\documentclass[10pt,letterpaper]{article}
\usepackage[top=0.85in,left=2.75in,footskip=0.75in]{geometry}

\usepackage[finalnew]{trackchanges}
\addeditor{RG}
\addeditor{NB}
\addeditor{New} % response to reviewers

\usepackage{lastpage}
\usepackage{enumerate}
\usepackage{mathtools}
\usepackage{caption}
\usepackage{afterpage}
\usepackage{pbox}
\usepackage{subcaption}
\usepackage{float}   
\usepackage{hyperref}
\usepackage{booktabs} % for much better looking tables
\usepackage{array} % for better arrays (eg matrices) in maths
\usepackage{paralist} % very flexible & customisable lists (eg. enumerate/itemize, etc.)
\usepackage{verbatim} % adds environment for commenting out blocks of text & for better verbatim
\usepackage[export]{adjustbox}
\usepackage{dirtytalk}
\usepackage{titlesec}
\usepackage{footmisc}
\usepackage{placeins}
\usepackage{colortbl}
\usepackage{multirow}
\usepackage{hhline}

\usepackage{wasysym} % for mars, venus

\usepackage{amsmath,amssymb}

\usepackage{changepage}

\usepackage[utf8x]{inputenc}

\usepackage{textcomp,marvosym}

\usepackage{cite}

\usepackage{nameref,hyperref}

\usepackage{microtype}
\DisableLigatures[f]{encoding = *, family = * }

\usepackage[table]{xcolor}

\usepackage{array}

\newcolumntype{+}{!{\vrule width 2pt}}

\newlength\savedwidth

\raggedright
\usepackage[aboveskip=1pt,labelfont=bf,labelsep=period,justification=raggedright,singlelinecheck=off]{caption}

\usepackage[table]{xcolor}
\definecolor{LightCyan}{rgb}{0.88,1,1}

\makeatletter
\renewcommand{\@biblabel}[1]{\quad#1.}
\makeatother

\date{}

\usepackage{lastpage,fancyhdr,graphicx}
\usepackage{epstopdf}
\fancyheadoffset[L]{2.25in}
\fancyfootoffset[L]{2.25in}
\long\def\comment#1{}
\def\ie{{\em i.e.},\ }
\def\eg{{\em e.g.},\ }
\def\etal{{\em et al.}}
\def\feat#1#2{{\sc #1}$_{#2}$} % or could be small caps or ...

\long\def\ShowFig#1{#1}  % set this to {} to NOT show figures...

\def\BG#1{\feat{BG}{#1}}
\def\mBG#1{\hbox{{\sc BG}}_{#1}} % for \BG, but in math mode

\def\dts#1{\Delta t_{#1}}

\def\Model#1{M_{#1}}

\def\err#1#2{\hbox{err}_{#1}(\, #2\,)}

\def\errL{err_{L1}}  % or maybe L1 ?
\def\errrL{err_{rLl}}  % or maybe rL1 ?

\definecolor{Gray}{gray}{0.85}
\newcolumntype{g}{>{\columncolor{Gray}}l}
\newcolumntype{w}{>{\columncolor{white}}c}

\begin{document}
\vspace*{0.2in}
% \title{PLOS: The Challenge of Predicting Blood Glucose of Patients with Type I Diabetes} % to tell Overleaf ...
\begin{flushleft}
{\Large
\textbf\newline{
The Challenge of Predicting Meal-to-meal Blood Glucose Concentrations for Patients with Type I Diabetes
}
% Please use "title case" (capitalize all terms in the title except conjunctions, prepositions, and articles).
}
\newline
\\
Neil C. Borle\textsuperscript{1},
Edmond A. Ryan\textsuperscript{2,4},
Russell Greiner\textsuperscript{1,3,*}
\\
\bigskip
\textbf{1} Department of Computing Science, University of Alberta, Edmonton, AB, Canada
\\
\textbf{2} Department of Medicine, Division of Endocrinology and Metabolism, University of Alberta, Edmonton, AB, Canada
\\
\textbf{3} Alberta Machine Intelligence Institute, Edmonton, AB, Canada
\\
\textbf{4} Alberta Diabetes Institute, Edmonton, AB, Canada
\\
\bigskip

% Insert additional author notes using the symbols described below. Insert symbol callouts after author names as necessary.
% 
% Remove or comment out the author notes below if they aren't used.
%
% Primary Equal Contribution Note
%\Yinyang These authors contributed equally to this work.

% Additional Equal Contribution Note
% Also use this double-dagger symbol for special authorship notes, such as senior authorship.
%\ddag These authors also contributed equally to this work.

%\textcurrency Current Address: Dept/Program/Center, Institution Name, City, State, Country % change symbol to "\textcurrency a" if more than one current address note
% \textcurrency b Insert second current address 
% \textcurrency c Insert third current address


% Group/Consortium Author Note
%\textpilcrow Membership list can be found in the Acknowledgments section.

% Use the asterisk to denote corresponding authorship and provide email address in note below.
* rgreiner@ualberta.ca

\end{flushleft}
\section*{Abstract}
Patients with Type I Diabetes (T1D) must take insulin injections to prevent the serious long term effects of hyperglycemia -- high blood glucose (BG). 
These patients must also be careful not to inject too much insulin because this could induce hypoglycemia (low BG), which can potentially be fatal. 
Patients therefore follow a ``regimen'' that, based on various measures, determines how much insulin to inject at certain times. 
Current methods for managing this disease require 
adjusting the % the adjustment of a 
patient's regimen over time based on the disease's behavior (recorded in the patient's diabetes diary). 

If we can accurately predict a patient's future BG values from 
his/her % their
current features 
(\eg predicting today's lunch BG value given today's diabetes diary entry for breakfast, including insulin injections, and perhaps earlier entries), 
then it is relatively easy to produce an effective regimen.
This study explores the challenges of BG modeling by applying a number of machine learning algorithms, as well as various data preprocessing variations 
(corresponding to 312 [learner, preprocessed-dataset] combinations),
to a new T1D dataset that contains 29\,601 entries from 47 
different % unique 
patients. 
Our most accurate predictor is a weighted ensemble of two Gaussian Process Regression 
models, which
achieved an $\errL$ loss of $2.70$ mmol/L \add[New]{(48.65 mg/dl)}. 
This was an unexpectedly poor result given that one can obtain an $\errL$ of $2.91$ mmol/L 
\add[New]{(52.43 mg/dl)} using the naive approach of 
simply % only
predicting the patient’s average BG. 
\add[New]{For each of data variant and model combination we report several evaluation metrics, including glucose-specific metrics, and find similarly disappointing results
-- again, the best model was only incrementally better than
the simplest measure.}
These results suggest that
the diabetes diary data that is typically collected may not be sufficient to
produce accurate BG prediction models;
additional data may be necessary to build 
 accurate BG prediction models.

%In this study we analyze a new T1D dataset which has not previously been studied using machine learning algorithms and which contains $55$ different patients with a combined number of $30,983$ diabetes diary entries. We explore a number of these algorithms as well as several different data preprocessing variations with the goal of building a automated system that can predict a patient's future blood glucose given a set of current features (e.g. predicting today's lunch blood glucose given today's diabetes diary entry for breakfast). The most accurate predictor that was developed was a weighted ensemble of two Gaussian Process Regression (GPR) models trained and tested  on individual patient data. One GPR model learns from a patients entire history while to other models specific meals. This ensemble was able to obtain an expected $L_1$ loss of $2.72 \frac{mmol}{L}$. This was an unexpectedly poor result, leading the authors to believe that very accurate blood glucose prediction models may not be obtainable from diabetes diary data that has been collected in a typically  clinical setting. This then suggests that improved data collection methods will be necessary if we are to build fine grain blood glucose control systems that are built upon blood glucose prediction models.

% Please keep the Author Summary between 150 and 200 words
% Use first person. PLOS ONE authors please skip this step. 
% Author Summary not valid for PLOS ONE submissions.   
%\section*{Author Summary}

%\linenumbers

% Use "Eq" instead of "Equation" for equation citations.
\section{Introduction}
\label{sec:intro}

Individuals suffering from Type I diabetes (T1D) 
are unable to produce insulin, 
meaning their bodies cannot properly regulate their blood 
glucose~(\BG{})~\cite{daneman2006type}
-- 
\ie 
cannot % generally
maintain their \BG{}\ between  
4 -- 8 mmol/L~\cite{kok2004predicting}.
As a result, T1D is a serious long term condition 
that can lead to
microvascular, macrovascular, neurolgical and metabolic complications~\cite{daneman2006type,kok2004predicting}. 

To manage their diabetes, patients give themselves periodic injections of insulin as directed by their health care team. Injecting too much insulin may induce hypoglycemia 
 (\BG{}$ < 4$ mmol/L, in our study), 
which can be dangerous, possibly causing a coma.
However, % patients should not attempt to avoid hypoglycemia by consistently 
injecting too little insulin %; this 
can % will 
result in hyperglycemia (\BG{}$ > 8$ mmol/L, in our study), 
which may 
lead % give rise
to chronic complications such as blindness, kidney failure, nerve damage and circulatory problems~\cite{daneman2006type,kok2004predicting}.
In general,
a patient's \BG{}\ % at a given time
will depend on many factors, 
including past carbohydrate intake, the amount of bolus/basal insulin injected,
exercise, and stress~\cite{kok2004predicting}.

\comment{
While composing only
$5 - 10\%$ of diabetes cases, 
Type I diabetes (the most severe type of diabetes~\cite{plis2014machine}) is a serious chronic condition,
that % and 
is strongly associated with microvascular, macrovascular, neurolgical and metabolic complications~\cite{daneman2006type,kok2004predicting}. Patients suffering from Type I diabetes 
are unable % lose the ability
to produce insulin, 
meaning % resulting in the inability for 
their bodies 
cannot % to
properly regulate blood glucose~\cite{daneman2006type}.
This means that while normal human blood glucose levels are between $4 \frac{mmol}{L}$ and $8 \frac{mmol}{L}$, 
these
patients often experience hypoglycemia (blood glucose less than $4 \frac{mmol}{L}$) and hyperglycemia (blood glucose greater than $8 \frac{mmol}{L}$)~\cite{kok2004predicting}. 
It is important to note that while hyperglycemia negatively impacts an individual's health (particularly over time), hypoglycemia can be much more costly~\cite{daneman2006type,kok2004predicting}. Several factors affect a patient's blood glucose such as exercise, stress, carbohydrate intake, amount of bolus/basal insulin injected~\cite{kok2004predicting}.
}

% In general, 
Diabetes patients try to properly maintain their \BG{}\ in a normal range. 
This is challenging because tight glycemic control using bolus insulin injections (whether intermittent with insulin pens or boluses using insulin pumps) is associated with an increased the risk of having hypoglycemic events~\cite{daneman2006type}.
This challenge has led to attempts to create closed-loop systems and the use of computational techniques that assist in controlling patient's \BG{}\ levels~\cite{bastani2014model}. An extreme example of this is the effort to create an 
``artificial pancreas'',
which explicitly integrates automatic monitoring with automatic administration of insulin~\cite{lunze2013blood}. 

Another perspective on fully automated diabetes management views the \BG{}\ control problem as two sequential subproblems:
\begin{enumerate}
\item ``modeling'': learning an accurate {\em \BG{}\ prediction model}\
that, for example, predicts the \BG{}\ level at lunch 
given a description of the subject throughout breakfast
(including perhaps her previous \BG{}\ values, carbohydrate intake, etc., from earlier meals), 
as well as the amount of insulin injected at breakfast.

\item ``controlling'': given the current information (at breakfast), 
consider the effects of injecting various possible amounts of insulin
-- \eg \{1 unit,\ 1.5 units, 2~units, \dots\}.
For each, use the learned model to predict the \BG{}\ value at lunch,
then 
inject the amount that is predicted to lead to the best lunch time 
\BG{}-value 
(Of course, this assumes that decisions made at breakfast only affect lunch,
then lunch decisions will only affect dinner, etc.\ -- 
which does not consider the longer-range effects of actions;
see Bastani~\cite{bastani2014model}).
\end{enumerate}

This paper focuses on the first subtask: developing a 
\BG{}\ prediction system\change[New]{. We use machine learning techniques to 
learn models that can be used to estimate an individual's 
future BG using covariates that describe the current patient.
%\subsection{Our Specific Task} % 1.1
%\label{sec:SpecificTask}
In}{, where in}
 general, a model $\Model{}$ will predict the blood glucose 
$\widehat{\mBG{i+1}}
\ =\ \Model{}(\, x_{i},\, \dts{i+1}\,)$ 
at the next time point ($\dts{i+1}$ minutes into the future), 
based on information currently known about this patient,
including the amount of insulin 
(\feat{bolus}{i})
the patient decided to inject (%We define these terms later, in Table~\ref{tab:orig_features}.
Note that this abstracts some issues; see Appendix\hbox{~\ref{preprocessing_variants}} for details):

\begin{equation}
x_i\quad =\quad [\hbox{\feat{time}{i}, \ \feat{BG}{i},\ \feat{bolus}{i},\ \feat{basal}{i},\ 
\feat{EV}{i},\ \feat{PV}{i},\ \feat{IOB}{i},\ \dots}]
% x_i\quad =\quad [time_i, \ \hbox{$BG_{i}$},\ \hbox{$bolus_{i}$},\ \hbox{$basal_i$},\ 
% \hbox{$ExV_i$},\ \hbox{$PV_i$},\ \hbox{$IOB_i$},\ \dots]
% This is \Delta t^{bolus}_i .. in case have to go back.
\label{eqn:xi-info}
\end{equation}
\change[New]{(Think of predicting the blood glucose at 
your
12:01pm lunch on Tuesday,}{An example of this subtask might be to predict an individuals blood glucose at 12:01pm lunch on Tuesday, }
given information collected up-until 8am breakfast on Tuesday\change[New]{. Note that this could be only}{(Where this could only include} the Tuesday breakfast information, 
or it could include other earlier information -- \eg the ellipses in Equation~\ref{eqn:xi-info} might contain information about events from yesterday, or last week).
%See also Table~\ref{tab:orig_features}.

\add[New]{Explicitly, the goal of this work is to to determine if it is possible to accurately predict a T1D patient's recorded BG, from one meal to the next, based only on the information typically recorded by the patient. To do this, a model must be able to deal with the data provided in a patient's diabetes diary, which has varied prediction horizons.}

\note[New]{Last paragraph of the introduction moved here}

This work is an extensive effort to learn an {\em accurate}\ \BG{}\ prediction model, 
which involved exploring 312 different combinations of
learner and preprocessing variant.
To train and evaluate each of these variants,
% For the training of our models, 
we used a dataset % consisting 
of 29\,601 
{entries} collected from 47 unique patients,
where
each {entry}
included the information typically collected,
including: 
the time of day,
the patient's current \BG{}, 
the carbohydrate about to be consumed and the anticipated exercise.

\subsection{\change[New]{Limitations of Prior Diabetes Modeling Studies}{Background Literature and its Limitations}} % 1.2
\label{sec:Limitations}
{As the mechanism of diabetes are not completely understood,
we are considering the machine learning approach
of learning a model, 
from a large labeled dataset, containing many records from many Type-1 diabetes patients.
Of course, this requires access to such a dataset.
}
Previous studies have been based on data from small numbers of subjects and/or data collected over a short time period.
For example, several studies have been based on data from a single patient where records were only collected for fewer than 100 days~\cite{tresp1999neural,kok2004predicting,baghdadi2007controlling,Zainuddin2009neural}. 
These studies are limited because different patients behave very differently.
Other studies include more patients (12 -- 15) but only have 3 -- 22 days worth of data~\cite{georga2015online,andreassen1994probabilistic}.
Another study used three patients with two years of data~\cite{magni2006stochastic}. \add[New]{Further, studies analyzing  continuous glucose monitoring (CGM) data from larger patient sets (89 T1D patients) exist but are again sampled over relatively short periods of time (1 week)}~\cite{gadaleta2018prediction}.
In these last \change[New]{three}{four} cases, datasets either had short histories for their patients or they had a small number of patients in total.
In contrast, our work uses a larger number of patients who had up to two years worth of data, where records were collected multiple times each day.

There are % exist 
large datasets of type-2 diabetes patients --
\eg Quinn~\etal~\cite{quinn2011cluster} % who 
measured glycated hemogloben changes in data collected data from 163 patients over the course of a year.
However, studies that model type-2 diabetes~\cite{chemlal2011blood,sudharsan2014hypoglycemia} should not be directly compared to those that model type-1 diabetes because there are significant differences between these diabetes types.
In particular, there is less variance in the blood glucose readings over time for type-2 patients than there is for type-1 patients, 
making type-2 patients easier to model.

While we focus on predicting \BG{}\ values {\em many hours later},
some studies instead attempt to predict the occurrence of hypoglycemic events,
and only within a short window 
(\eg 30 to 120 minutes)~\cite{bunescu2013blood,eren2010hypoglycemia,georga2015online,pappada2011neural,plis2014machine}. 
\add[New]{Indeed, a literature review by Contreras~\hbox{\etal} identified 49 publications using modeling techniques for blood glucose prediction (primarily with T1D patients) of which 38 used predictions horizons that were 60 minutes or less~\hbox{\cite{contreras2018artificial}}.
Of these, one of these publications included prediction horizons of 180 minuets but on simulated data, and another had 1440 minute prediction horizons but used a dataset of 8 patients collected over only 3 days~\hbox{\cite{contreras2018artificial}}.} 
While this might help to protect patients from a very serious situation, 
it is lacking in several ways. 
First, such fine-grained measurements are often not practically obtainable outside of a study setting and without using a \change[New]{continuous glucose monitoring  (CGM)}{CGM} device that provides measurements every 5 minutes.
Second, these short-term predictions are
not adequate for spanning the time between meals.
Third, the goal of building a diabetes control system is better served with a more expressive model,
as opposed to one that can only provide binary classifications
-- hypoglycemic or not. 
Note that these model
provides no useful feedback for situations where patients are hyperglycemic.

In our work, we try to model blood glucose dynamics 
(including both hyperglycemia and hypoglycemia) and 
using only the standard records collected at meal times.
While this makes our task more challenging,
we do this because it involves only the data that medical professionals most often encounter in practice.

\remove[New]{
\subsection{Additional Background} % 1.2
\label{sec:Addition-BG}

This section describes several techniques
within the diabetes modeling literature,
including the approaches that we used.}

Like Pappada \etal~\cite{pappada2008development},
we also considered neural network models.
After training on 17 T1D patients,
Pappada~\etal's~\cite{pappada2008development} evaluation, on a single held-out patient,
yielded ``scores'' 
(a version of the rL1 measure defined in Equation~\ref{eqn:rL1-loss})
of 0.067
(resp., 0.089, 0.117, 0.145, 0.166, 0.189)
when using predictive windows of 50
(resp., 75, 100, 120, 150 and 180)
minutes in the future.
While 
they were % Pappada \etal~are 
able to achieve a score of 0.189 using 180 minute predictive windows, 
% it is important to 
note that their result is based on the test data of a single patient (other patients may be more difficult to predict) and that the dataset used in 
their study involved relatively short-range predictions
(50 -- 180 minutes),
while our study involves predictions 
made, on average, 
310.6 minutes in the future 
(averaging 593 minutes for overnight predictions and
236 minutes otherwise). 
Since Pappada \etal~showed that larger predictive windows decrease the accuracy of their models, 
we expect that our data should be more difficult to model well.
Also, 
their study involved
% note that the participants from their study provided 
only 3 to 9 days of data with continuous glucose monitoring, 
whereas our data were collected over a period of months to years. 

Our work resembles
previous works~\cite{valletta2009gaussian,duke2010intelligent}
that use Gaussian Process Regression (GPR)
for modeling diabetes. 
In particular, Duke~\cite{duke2010intelligent} used GPR to learn models of individual patients that could 
be
used to aid in cross-patient prediction. 
We similarly explore some transfer learning techniques with GPR, 
along with ensembles of learners and various other machine learning algorithms. 

Prior works have also addressed the 
blood glucose modeling problem that we 
explore in this work~\cite{kok2004predicting,baghdadi2007controlling,Zainuddin2009neural}.
These latter two % both Baghdadi~\etal\ and Zainuddin~\etal\ 
evaluate their results using normalized blood glucose values;
since they are unitless normalized values,
they cannot be directly interpreted in terms of mmol/L,
which % This also
means that we cannot compare our results to theirs. 
However, 
% in our results,
we do evaluate the performance of a model that 
is similar to the Gaussian Wavelet Neural Network used in 
the third%Zainuddin~\etal
~\cite{Zainuddin2009neural}.

\add[New]{
One work of interest to our study is the previously mentioned 2018 publication by Gadaleta~\hbox{\etal~\cite{gadaleta2018prediction}} that analyzes CGM data. In comparison to our work, this work looks at shorter term blood glucose predictions over a shorter period of time but is similar in that it provides a fairly comprehensive analysis of the predictive performance of many different machine learning models. Gadaleta~\hbox{\etal} focus on two different methods for training and evaluating models, ``static'' and ``dynamic''. Similarly to their ``dynamic'' training process, the majority of our study focuses on training and testing on data from each patient separately. However, the stacking models used in our work effectively train a ``static'' model with a Leave One Patient Out and combine this model with a patient-specific model. Gadaleta~\hbox{\etal} consider combining ``static'' and ``dynamic'' training processes as future work.
}

\add[New]{There are other works that define measures for evaluating the quality of glucose predictors, in general.
In particular, 
Del Favero~\hbox{\etal~\cite{del2012glucose}}
describes several different measures for comparing 
a patient's specific glucose reading, with a predicted one,
including both standard measures (like L1, relative L1 error, and L2 losses -- there called MAD, MARD and RMSE) and some  ``glucose-specific metrics'', such as gMAD and gRMSE~\hbox{\cite{clarke2005original}}.
While our paper focuses on the L1 and relative L1 errors,
we also include the others mentioned there. 
}

\subsection{Main Contributions} % 1.4
\label{sec:main-contrib}
Below we list the main contributions of this work:
% The contributions of this work are as follows: 
\begin{enumerate}
\item 
To our knowledge, this study examines the largest multi-year dataset of diabetes diary records, 
collected from Type~1 diabetes patients, used for modeling future \BG{}.
\comment{
 this is the largest multi-year study of diabetes diary data, collected from Type~1 diabetes patients, that has been used for \BG{}\ modeling.}

\item
We provide a comprehensive study of this data,
considering 
312 combinations of learning algorithm and type of data,
to determine if machine learning can be used to create 
an accurate blood glucose prediction model.

\item
Our results demonstrate % We illustrate 
that it is difficult for a machine learned model to perform 
better than a naive baseline (in this case, predicting a patient's average \BG{}),
\add[New]{when considering both standard error measures, like L1 and L2 loss, and also for glucose-specific measures, such as gMAD.}

%\item
%We compare our best model to a human expert and show that both perform similarly on this \BG{}\ prediction task.
\end{enumerate}

The 
\add[New]{publically available}
MSc thesis~\cite{borle2017challenge} that corresponds to this work provides additional information, including
a breakdown of the individual patients in the study, more detailed results, 
and a comparison to a diabetologist's performance on this prediction task.

\section{Materials and Methods} % 2

Section~\ref{Sec:data} first summarizes how we obtained this (real world) data;
Section~\ref{sec:preproc} then describes the pre-processing steps required to make this data usable. 
We then consider two ways to modify this dataset:
Section~\ref{sec:EP} considers modifying the set of records;
one class of studies involves the complete set of entries, and 
another class included just the subset of ``Expert Predictable'' records
(defined below).
For each of these two ``sets of record'',
we consider various ``feature sets''
-- the original set of features, and also 12 other variants,
each of which includes various new features, 
that are combination of those original features; 
see Section~\ref{sec:FeatureEng}.
Section~\ref{sec:ML-algs} then summarizes the 12
different learning algorithms we considered 
(based on 7 distinct base learners);
and Section~\ref{sec:Naive} describes a trivial baseline, to help us determine whether the results of any of the learned models are actually meaningful.
%\footnote{Spoiler alert: they are not.}
This requires describing how we evaluate the quality of a learned model -- see Section~\ref{sec:ModelEval}.
(This segues naturally to Section~\ref{sec:results}, which provides those empirical results.)

\subsection{Dataset} % 2.1
\label{Sec:data}

This study used 47 patient histories from Type I diabetes patients, which were collected using the ``Intelligent Diabetes Management'' (IDM) software%
% \footnote{
% The website \url{https://idm.ualberta.ca/} has since been decommissioned. 
% }
(described in Ryan~\etal~\cite{ryan2017improved}). Note that the associated website \url{https://idm.ualberta.ca/} has since been decommissioned.
This data included patients who participated in Ryan~\etal's study, 
as well as additional patients who began using the IDM software after the completion of the study (up until December 2016). 
The participants gave their informed written consent, and the Research Ethics Board of the University of Alberta approved the collection and analysis of the data. 
For further details regarding patient participation, see Ryan~\etal~\cite{ryan2017improved}.
Some of the participants only used the system a few times. 
As we wanted to focus on patients that had sufficient information
to find relevant patterns,
% To get meaningful information about each patient,
we
only included patients who made at least 100 diabetes diary entries with the system -- 
\ie produced at least 100 ``sufficient'' records.
This led to a dataset % consisting 
of 16 pump users and 31 non-pump patients.
Table~\ref{tab:demsummary} provides summary statistics for our data.
The dataset used for this work differs from the one described in Borle~\cite{borle2017challenge} 
in that we limit the number of patients included to those with complete data.
%
% \footnote{
% %\note[RG]{Eddie: do we need this? Ie, do you think any reader will actually
% %go to Neil's MSc, and look this carefully?}
% The dataset used for this work differs from the one described in Borle~\cite{borle2017challenge} 
% in that we limit the number of patients included to those 
% with complete data.
% % as it has removed patients \#48 and \#51 
% % as they had incomplete demographic information. 
% % % Further note that two pairs of patients, \#4 and \#25, and \#5 and \#29, are two individuals who participated in the study twice and therefore each producing two separate patient histories (which were each combined for the purpose of this study).
% % Further note that the pair of patient-ids, \#4 and \#25, 
% % both correspond to the same patient who participated in the study twice;
% % similarly for \#5 and \#29.
% % Each pair is combined in this current study.
% % Patient \#16 also produced two patient histories,
% % however we used only one for evaluation (as described later). 
% }
%In total, this gave us 49 patient histories, which were used for evaluating model performance.
% and Table~\ref{tab:demographics} presents the complete demographic information for all patients included in the study.

%The data collected in this study also includes particularly noteworthy patient, 
Patient \#16 % (See Borle~\cite{borle2017challenge}) 
is 
% a particularly 
noteworthy % individual, 
for having % who has
by far the most records of any patient in our dataset;
it is unusual for a patient to consistently produce diabetes entries over the course of many years. 
Because of the large number of records, 
we use part of this patient's dataset as our hyper-parameter tuning (validation) dataset, as well as for visualization.
%(such as Figure~\ref{fig:118_gpr} later).

\begin{table} % [!ht]
% \begin{adjustwidth}{-2.25in}{0in} % Comment out/remove adjustwidth environment if table fits in text column.
\centering
\caption{
{\bf Summary of Demographics}}
% \note[RG]{Why adjust the width? Does PlosOne require this?? Also "Distinct" not 'Unique'}
\small
\begin{tabular}{|c|c|c|c|c|c|c|}
\hline
\rowcolor{red!50} Distinct % Unique
Patients & Age & Height & Weight & Sex & Pump Users\\ \hline
47 & $42 \pm 13$ & $166 \pm 8$ cm* & $74 \pm 14$ kg & 47 (9 $\mars$ / 38 $\venus$) & 16\\ \hline
\end{tabular}
\label{tab:demsummary}
\vspace{0.05cm}
\begin{flushleft}
* Height 
could not be obtained for 7 individuals,
so this average value was calculated using only the remaining patients.
See Borle~\cite{borle2017challenge} for more details about individual patients.
\end{flushleft}
% \end{adjustwidth}
%\note[RG]{Don't you need to know the Pump-status of every user?  Shouldn't there be 52 of them?}
% Pregnant: Confusing have 41 for denominator. How does this relate the 32 who are female?}\note[NB]{41 was the number of responses to the question (including men), I've changed this to be out of the 32 self-identified women, 31 provided an answer where 1 person was pregnant.
\end{table}

Each record $i$ %associated with a patient 
corresponds to an entry in a patient's ``diabetes diary'',
which includes the meal associated with the record \feat{meal}{i},
a time stamp (\feat{date}{i}\ and \feat{time}{i}), 
%\annote[RG]{the time when the event took place,}{? do we use this? It is not in the tables...}\note[NB]{Not directly, but it is require during preprocessing to produce all the time related features}
the blood glucose value \feat{BG}{i},
the grams of carbohydrates consumed \feat{CHO}{i}, 
and the units of bolus (resp., basal) injected \feat{bolus}{i}, 
% and basal insulin injected 
(resp., \feat{basal}{i}).
%\remove[RG]{and the level of anticipated future activity $ExV_i$.}
The patients also entered the anticipated level of exercise
using the non-numeric values \{``less than normal'', ``normal'', ``active'', ``very active''\}.
%Following advice from our expert diabetologist\footnote{Dr. Edmond A. Ryan, a co-author},
We converted these into 
% machine interpretable
numeric values ($2$, $4$, $7$ and $10$ respectively)
for use by standard learning algorithms.

\begin{figure}[t] % 1
 \centering
 \ShowFig{
\includegraphics[width=%0.75
\textwidth,height=2in]{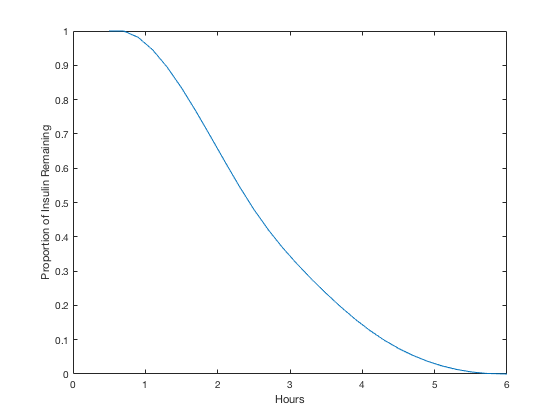}
}
\caption{{\bf Spline of Insulin on Board over Time.}}
\label{fig:IOB_spline}
\end{figure}

As was mentioned, 16 of the patients in this study used insulin pumps,
which each % . These pumps work by 
directly infuse insulin from a reservoir, via a catheter, just under a patient's skin at a basal rate. 
%\change[RG]{However, they are also used to}
{Moreover, they also}
self-inject larger amounts of bolus insulin when a patient ingests carbohydrates (as a patient would with an insulin pen 
\url{http://www.diabetes.org/living-with-diabetes/treatment-and-care/medication/insulin/how-do-insulin-pumps-work.html}).
{ Each record of each insulin pump patient 
includes the basal infusion rate value \feat{PV}{i},
in $\frac{units}{hour}$.
}
The insulin pump settings work by partitioning the 24h clock into intervals,
and setting a particular delivery rate of insulin for each interval. 
The \feat{PV}{i}\ values for any specific record 
was then 
set to  the % found by obtaining the corresponding 
insulin delivery rate for the interval containing the record's time stamp
We also computed two other features: 
$\Delta t_i$, which is the elapsed time since the previous record (Actually, $\Delta t_i$ is based on previous 
\feat{bolus}{i-1}\ and \feat{CHO}{i-1}\ values; see Appendix~\ref{preprocessing_variants}) 
and 
% 
% , which is the amount of residual insulin from previous bolus injections. 
% We use the 
``Insulin on Board'' \feat{IOB}{i}, 
% (see the 11th row in Table~\ref{tab:orig_features})
which captures the effect of any insulin remaining in a person's system from previous injections~\cite{al2015smart}. 
This was based on the following pairs of elapsed time and 
percentage of post-injection insulin remaining~\cite{mudaliar1999insulin}: 
(1.66 hours, 78\%), (2.5 hours, 48\%), (3.33 hours, 27\%), (4.15 hours, 12\%), (5 hours, 3\%). 
We then used a simple spline to interpolate these values; see % and is visualized in 
Fig~\ref{fig:IOB_spline}. 
Table~\ref{tab:orig_features} formalizes all of these features and Table~\ref{tab:rawdata} provides example data.

\def\ebolus#1{\feat{bolus}{#1-}}
\def\eCHO#1{\feat{CHO}{#1-}}
\def\normal{} % or {(normal)}

\subsection{Data Preprocessing} % 2.2
\label{sec:preproc}
The first step in analyzing % processing
the data was dealing with the missing or erroneous (unreasonably low) values.
We discarded any record that did not have an associated \BG{}\ value (572 records).
This was necessary as we cannot evaluate a model on records where we do not have a ground truth for \BG{}. 
%\note[RG]{Move this deletion here, by other deletion.}\note[NB]{I'm not sure what is meant by this comment}
We also discarded any records that had missing dates (6 records)
because these timestamps are integral to deriving features from the data. 
%\change[RG]{After this, we take records with BG values less than $1 \frac{mmol}{L}$ (8 records) and set the BG values to $1 \frac{mmol}{L}$.}
Second, we changed any \BG{}\ value less than 1 mmol/L (8 records) to 1 mmol/L,
% We rounded these eight values to 1 mmol/L because 
as glucose meters are simply not accurate at these low values,
other than to state % confirm
that the values are very low.
%We did this based on the assumption that the glucose meter reported an erroneous BG value because it is not likely that a patient would be healthy enough to report or survive such low values.
We then log-transformed these blood glucose values for all our predictors,
anticipating that this log-linear model would have better performance.
% as it may dampen the variation.
This means that after this model makes a prediction,
we must use simple exponentiation to transform that prediction 
back into the original interpretable units.
\comment{ Furthermore, 
our subsequent processing steps involve creating features based on the blood glucose values at previous meals;
this cannot be done if those previous blood glucose value is missing.}

To address missing \feat{bolus}{}\ insulin and \feat{CHO}{}\ carbohydrate values,
we imputed average values into the missing entry 
(variants for this step are described in Section~\ref{sec:FeatureEng}). 
This was done on a per-subject, per-meal basis
-- that is, we imputed % we look at 
an individual's average value for a particular meal.
For example, say a specific patient injected, on average, 3 units of bolus insulin before breakfast. 
Whenever she does not enter the before-breakfast insulin,
we replace that missing value with ``3 units''. 
When dealing with missing exercise values, we imputed the ``normal'' value.
% instead of an average value as is done with insulin and carbohydrates. 
For missing basal insulin values, we always imputed a constant value of 0.
%\note[RG]{Explain... is this constant value different from other values??}
This allows a learner to distinguish when basal insulin was recorded and when it was not.
% for patient $i$ we would determine what the average units of bolus insulin injected was before before breakfast. Then for any before breakfast records where the injected bolus insulin was missing, we would replace that missing value with the average before breakfast value for that individual. When dealing with missing exercise values, we impute the ``normal'' value, instead of an average value as is done with insulin and carbohydrates.
After this preprocessing, we computed the auxiliary features 
($\Delta t_i$, \feat{IOB}{i}) from the improved data. 
We describe the complete set of features in Table~\ref{tab:features}, and we show example records as columns in Table~\ref{tab:prodata}; 
see also Appendix~\ref{preprocessing_variants}.

\begin{table}[t] % [!ht]  % this table was AFTER the example...
\begin{adjustwidth}{-2.25in}{0in} % Comment out/remove adjustwidth environment if table fits in text column.
\centering
\caption{
{\bf Description of Original Features, and some Computed Features, used in this Study}}
\label{tab:orig_features}
% \note[RG]{Why not have the order of features here, match text earlier?\\ Should first have "basic values", then computed values.}

\small
\begin{tabular}{| >{\columncolor{LightCyan}}l|p{12cm}|}
\hline\hline
\feat{meal}{i} & The time of day:
\{ Before Breakfast, After Breakfast, Before lunch, After Lunch, Before Supper, After Supper, Before Bed, During the Night\}\\ \hline
\feat{date}{i} & The date as year-month-day\\ \hline
\feat{time}{i} & The time as hour:minute:second\\  \hline
\feat{BG}{i} & The BG value at the current time ($\frac{mmol}{L}$) \\ \hline
\feat{CHO}{i} & The amount of carbohydrates ingested (grams)\\ \hline
\feat{bolus}{i} & The amount of insulin injected (units)\\ \hline
\feat{basal}{i} & The units of background insulin injected\\ \hline
\feat{EV}{i} & Numeric encoding of exercise value: $\{2, 4, 7, 10\}$ \\ \hline
\feat{PV}{i} & Pump Value: The rate at which the insulin pump is infusing ($\frac{units}{hour}$). 
This is always 0 if the patient does not have a pump. 
%and \annote[RG]{is replaced by}{What does this mean?} \feat{basal}{i}
\\
\hline
\hline
$\Delta t_i$ & The elapsed time since last record\\ \hline
\feat{IOB}{i} & Insulin on Board: Estimated residual insulin from the previous injection ($\frac{mmol}{L}$)\\ \hline
\end{tabular}
\vspace{0.05cm}
\begin{flushleft}
See text for further description of these terms.
Note this is a simplified set of features;
see Table~\ref{tab:features} in the Appendix for the complete set of feature descriptions.
\end{flushleft}
\end{adjustwidth}
\end{table}

\begin{table}[t] % [!ht]
\begin{adjustwidth}{-2.25in}{0in} % Comment out/remove adjustwidth environment if table fits in text column.
\centering
\caption{
{\bf Example of Data, over a single day, from Patient 16}}
\label{tab:rawdata}
\small
\begin{tabular}{|g|c|c|c|c|c|c|}
\hline
index $i$ & 27 & 28 & 29 & 30 & 31 & 32\\ 
\hline
\feat{meal}{i} & Before Breakfast & After Breakfast & Before Lunch & After Lunch & Before Dinner & After Dinner \\ %\hline
\feat{date}{i} & 2015-11-25 & 2015-11-25 & 2015-11-25 & 2015-11-25 & 2015-11-25 & 2015-11-25 \\ %\hline
\feat{time}{i} & 08:36:00 & 10:19:00 & 12:19:00 & 15:35:00 & 18:42:00 & 20:11:00\\ %\hline
\feat{BG}{i} & 16.2 & 14.7 & 5.6 & 6.8 & 10.5 & 3.0\\
\feat{CHO}{i} & 30.0 & 0 & 30.0 & 0 & 15.0 & 0\\ %\hline
\feat{bolus}{i} & 10.4 & 0 & 3.0 & 0 & 3.8 & 0\\ %\hline
\feat{basal}{i} & 0 & 0 & 0 & 0 & 0 & 0 \\ 
\feat{EV}{i} & 4 \normal& 4 \normal & 4 \normal & 4 \normal & 4 \normal & 4 \normal\\ %\hline
\feat{PV}{i}  & 0.50 & 0.50 & 0.63 & 0.45 & 0.90 & 0.90\\ 
\hline
$\Delta t_i$ & 540 & 103 & 120 & 196 & 187 & 89\\% & 615\\ %\hline
\feat{IOB}{i} & 0.00 & 7.90 & 3.61 & 0.89 & 0.81 & 3.35\\ %\hline
\hline
\end{tabular}
\vspace{0.05cm}
\begin{flushleft} 
Note this is a simplified version of the data; 
% For more details on the complete set of features, see 
Table~\ref{tab:prodata} in the Appendix provides the general, complete set of features. 
% Further, note that \feat{IOB}{i}\
% \note[RG]{Why mention this one feature?}
% specifically refers to residual bolus insulin.

\end{flushleft}
\end{adjustwidth}
\end{table}

\comment{
\subsubsection{Using Log of BG values} % 2.2.1 Training in the LOG Space
\label{sec:log}
\comment{ \note[RG]{Being positive means we CAN do it.  
But it does not mean that we SHOULD do.  
Do you have any reason for why log-xform is good?}
\note[NB]{Only that this might help with model misspecification for linear models}
}
As our blood glucose are always at least $1 \frac{mmol}{L}$
(after pre-processing),
we decided to train our models with log transformed blood glucose values,
anticipating this log-linear model might have better performance.
% as it may dampen the variation.
After this model makes a prediction,
we use simple exponentiation to transform that prediction into an interpretable result.
}

\subsection{Subset of only ``Expert Predictable'' Entries} % 2.3
% On Deciding Whether to Predict a Glucose Value
\label{sec:EP}

%\note[RG]{This used to be Sec 2.2.1 -- but changed to be 2.3.}
As our data was collected voluntarily 
from patients at their own convenience, 
sampling intervals are not uniform, %  in the data 
and 
the relevant
data is not 
recorded % available
for every meal. 
% This becomes problematic when modeling the data because blood glucose values become more difficult to predict as more time is allowed to elapse between readings. 
This is problematic for our predictive task as blood glucose values are more difficult to predict as more time % is allowed to 
elapses between readings. 
To address this issue,
our clinician co-author (E.A.R.) established the following criteria of when it is % may be
reasonable % to attempt 
to predict the next glucose value;
\comment{
we asked an expert diabetologist (E. A. Ryan) when an expert would feel comfortable making a prediction, given a patient's history. % From this discussion, we developed the following criteria.
This discussion led to the following criteria --
}
the \BG{}\ is ``expert predictable'' (EP) at a given time
% that one should only make a prediction for a given time-point
if all of the following are true:

\begin{enumerate}
\item The preceding record 
is not % cannot be
a hypoglycemic event (
Note that this is difficult to predict 
due to potential glucose counterregulation effects~\cite{gerich1988glucose} and the uncertainty in \BG{}\
that follows from a physiological response to hypoglycemia).

\item 
% In order to make a prediction, a 
The blood glucose reading is % must be
present for the preceding meal.
For example, to make a prediction about a patient's blood glucose value before lunch, 
a record detailing his/her previous breakfast must be available.

\item
Six of the last eight days prior to a prediction must have records for 
both the 
current meal time and the previous meal time. 
For example, to predict the blood glucose before lunch, 
six of the last eight days must have both ``before lunch'' and 
``after breakfast'' entries,
to help capture 
this ``after breakfast to before lunch'' transition pattern.
\end{enumerate}

\begin{figure}[t] % 2
\centering
\ShowFig{\hspace*{-0.1in}\includegraphics[width=1.2\textwidth,height=1.95in]{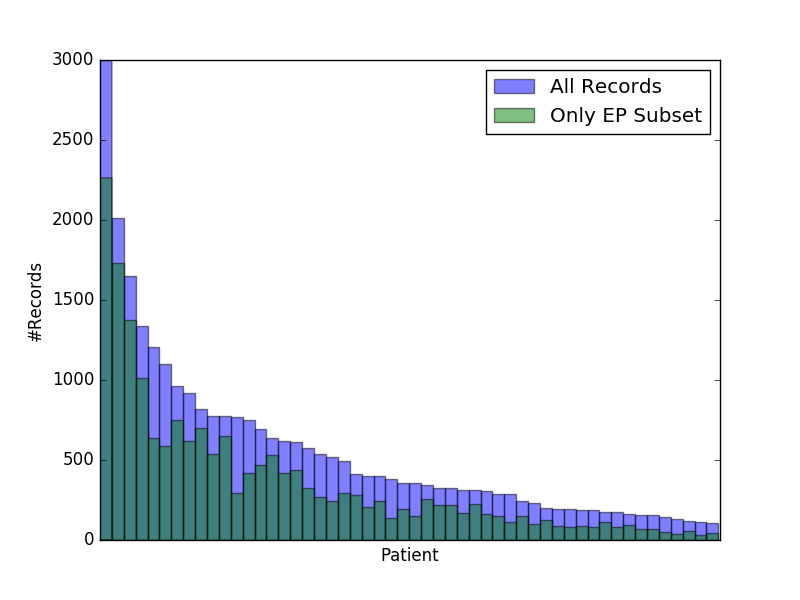}
}
\caption{{\bf Records Meeting the EP Criteria From Section~\ref{sec:EP}.} 
Patients are sorted by descending total numbers of records. 
See Borle~\cite{borle2017challenge} for further details.}
\label{fig:eddie_rule}
\end{figure}

Fig~\ref{fig:eddie_rule} shows the number of 
records from each patient that 
qualify as EP -- the number of records for which our expert would feel comfortable making predictions.

Later, we trained and evaluated 13 models using the entire dataset $D$
\change[New]{(called \{ D13, D14, \dots, D25\})}{(called \{ $D_{a}$1, $D_{a}$2, \dots, $D_{a}$13\})}
%\note[RG]{Can these be number 1..13 -- so the numbers match: $D_a$3 corresponds to $D_e$3, etc ??}
as well as 13 corresponding models 
that were trained  
% \change[RG]{on $D$ and }
{and then}
% {Below -- li1198-ish -- you claim you TRAIN on $D^{EP}$ before testing on $D^{EP}$??}
evaluated, using only the records
that met the expert's EP criteria, $D^{EP}$
\change[New]{(called \{ D0, D1, \dots, D12 \})}{(called \{ $D_{e}$1, $D_{e}$2, \dots, $D_{e}$13 \})}. 
\add[New]{Note that the $D_a$ and $D_e$ notation reflects that a dataset was derived from either 
all % \hbox{\em \underline{a}ll}\ 
the data points ($D_a$) or from the 
 expert's subset % \hbox{\em \underline{e}xpert's subset}, 
$D^{EP}$ ($D_e$).}

\subsection{Feature Engineering} % 2.4
\label{sec:FeatureEng}  % Preprocessing} % variants

%\note[RG]{Should this be earlier -- just after EP section? }
%\note[RG]{Rewrote this -- original appears in comment below.}
Table~\ref{tab:orig_features} shows the basic features used to describe each event.
Additionally, we also considered many other feature sets 
to see if any could lead to better performance.
Some of the variants completed records that were missing entries for carbohydrates or bolus insulin, 
which others % and some 
removed those deficit records.
Some added in the day of the week as an integer feature or as a one-hot encoded feature
(\url{http://scikit-learn.org/stable/modules/generated/sklearn.preprocessing.OneHotEncoder.html}), 
while others removed the ``basal insulin'' feature.
A few variants included non-temporal patient characteristics: 
age, gender, height and weight. 
Some replaced the 
set of features with just % features
the first 4 principal components 
(obtained by principal component analysis, PCA). 

Our ``Kok Features'' variant  
uses computed features similar to Kok~\cite{kok2004predicting}, 
and subsequently used by Baghdadi~\etal~\cite{baghdadi2007controlling} and Zainuddin~\etal~\cite{Zainuddin2009neural}. 
% Note that 
Unlike Kok's data, however,
we do not have stress level values in our data and were therefore unable to incorporate that feature.

For any given dataset variant, 
some models have % distinct 
components that train on different subsets of the data. 
In addition,
% the typical model that we use (training and testing on all the data from a single patient), 
we also use models that include components that are trained on the data from all patients other than the current test patient% under test
, as well as models that involve sub-models 
that are each trained only on data from one meal type (\eg before breakfast).

%\note[RG]{Moved this here, from earlier...}
Note we consider each of these variants applied to the original dataset, and also to the reduced ``EP-filters'' datasets;
see Section~\ref{sec:EP}.
% Finally, some variants applied the EP filter (from Section~\ref{sec:EP})
% to remove problematic 
% situation % records
% -- \ie ones that (our expert believes) do not contain enough information to be reliably predicted.
Note that this reduces the predictions that our system attempts. 
For example, we do not attempt to predict a 
before-lunch % after-breakfast 
\BG{}\ value if there is no preceding 
after-breakfast reading.
%\note[RG]{Example was before-bkft to after-bkft.  Do you really need 2 readings for each meal?}
{For each variant, we only considered the subset of the records that belonged to that variant,
both for producing the model and also for estimating the quality of that model -- 
in particular, the thirteen ``EP''-models were trained and tested on the
$D^{EP}$ set of records.}
% {See earlier comment - line 1107...}

Table~\ref{tab:datasets} (in Appendix~\ref{preprocessing_variants}) describes all our dataset variants.

% \add[RG]{The next sections will focus on only a few of these datasets... }
% Table~\ref{tab:datasets} (in Appendix~\ref{preprocessing_variants})
% succinctly describes these 26 datasets,
% showing which set of modifications is applied to each.
% That appendix also shows the results of applying the various learning models
% (Section~\ref{sec:ML-algs})
% to each of these dataset-variants.

\subsection{Machine Learning Algorithms} % 2.5
\label{sec:ML-algs}

\note[New]{Removed section on describing the standard ML framework and first subsection heading}
This work considers 
twelve different learners, based on % a range of possible 
{seven} base learning algorithms,
each run on each of the various different datasets
(differing based on the feature preprocessing used; 
see % that we have described previously, in 
Section~\ref{sec:FeatureEng}
% The standard learning algorithms that we use are: 
K-Nearest Neighbors (KNN), 
Support Vector Regression (SVR), 
Artificial Neural Networks (ANN), 
Wavelet Neural Networks (WNN), 
Ridge Regression (RR),
Random Forest Regression (RFR) 
and Gaussian Process Regression (GPR). 
We used patient\#16's first 3260 diabetes diary entries from dataset D21
to tune the hyper-parameters for the different base learners --
\eg for the GPR model (nugget = 0.25), KNN model (K = 10, weighting = uniform), RF model (maximum depth = 4), and our neural network model (batch size = 20, epochs = 1000). 
% Note that hyper-parameter tuning for the GPR model (nugget = 0.25), KNN model (K = 10, weighting = uniform), RF model (maximum depth = 4), and our neural network model (batch size = 20, epochs = 1000) was done using patient\#16's first 3260 diabetes diary entries from dataset D21. 
These 3260 records were then excluded from our testing data 
% in order 
to reduce the risk of % avoid
overfitting.
All other unspecified parameters were defaults
-- \eg
the linear SVR model (C = 1), SVR with RBF kernel model (C = 1, $\sigma^2 = \frac{1}{\hbox{\# of features}}$) and Ridge regression model ($\alpha$ = 1) 
are the defaults provided by
{\tt scikit-learn}~\cite{scikit-learn}. 
\comment{
The SVR model (C = 1), SVR with RBF kernel model (C = 1, $\sigma = \frac{1}{\hbox{\# of features}}$) and Ridge regression model ($\alpha$ = 1) were not tuned and are the defaults provided by scikit-learn~\cite{scikit-learn}. 
All other unspecified parameters were defaults. 
}
The ANN architecture included one output neuron with linear activation and two hidden layers of $3 \times (\hbox{\# of features})$
with rectified linear activation. 
The WNN architecture included one output neuron with linear activation and one hidden layer of $(\hbox{\# of features})$ neurons with 
Gaussian wavelet activations ($\Psi(x) = -xe^{-\frac{1}{2}x^2}$).
%Additionally,  Table~\ref{tab:knn} (in the Supporting Information) shows the results of exploring the KNN parameters on dataset D9. 
Most of these models were implemented with the help of {\tt scikit-learn}~\cite{scikit-learn}, 
except for the {\tt ANN} that was implemented using Keras~\cite{chollet2015keras} 
and the {\tt WNN} that was implemented in part with
{\tt scikit-neuralnetwork}\\
(\url{http://scikit-neuralnetwork.readthedocs.io/en/latest/index.html}).
We assume the reader is familiar with these fairly standard learners.

% In this work
We also combined base learners to develop more complex learners.
The following section (Section~\ref{weighted_gpr}) describes our GPR ensemble approach. 
%\change[RG]{To incorporate the information from the other patients ({eg} use patients \#1 to \#51 to help train a model for patient \#52), we use a ``Stacking'' approach where a model is trained on the auxiliary patients and then used to produce a new feature for the test patient by making predictions on the test's patients data.}
%\note[RG]{Should this be earlier -- as these are other types of features?}\note[NB]{I don't consider these new features}
We also considered another approach, which incorporates
 the information from the other patients
 -- \eg use patient histories \#1 to \#46 to help train a model for patient \#47.
%\note[RG]{What number should we use here? Also, if a patient has 2 entries ... }
 This ``Stacking'' approach first trains a model on the auxiliary patients, then runs this model on the test patient's data
 to produce a new feature for each meal -- \ie a 14th feature,
 to augment the 13 features shown in Table~\ref{tab:prodata}.
%\note[RG]{Need to use ref to table, and ie, eg... }
%-- that is, learn a model from patients \#1 to \#51, and use this, as well as information from patient\#52, to make prediction for patient\#52.
%Fig~\ref{fig:stacking} in 
Appendix~\ref{stacking} shows the entire stacking process.

\subsubsection{Modeling with a confidence weighted GPR Ensemble, $M_{gpr}^w$} % 2.7.2
\label{weighted_gpr}

%To make an accurate model that can predict a patient's blood glucose given their diabetes history, we constructed a model that is a combination of two models produced from  Gaussian Process Regression. 
%We describe Gaussian Process Regression in Section~\ref{gpr} and the way in which we combine our regression models in Section~\ref{weighted_gpr}.
%\subsubsection{Weighting by Model Confidence}
%\note[NB]{NEED TO FIND CITATION FOR THIS APPROACH}
For each patient, our ``GPR ensemble'' model first creates two different GPR models,
then combines them into a single model called $M^{w}_{gpr}$.
%  to predict that patient's blood glucose values. 
The first of these two models, $GPR_{p}$, learns from the entirety of a patient's training data. 
%\note[RG]{re-wrote next sentence; see comment below.}
The second model, $\{ GRP_{m} \}$, is actually a collection of GPR models 
-- one for each possible meal category $m$ (corresponding to \feat{meal}{i}\ in Table~\ref{tab:orig_features}.
Each of these $GPR_m$ models is trained using only the occurrences of that particular meal category 
in the patient's training data (\eg all occurrences of 
``Before Lunch'').
\comment{
The second model $\{ GRP_{m} \}$ is a collection of GPR models where for each possible meal category (See $meal_i$ in Table~\ref{tab:features})
% \change[RG]{($meal_i \in \{\ \hbox{before-breakfast},\ \hbox{after-breakfast},\ \dots\ \}$),}{ 
% (see $meal_i$ in Table~?   [which lists all of the features])
% }
a GPR model is trained using only the occurrences of that particular meal category 
in the patient's training data (\eg all occurrences of before-breakfast).
}
Once we have obtained $GPR_{p}$ and the set of $GRP_{m}$ models and wish to make a prediction for instance $x_i$,
we produce a weighted prediction of the form
\begin{equation}
\widehat{BG_{i+1}}\quad =\quad \frac{1}{\alpha + \beta}\left[ 
   \alpha\ GPR_{p}(\,[x_i, \dts{i+1}]\,)\ +\ \beta\ GPR_{m}(\,[x_i, \dts{i+1}]\,)\ \right]
\comment{\hat{y_i}\quad =\quad \frac{1}{\alpha + \beta}\left[ \alpha\ GPR_{p}(x_i)\ +\ \beta\ GRP_{m}(x_i)\ \right]}
\end{equation}
where $\alpha = \frac{1}{{\sigma_p}_i}$ and $\beta = \frac{1}{{\sigma_m}_i}$,
and where ${\sigma_p}_i$ and ${\sigma_m}_i$ are respectively
the standard deviations of the posterior Gaussian distributions
% \note[RG]{What is mean for N(...) -- should this be $[X, \Delta T]$? Or is it specific to $i$?}
$\mathcal{N}(\,GPR_{p}(\,[x_i, \dts{i+1}]\,), \,{\sigma^2_p}_i\,)$ and 
$\mathcal{N}(\,GPR_{m}(\,[x_i, \dts{i+1}]\,), \,{\sigma^2_m}_i\,)$ 
at the point $x_i$.

\subsection{Model Evaluation} % 2.7
\label{sec:ModelEval}

%\subsubsection{Evaluating Model Quality} % 2.7.1
%\label{sec:quality}

\add[New]{
To assess the performance of our models in general, evaluation functions are required to measure the quality of model predictions with respect to known true outcomes. 
In this paper we report our results in terms of ``$L_1$-loss'' ($\errL$), ``relative $L_1$-loss'' ($\errrL$) and Root Mean Squared Error (RMSE).}

Given a model $\Model{}( \cdot)$  
and a dataset
\begin{equation}
D % [\hbox{patient\#j}]
\quad =\quad \{\,[\, [x_i , \dts{i+1}],\ \mBG{i+1} \,] \,\}_i
\label{eqn:Dataset1}
\end{equation}
% $D = \{ x_i\}_i$ 
where each $x_i$ provides
the  ``temporal information'' shown in Equation~\ref{eqn:xi-info},
% (and in particular, the next time point $meal_{i+1}$ and associated blood glucose value $\BG{i+1}$),
% \note[RG]{Hmmm.. we should probably move the rL1 to the Sec 3.2 --
% ie, have the main text just deal with L1, ...}
\remove[New]{we considered both the} ``$L_1$-loss'' ($\errL$), ``relative $L_1$-loss'' ($\errrL$)\add[New]{, and RMSE are defined as}
\begin{eqnarray}
\err{L1}{\Model{}(\cdot), D}&=& 
\frac{1}{|D|-1}\sum_{i=1}^{|D|-1} |\ \Model{}(x_i, \,\dts{i+1}) - \mBG{i+1}\ | 
 \label{eqn:L1-loss}\\
 \err{rL1}{\Model{}(\cdot), D}&=& 
\frac{1}{|D|-1}\sum_{i=1}^{|D|-1} \frac{|\ \Model{}(x_i,\, \dts{i+1}) - \mBG{i+1}\ |}{\mBG{i+1}}
  \label{eqn:rL1-loss}\\
RMSE ({\Model{}(\cdot), D})&=& 
\sqrt{\frac{1}{|D|-1}\sum_{i=1}^{|D|-1} (\ \Model{}(x_i, \,\dts{i+1}) - \mBG{i+1}\ )^2}
  \label{eqn:rmse}
\end{eqnarray}
%\note[RG]{If you are going to use RMSE later -- should define. Perhaps here?}\note[NB]{Defined in following equation}
where \BG{i+1}\ is the blood glucose associated with the next time point, occurring  $\dts{i+1} %= \min\{\,\dt{i},\, \Delta t^{CHO}_{i}\,\}
$ minutes later \add[New]{(Note that \hbox{Del Favero~\etal~\cite{del2012glucose}} refers to $\errL$ and $\errrL$ as MAD and MARD, respectively)}.

\add[New]{For each of these three metrics we also consider ``glucose-specific'' variants. These ``glucose-specific'' variants use a ``Clark Error Grid inspired penalty function'' to re-weight the relative costs of different mispredictions (see \hbox{Del Favero~\etal~\cite{del2012glucose}} for descriptions and definitions). Therefore in total, we consider the performance of our models across 6 different evaluation functions (See Appendix\hbox{~\ref{sec:whyrl1}} for why we include both $\errL$ and $\errrL$).
}

\note[New]{Moved the Baseline model description here}

\subsubsection{Baseline: Using a Naive Predictor} % 2.6
% Establishing a Baseline: Comparison to a Naive Predictor
\label{sec:Naive}

%\change[RG]{To evaluate the quality of the models used in this work and to establish a baseline for comparison,}
{To establish a baseline for evaluating these learned models, 
}
we created a naive model ($\Model{avg}$) that,
for each patient, 
simply predicted that patient's 
average \BG{}\ value (over all meals/records) based on 
his/her % their 
diabetes history
-- that is, the naive model predicted the same average value (for that patient), 
independent of any other
information about that patient.
% (See Section~\ref{sec:Naive}).
%To evaluate the quality of the models used in this work, 
%we compared each to a naive baseline model ($\Model{avg}$):
%which is %. This baseline model was designed so that it always predicts 
%the average blood glucose value of a patient's history.%
% \footnote{
% This would be like a weatherman just predicting that the temperature tomorrow will be 3.6$^o$C every day -- independent of the season, or today's temperature, or any other climatic features~\url{http://www.edmonton.climatemps.com/temperatures.php}.
% }.
More concretely, given a patient's data $D = \{ [\dots, BG_i, \dots] \}_i$,
 partitioned into a training set $D^{train}$ and a test set $D^{test}$, 
the model calculates the average blood glucose for the entire training set 
% $D^{train}$
% \note[RG]{Use the notation ... Dtrain(X) -- to be explicit that this is wrt a specific person.... And have summation be $[BG_i, \dots] \in D^{train}$, or whatever}
\begin{equation}
BG_{avg}(D^{train})\quad =\quad 
\frac{1}{|D^{train}|} \sum_{[\dots, BG_i, \dots] \in D^{train}} BG_{i}
\end{equation}
including readings for all meals and all days.
Then for all instances $x_j$ in the associated test set $D^{test}$,
%\note[RG]{Need to decide about args to Model ... is it $\Delta t$? Or ...}\note[NB]{it should be $\Delta t$}
this model sets $\Model{avg}(\,x_j,\, \dts{j+1}\,)\ =\ BG_{avg}(D^{train})$.
So, for patient $16$:
as the \BG{avg}\ value for her first training set was 8.4,
this trivial model predicts that her blood glucose value will be 8.4 for each meal in the associated test set.
We can then evaluate this trivial model
using Equations~\ref{eqn:L1-loss}, ~\ref{eqn:rL1-loss} and ~\ref{eqn:rmse};
we clearly hope that the less-trivial models will do significantly better.

\comment{
Then, for each prediction made for the test set, the loss is calculated as 
$\mathcal{L}(\BG{avg},\ \BG{true})$,
which is either
$|\BG{avg}\,-\, \BG{true}|$ or $|\BG{avg}\,-\, \BG{true}|\,/\,\BG{true}$;
see Equations~\ref{eqn:L1-loss} and ~\ref{eqn:rL1-loss} respectively
}

\subsubsection{10-Fold Cross Validation}
\label{sec:10f-CV}

\begin{figure}[t] % [!ht]  3
\centering
\ShowFig{\includegraphics[width=4in%\textwidth
]{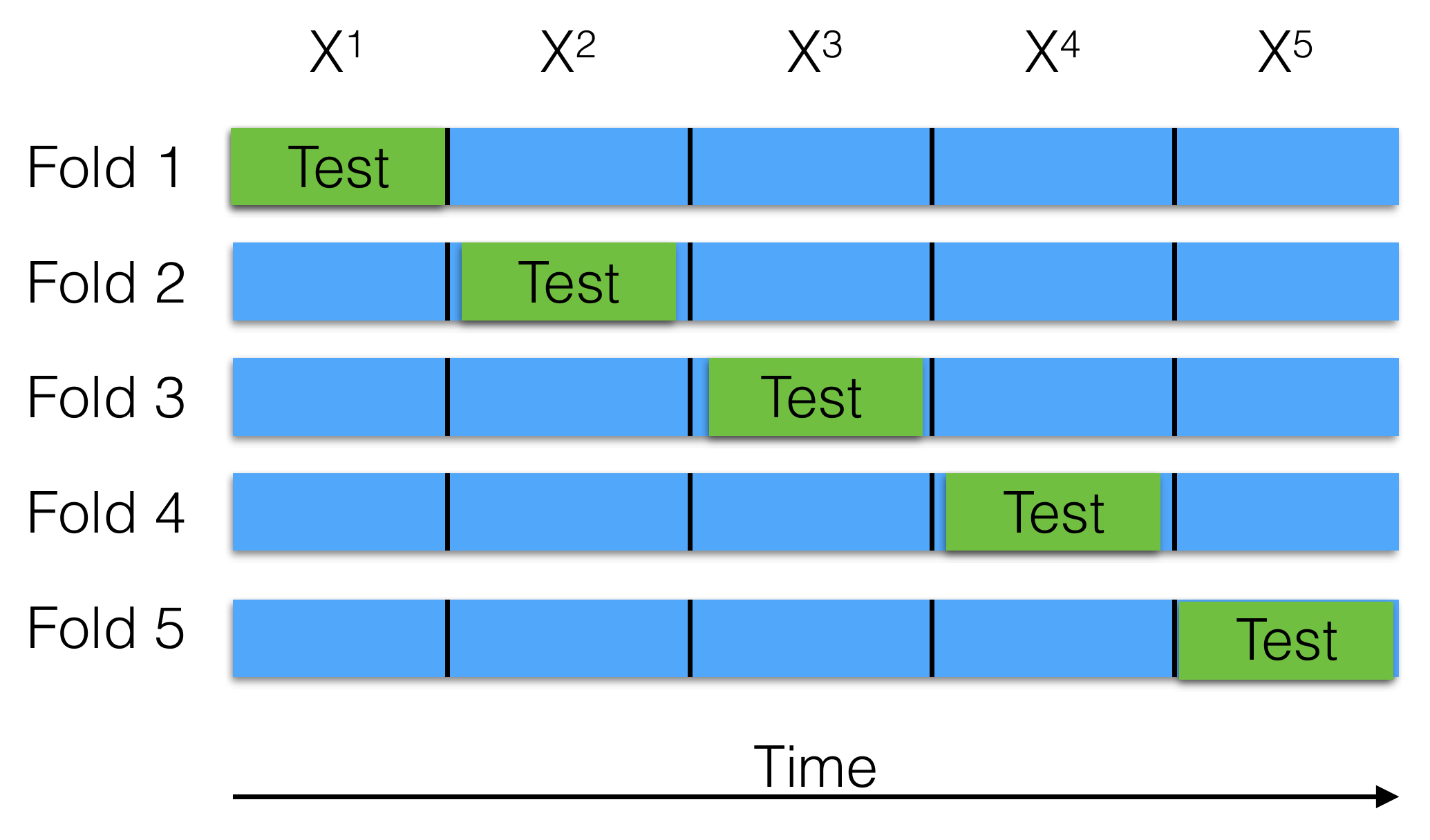}
}
\caption{{\bf Illustration of 5-Fold CV with Contiguous Segments.}
In each CV iteration training is done on the blue segments and testing on the green segment.} 
%\annote[RG]{Independence is assumed between segments.}{Independence? Perhaps conditioned on being same person?  Do we need to make this claim?}
\label{fig:contiguous_CV}
\end{figure}

Each of our learners will take the entire dataset and produce a model.
The next challenge is evaluating this learned model.
To evaluate the predictive quality of 
each learner, % our approach,
we use 10-fold cross validation (CV),
with respect to each patient. 
% Given the time series history of a patient, denoted $X_i$,
We first partition 
time series history of a patient, denoted $X_i$,
% the data
into ten contiguous segments
$X_i = \bigcup_{j=1..10} X_i^j$.
We then use % where
nine of the ten segments % are used 
for training 
% ($X^j_{i,tr}$)
in each CV round and 
use the remaining one segment % is used 
for testing % ($X^j_{i,te}$)
-- so the first split would be 
train $X^1_{i,tr} = \bigcup_{j=2..10} X_i^j$
and test $X^1_{i,te} = X_i^1$.
% Note that 
While the testing partition always consists of contiguous data, the training partition will not always be completely contiguous.
%% For example, if the data is partitioned such that $X = \{X_1, X_2, X_3, X_4, ...\}$ and $X_{test} = \{X_2\}$ is our testing partition, then our training partition, $X_{train} = \{X_1, X_3, X_4, ...\}$, will not be contiguous between $X_1$ and $X_3$.
Fig~\ref{fig:contiguous_CV} provides a visualization of what it means to partition time series data into contiguous segments for the purposes of cross validation -- for simplicity, here we show ``5-fold CV'' rather than 10.

% Results and Discussion can be combined.
\section{Results} % 3
\label{sec:results}

\subsection{Cross Validation Results}
\label{sec:CVresults}

% \note[RG]{Should start by quickly summarizing the 12 and the 26.
% Especially as earlier section mentioned 7 base learners, ...
% Should suggest what D0, D9, D21 ... -- ie the ones mentioned explicitly below.}

In this section we will discuss the results of our models when evaluated with cross validation on the dataset variants created. Again, 12 different learners were evaluated (derived from 7 base learners) and 26 different dataset variants (\change[New]{D0 through D25}{$D_e$1 through $D_e$13, and $D_a$1 through $D_a$13}) were created for the purpose of this analysis. 
For each of the 47 patient histories not used for hyperparameter selection 
%\annote[RG]{validation}{??? do you mean to set the hyperparameters?} 
(described in Section~\ref{Sec:data}),
we perform 10-fold CV using $12 \times 26$ 
different learner/dataset-variant combinations to determine their effectiveness
and how well they compare to the baseline model, $M_{ave}$ from Section~\ref{sec:Naive}.
Appendix~\ref{model_variants} provides details about the models,
and Appendix~\ref{app:heatmaps} provides heat-maps that
show both the performance of models on different datasets (in terms of $\errL$ and $\errrL$),
as well as the improved performance relative to $M_{ave}$. 
\add[New]{Appendix~\hbox{\ref{app:tables}}
also shows the performance of these learned models
in terms of other measures.}
\comment{
The model details and heat-map results can be found in Appendix~\ref{model_variants} and Appendix~\ref{app:heatmaps} respectively. 
These heat maps show both the performance of models on different datasets (in terms of $\errL$ and $\errrL$), as well as the improved performance relative to $M_{ave}$. 
}
% The heat-maps are organized so that 
% \note[RG]{This used to hold a description of the heat-maps ...
% I moved it to where it belongs: Appendix A.3}
\comment{
The left half of each heat map contains the datasets that adhere to our EP rules, while the right half contains those datasets that do not.
Models (on the y axis) are  sorted in terms of their average $\errL$ error (over all 26 datasets), so that the model with the best average $\errL$ error across all datasets appears at the top of these figures. 
Further, datasets D0 to D12 and D13 to D25 (in each half of these heat maps) are sorted horizontally,
in increasing order of average $\errL$ error across all models, with the left most dataset in each half having the smallest error. 
Note, that for Figure~\ref{AbsL1Te} 
the best model/dataset combination is found in the top left corner.
}

\begin{figure}[t] % !htbp]
\centering
\ShowFig{
\hspace*{-1in}\includegraphics[width=1.35\textwidth,height=3in]{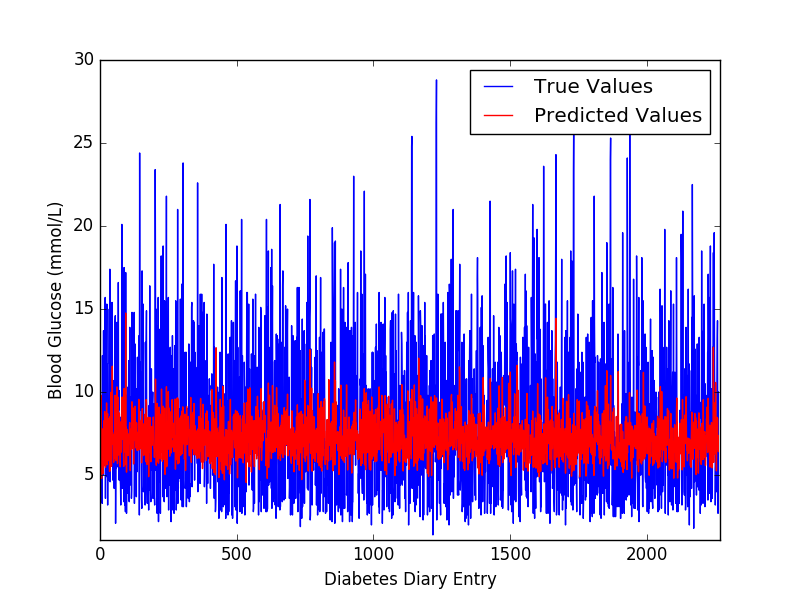}
}
\caption{{\bf Model $M^{w}_{gpr}$: Our GPR ensemble's predictions on data from patient 16.}}
\label{fig:118_gpr}
\end{figure}

For each learner and dataset variant pair,
we compute the $\errL$ error as a micro-average
over all the records of each patient.
These results allow us to determine % , we can identify 
the pair with the lowest average $\errL$, over all 47 patient histories:
These studies found that $M^{w}_{gpr}$ had the lowest $\errL$
on average across all of the 26 different preprocessing variants of the data.
On dataset \change[New]{D0}{$D_e$1} (the preprocessing variant with the lowest average $\errL$ across all models),
$M_{ave}$'s average $\errL$ was 2.91 mmol/L,
while $M^{w}_{gpr}$'s average $\errL$ was 2.70 mmol/L 
--
\ie our best model saw an improvement of only 7.1\% relative to the baseline!

To help understand why the improvement is not greater, 
Fig~\ref{fig:118_gpr} shows the predictions of $M^{w}_{gpr}$ 
for the processed entries from patient\#16 that were used for selecting hyperparameters. 
Here, we can see that the model is unable to account for 
the high amount of variance present in the \BG{}\ records for this patient.

\begin{figure}[t] % !htbp]
\centering
\ShowFig{
\includegraphics[width=\textwidth]{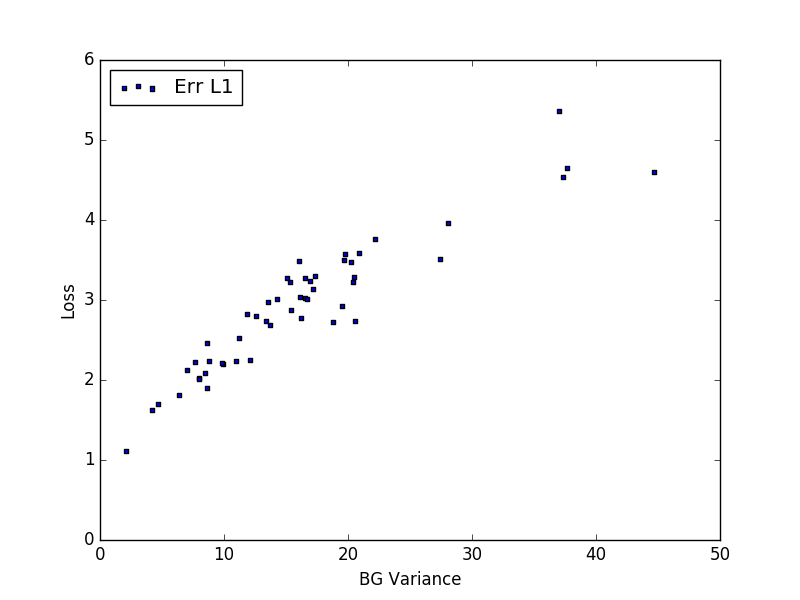}
}
\caption{{\bf Model $M^{w}_{gpr}$: average $\errL$ as a function of BG variance, for all patient histories.}}
\label{fig:bg_var}
\end{figure}

Fig~\ref{fig:bg_var} plots the variance in each patient's \BG{}\ history and the corresponding patient's $\errL$ loss 
(Equations~\ref{eqn:L1-loss})
that $M^{w}_{gpr}$ was able to achieve.
This figure shows
that the variance of a patient's blood glucose was highly correlated with 
the $\errL$ test loss ($0.93$ Pearson Correlation). 

\begin{figure}[tb] % !htbp]
\centering
\ShowFig{
\includegraphics[width=\textwidth]{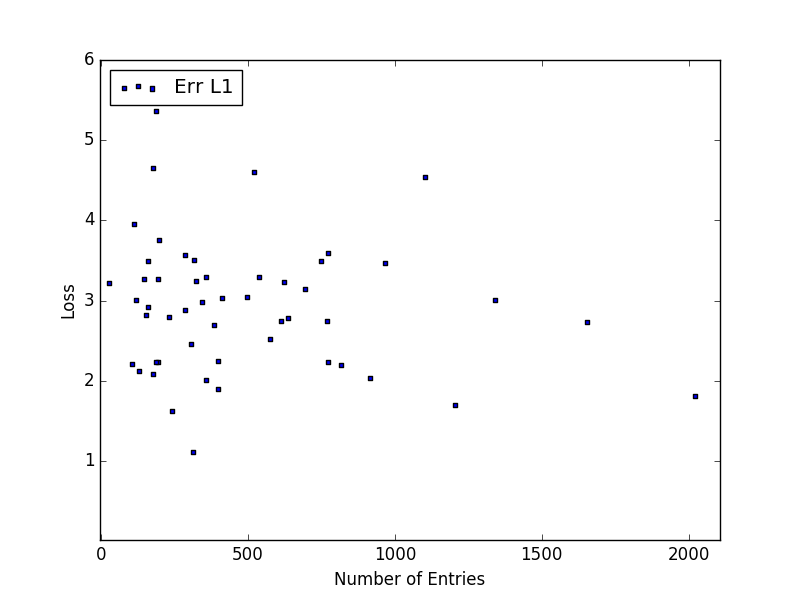}
}
\caption{{\bf Model $M^{w}_{gpr}$: average $\errL$ as a function of the \# of diabetes diary entries for a patient, for all patient histories.}}
\label{fig:entries_scatterplot}
\end{figure}

Fig~\ref{fig:entries_scatterplot} is a scatter plot of $\errL$ loss as a function of the number of data points that were available for each patient in the dataset. 
% Fig~\ref{fig:entries_scatterplot} shows that there seems to be
This figure suggests that there is no relationship between how well the model performs (in terms of $\errL$ test loss) on any particular patient,
and how many data points were collected from that patient -- 
Pearson Correlation: -$0.14$.

\subsection{Other Evaluation Measures} % 3.2
\label{sec:OtherEval}
\note[New]{This section heading is new, summarizing the various other evaluation measure.}
%\note[RG]{Here, we should first define the various measures,}
%\note[NB]{Since this is part of the results I think the definitions might fit better in Section 2.7.1 }
We then considered the $\errrL$ loss and found that
$M^{w}_{gpr}$'s $\errrL$ loss on \change[New]{D0}{$D_e$1} was 0.361.
We also saw that $M^{w}_{gpr}$ achieved the best $\errrL$,
although this was on a different dataset-variant.
The best [model, dataset-variant] pair was $M^{w}_{gpr}$ on dataset \change[New]{D9}{$D_e$11}, which achieved an average $\errrL$ error of 0.348; this was an improvement of 19.0\% relative to the $M_{avg}$ baseline of 0.430.
Note that $M^{w}_{gpr}$ achieved an $\errL$ of 2.78 mmol/L dataset \change[New]{D9}{$D_e$11}.

\note[New]{New Paragraph}
In addition to our $\errL$ and $\errrL$ metrics we also report RMSE and the glucose-specific versions of these metrics in Table~\ref{table:metrics}. Interesting the best model across all metrics was $M^{w}_{gpr}$, and the best dataset preprocessing variant was either $D_e$1 or $D_e$11 (both $D_e$ variants).

\def\spc{\hspace*{3em}}
% \note[RG]{Need to have another table to the ALL datasets $D_a$
% -- not just the Expert ones $D_e$ ...
% }
\begin{table}[!htbp] %[!ht]
\begin{adjustwidth}{-2.25in}{0in}
\centering
\caption{
{\bf Performance of the best models across all metrics on $D_e$ Datasets.}}
\begin{tabular}{|>{\columncolor{LightCyan}}l||c|c|r|c|c|}
\hline
\rowcolor[gray]{0.9}
Metric & Naive Model Error & Best Model Error & 
{Percent Improvement} & Best Model & Best Dataset Variant \\ \hline
RMSE & 3.58 & 3.47 &  2.98\%\spc & $M^{w}_{gpr}$ & $D_e$1\ \\ \hline
$\errL$ (MAD) & 2.91 & 2.70 & 7.12\%\spc & $M^{w}_{gpr}$ & $D_e$1\ \\ \hline
$\errrL$ (MARD) & 0.430 & 0.348 & 18.97\%\spc & $M^{w}_{gpr}$ & $D_e$11 \\ \hline
gRMSE & 5.01 & 4.86 & \ 2.95\%\spc & $M^{w}_{gpr}$ & $D_e$1\ \\ \hline
gMAD & 5.31 & 4.98 &\  6.28\%\spc & $M^{w}_{gpr}$ & $D_e$1\ \\ \hline
gMARD & 0.783 & 0.648 & 17.28\%\spc & $M^{w}_{gpr}$ & $D_e$11 \\ \hline
\end{tabular}
\vspace{.1cm}
\begin{flushleft} Note that these error values are micro-averages. These are selected entries from the tables in Supporting Information Section~\ref{app:tables}.
\end{flushleft}
\label{table:metrics}
\end{adjustwidth}
\end{table}

\begin{table}[!htbp] %[!ht]
\begin{adjustwidth}{-2.25in}{0in}
\centering
\caption{
{\bf Performance of the best models across all metrics on $D_a$ Datasets.}}
\begin{tabular}{|>{\columncolor{LightCyan}}l||c|c|r|c|c|}
\hline
\rowcolor[gray]{0.9}
Metric & Naive Model Error & Best Model Error & Percent Improvement & Best Model & Best Dataset Variant \\ \hline
RMSE & 3.67 & 3.58 & 2.49\%\spc & $M^{w}_{gpr}$ & $D_a$1\ \\ \hline
$\errL$ (MAD) & 2.96 & 2.78 & 6.14\%\spc & $M^{w}_{gpr}$ & $D_a$1\ \\ \hline
$\errrL$ (MARD) & 0.422 & 0.355 & 15.88\%\spc & $M^{w}_{gpr}$ & $D_a$11 \\ \hline
gRMSE & 5.16 & 5.02 & 2.70\%\spc & $M^{w}_{gpr}$ & $D_a$1\ \\ \hline
gMAD & 5.45 & 5.14 & 5.70\%\spc & $M^{w}_{gpr}$ & $D_a$1\ \\ \hline
gMARD & 0.772 & 0.660 & 14.50\%\spc & $M^{w}_{gpr}$ & $D_a$11 \\ \hline
\end{tabular}
\vspace{.1cm}
\begin{flushleft} Note that these error values are micro-averages. These are selected entries from the tables in Supporting Information Section~\ref{app:tables}.
\end{flushleft}
\label{table:metrics}
\end{adjustwidth}
\end{table}

%\note[RG]{fig:118\_grp used to be here.. Why?}
\comment{
\begin{figure}[t] % !htbp]
\centering
\hspace*{-1in}\includegraphics[width=1.35\textwidth,height=3in]{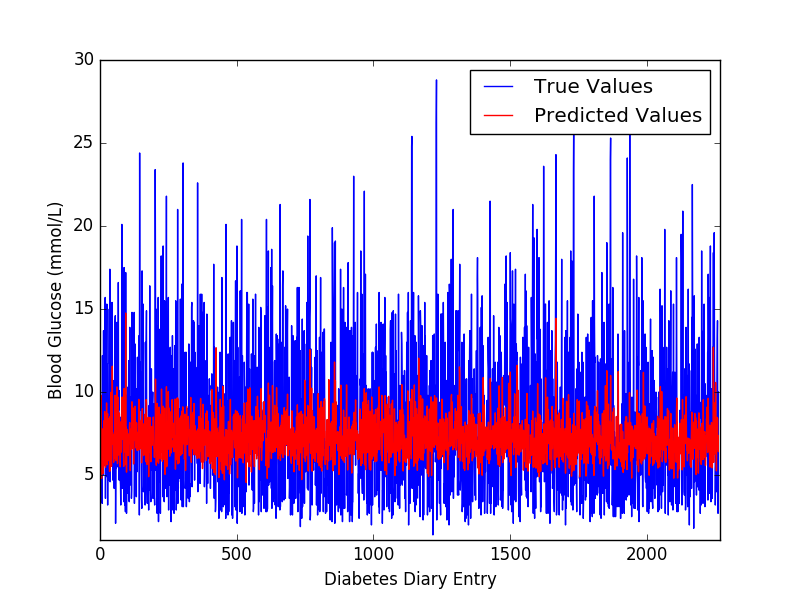}
\caption{{\bf Model $M^{w}_{gpr}$: Our GPR ensemble's predictions on data from patient 16.}}
\label{fig:118_gpr}
\end{figure}
}

\section{Discussion} % 4
\label{sec:discussion}

Our results show that our best learning algorithm is more accurate than a naive baseline -- but only slightly -- and that it can only achieve an average $\errL$-loss of approximately 2.70 mmol/L.
This loss means that, on average, if the patient's blood glucose was normal
(\eg 6 mmol/L), 
the learned model may incorrectly identify the patient as either hypoglycemic 
(as $6 - 2.70 < 4$ mmol/L)
or hyperglycemic ($6 + 2.70 > 8$ mmol/L).
Together with the strong relationship between glucose variance and prediction error, 
this highlights how challenging it is to create models that produce fine-grained blood glucose predictions when only using diabetes diary entries 
-- \ie using only the information that is commonly available to medical practitioners. 

Having tried 312 different combinations of learners and dataset variants, and observing minimal differences in their performance, 
it seems unlikely 
% \change[RG]{(although not impossible) that further performance gains can be achieved
% without overfitting new models to the data. 
% Of course, avoiding the overfitting of models to data is necessary when one wishes to make generalized claims about model performance.}
{that another [learner, dataset-variant] combination would be better.}
% \change[RG]{One noteworthy finding of these results is that efforts to combine patient data to build enhanced predictors were unsuccessful. 
% This is interesting because it indicates that simply including more patients in the study is not likely to improve model performance.}
{Note that these limited results include models that use data from multiple patients,
which suggest that simply including more patients in the study is not likely to improve model performance.}
% \change[RG]{Moreover, since there does not seem to be a strong relationship between the number of 
% data points
% that a patient has recorded and the performance of a model on that patient,
% collecting more of this type of data (sampled before and after meals) for each individual likely will not improve model performance.}
{ 
Moreover, since the model accuracy did not seem to improve as 
the number of 
recorded entries increased,
we suspect that simply collecting more of these entries
for each individual will not improve model performance.
}

There are many possible reasons why modeling T1D glucose levels based on 
this standard type of
diabetes diary data is so challenging. 
% \change[RG]{For our study, 
% % this may be because % 
% one reason may just be that some}
{It is possibly just an artifact of our study:
perhaps many
}
 of the patients who volunteered,
did so because their diabetes was difficult to manage.
Another reason could be that inaccuracies and omissions of variables in data prevent the model from producing accurate predictions. 
These omissions
could possibly include: 
not knowing the site where the bolus insulin was injected, 
how much scar tissue was present at the injection site, 
skin temperature, how accurately the carbohydrate value was recorded, the accuracy of the recorded insulin dose, 
the levels of different hormones, 
whether % or not 
the patient was menstruating, stress levels, accuracy of recording exertion, insulin age or storage conditions, amount of blood flow at the injection site 
and likely yet other factors.
Given our belief that training more accurate models will require additional relevant variables, future research might incorporate more confounding variables, such as injection location~\cite{koivisto1980alterations}, glucagon levels~\cite{unger2012glucagonocentric} and/or meal protein/fat content~\cite{paterson2017increasing}. However, it is not clear which, if any, of such variables are sufficient to explain the response, nor whether they can be practically captured in a clinical setting.
%As discussed in the Introduction, other systems obtain smaller error by making short term forecasts -- on the order of tens of minutes, instead of hours. While it may be the case that this is necessary to obtain accurate BG predictions, this is a more reactive than proactive approach to treating diabetes.

\section{Conclusion} % 5
\label{sec:concl}
%\change[RG]{In this work we investigated the feasibility of using machine learning algorithms to build an accurate prediction model for the future blood glucose values in Type I diabetes patients. }
This work explored the challenge of 
accurately predicting future blood glucose values 
in Type I diabetes patients, 
based on 
% \change[RG]{a model learned using}
{various models learning using several} machine learning algorithms.
Our extensive explorations 
-- involving 12 different learning algorithms, and 26 different encodings of the data (312 combinations) 
--
found that, on average, 
the model with the lowest expected $\errL$ was a confidence weighted Gaussian process regression model ($M^{w}_{gpr}$).
Using 10-fold cross validation on 29\,601 blood glucose records from
47 different patients,
our $M^{w}_{gpr}$ model 
%that % which
performed only $7.1\%$ better than 
the na{\"\i}ve ``mean predicting'' model ($M_{avg}$). 
Anecdotally,
a diabetologist also attempted to do this task --
predicting the BG for the next meal.
We found that this 
{model's predictions} (insignificantly) outperformed 
the diabetologist's, % an expert diabetologist
{in terms of a simple unbiased loss function},
but that the diabetologist performed (insignificantly) better when the evaluation was biased toward predicting hypoglycemic events;
see Borle~\cite{borle2017challenge}.

These results showed that our model could achieve an expected absolute error of $2.70$ mmol/L \add[New]{(48.65 mg/dl)}, which is disconcertingly large given that this is based on the type of data that is frequently collected and used for clinical practice (records are collected at meal times by the patients themselves). 
These results strongly suggest that the standard data collected by T1D patients, while apparently appropriate for clinical treatment of T1D, is not sufficient for accurately predicting blood glucose levels. 
We conjecture that using patient data that is sampled more frequently and that includes additional features would improve both the ability of professionals and machine learning practitioners to more accurately predict patient's blood glucose levels, but there is a practical trade-off between patient convenience and highly detailed record keeping.

\newpage

\appendix
\section{Supporting Information}
\label{supporting_information}

\subsection{Dataset Preprocessing Variants and Example}
\label{preprocessing_variants}

%\note[RG]{Someplace ... should motivate and describe the $\Delta t_i^{bolus}$... What would happen if ..}

As mentioned earlier, the features that we described in Section~\ref{sec:preproc} are simplified so that they can be easily understood. 
This appendix describes the complete set of features
% (shown in Table~\ref{tab:features}), 
that we 
used.
% created after imputing missing values and removing unusable records.

The previously described $\Delta t_i$ and \BG{i}\ features
implicitly assume that the previous event involved both
injecting a quantity of insulin and also consuming some carbohydrates.
While this is typically the case, 
there are exceptions.
To accommodate such situations,
where an event involves only one of these,
we decompose each of these terms
into two different features:
$\Delta t_i$ is separated into $\Delta t^{bolus}_i$ and 
$\Delta t^{CHO}_i$,
and \BG{i}\ is separated into
\BG{i}$^{bolus}$ and \BG{i}$^{CHO}$. 
These correspond to the time since most recent (previous) bolus injection $\Delta t^{bolus}_i$ (when \ebolus{i}\ bolus units were injected and the blood glucose was \BG{i}$^{bolus}$),
and the time since the most recent previous carbohydrate consumption
$\Delta t^{CHO}_i$ 
(when the subject consumed  \eCHO{i}\ and 
his/her blood glucose was \BG{i}$^{CHO}$).

Table~\ref{tab:features} 
provides the complete set of features that we used and 
Table~\ref{tab:prodata}
provides the correct % an updated
version of Patient\#16's data from Table~\ref{tab:rawdata} with these features.
%\annote[RG]{This distinction is made because while many events correspond to both injecting a quantity of insulin and also consuming some carbohydrates, this is not always true.}{ Should START by explaining why you are doing this.  So I moved this above... delete this sentence.}
%\note[RG]{Need to finish this, be describing what you do with these terms: Hence, our feature set is instead the one shown in Table ??. We also show the extended version of Patient\#16's data in Table ???.}

\begin{table}[t]
\begin{adjustwidth}{-2.25in}{0in} % Comment out/remove adjustwidth environment if table fits in text column.
\centering
\caption{
{\bf Description of Original and Processed Features used in this Study}
\label{tab:features}
}
% \note[RG]{Why not have the order of features here, match text earlier?\\ Should first have "basic values", then computed values.}
\small
\begin{tabular}{|>{\columncolor{LightCyan}}l|p{12cm}|}
\hline
\feat{meal}{i} & The time of day:
\{ Before Breakfast, After Breakfast, Before lunch, After Lunch, Before Supper, After Supper, Before Bed, During the Night\}\\ \hline
\feat{DOW}{i} & The day of the week\\ \hline
\feat{xV}{i} & Numeric encoding of exercise value: $\{2, 4, 7, 10\}$ \\ \hline
\feat{PV}{i} & Pump Value: The rate at which the insulin pump is infusing ($\frac{units}{hour}$). This is always 0 if the patient does not have a pump  \\ \hline
\feat{basal}{i} & The units of background insulin injected\\ \hline
\feat{BG}{i} & The BG value at the current time ($\frac{mmol}{L}$) \\ \hline
\hline
\feat{IOB}{i} & Insulin on Board: Estimated residual insulin from the previous injection ($\frac{mmol}{L}$)  \\ \hline
\eCHO{i} & The previous most recent amount of carbohydrates ingested (grams)\\ \hline
\ebolus{i} & The previous most recent  amount of insulin injected (units)\\ \hline
\BG{i-}$^{CHO}$ & The BG value at the time that \eCHO{i}\ was ingested ($\frac{mmol}{L}$)\\ \hline
\BG{i-}$^{bolus}$ & The BG value at the time that \ebolus{i}\ was injected ($\frac{mmol}{L}$)\\ \hline
$\Delta t^{CHO}_i$& The time between \eCHO{i}\ and \BG{i+1}\ (min)  \\ \hline
$\Delta t^{bolus}_i$ & The time between \ebolus{i}\mbox and \BG{i+1}\ (min)\\ \hline
% $BG_{i+1}$ & The BG value at the next time step to be predicted ($\frac{mmol}{L}$) \\ \hline
\end{tabular}
\vspace{0.05cm}
\end{adjustwidth}
\end{table}

\begin{table} % [t] % [!ht]
\begin{adjustwidth}{-2.25in}{0in} % Comment out/remove adjustwidth environment if table fits in text column.
\centering
\caption{
{\bf Example of Processed Data, over a single day, from Patient 16 (Variant D1)}}
\label{tab:prodata}
\small
\begin{tabular}{|g|c|c|c|c|c|c|}
\hline
index $i$ & 27 & 28 & 29 & 30 & 31 & 32\\ 
\hline
\feat{meal}{i} & Before Breakfast & After Breakfast & Before Lunch & After Lunch & Before Dinner & After Dinner \\ %\hline
\feat{DOW}{i} & Tuesday & Tuesday & Tuesday & Tuesday & Tuesday & Tuesday \\ %\hline
\feat{EV}{i} & 4 \normal& 4 \normal & 4 \normal & 4 \normal & 4 \normal & 4 \normal\\ %\hline
\feat{PV}{i}  & 0.50 & 0.50 & 0.63 & 0.45 & 0.90 & 0.90\\ %\hline
\feat{basal}{i} & 0 & 0 & 0 & 0 & 0 & 0 \\ 
\BG{i} & 16.2 & 14.7 & 5.6 & 6.8 & 10.5 & 3.0\\
\hline
\feat{IOB}{i} & 0.00 & 7.90 & 3.61 & 0.89 & 0.81 & 3.35\\ %\hline
\eCHO{i} & 17.5 & 30.0 & 20.5 & 30.0 & 18.5 & 15.0\\ %\hline
\ebolus{i} & 1.93 & 10.40 & 2.44 & 3.00 & 2.54 & 3.80\\ %\hline
\BG{i-}$^{CHO}$ & 10.3 & 16.2 & 14.7 & 5.6 & 6.8 & 10.5\\ %\hline
\BG{i-}$^{bolus}$ & 10.3 & 16.2 & 14.7 & 5.6 & 6.8 & 10.5\\ %\hline
$\Delta t^{CHO}_i$ & 540 & 103 & 120 & 196 & 187 & 89\\ %\hline
$\Delta t^{bolus}_i$ & 540 & 103 & 120 & 196 & 187 & 89\\ %\hline
% $BG_{i+1}$ & 16.2 & 14.7 & 5.6 & 6.8 & 10.5 & 3.0\\
\hline
\end{tabular}
\vspace{0.05cm}
\begin{flushleft} 
Note that 
$\mBG{i}^{CHO}$ will differ from $\mBG{i}^{bolus}$ and $\Delta t^{CHO}_i$ will differ from $\Delta t^{bolus}_i$%
{whenever} carbohydrates and insulin were not taken at the same time. 
% Data from 118 2705:2711
%\note[RG]{?? I am confused by the notation here. Isn't every column based on information taken at a single time??? Let's talk about this.}\note[NB]{I've remade this table entirely, please check it over again}\\ \note[NB]{Also, I think I've figured out the notational confusion. $\BG{i+1}$ is a label, not a feature. we don't use $\BG{i}$ to predict $\BG{i+1}$, we instead use the most recent $\Delta t^{CHO}_i$ or $\Delta t^{bolus}_i$ so given the most recent record $ * = argmin(\dt{i}, \Delta t^{CHO}_i)$ we are predicting $\Delta t^{*}_i$ min ahead starting with $BG^{*}_{i}$. Finally, it is not necessarily the case that $\BG{i+1} = BG^{bolus}_{i+1}$ or $\BG{i+1} = BG^{CHO}_{i+1}$}
\end{flushleft}
\end{adjustwidth}
\end{table}

Table~\ref{tab:datasets} lists all of the dataset variants that are generated by 
the various % of differing 
feature-creation/selection % preprocessing
steps. Note the first 13 datasets 
(\change[New]{D0 - D12}{$D_e$1 - $D_e$13}) contain only the subset of records that satisfy the EP criteria -- see Section~\ref{sec:EP}.
The second 13 rows (\change[New]{D13 - D25}{$D_a$1 - $D_a$13}) deal with, essentially, the entire set of records. 
The ``Missing Carbs'' and ``Missing Bolus'' columns address variants in handling missing data, ``PCA Transformation'' indicates if the features were transformed using PCA, the ``Kok features'' column indicates datasets that use the features described by Peter Kok and the remaining columns indicate 
whether specific other features are included.
% the presence of additional features that are included.

%\note[RG]{Why bars around Missing Bolus, but not others? Should have bars separating \#Subjects from rest.\\ Should also have \#Records -- especially relevant wrt EP instances}\note[NB]{This is actually a bug in Latex which I have not been able to find a solution to so far}

\begin{table}[!htbp] %[!ht]
\begin{adjustwidth}{-2.25in}{0in}
\centering
\caption{
{\bf Datasets Generated from Different Feature Sets.}}
\note[New]{On this table, and below:
renamed the datasets to reflect new naming conventions,
and reordered datasets, based on updated results.}
\resizebox{1.4\textwidth}{!}{
\begin{tabular}{|>{\columncolor{LightCyan}}l|| >{\centering\arraybackslash}m{1.6cm} | >{\centering\arraybackslash}m{1.6cm} | >{\centering\arraybackslash}m{1.7cm} | >{\centering\arraybackslash}m{1.7cm} | >{\centering\arraybackslash}m{1.7cm} | >{\centering\arraybackslash}m{1.7cm} | >{\centering\arraybackslash}m{1.7cm} | >{\centering\arraybackslash}m{1.7cm} | >{\centering\arraybackslash}m{2.5cm} | >{\centering\arraybackslash}m{2.5cm} |}
\hline
\rowcolor[gray]{0.9}
  &
  \parbox[t]{1.3cm}{\# of \\ Subjects} & 
  \parbox[t]{1.5cm}{Records \\ Predicted} & 
  \parbox[t]{1.0cm}{EP \\ Rules} & 
  \parbox[t]{1.3cm}{DOW \\ Features} &
  \parbox[t]{1.0cm}{Basal \\ Feature} & 
  \parbox[t]{1.2cm}{Patient\\ Specific \\ Features} & 
  \parbox[t]{1.3cm}{Kok \\ Features} & 
  \parbox[t]{1.5cm}{PCA \\ Transform} & 
  \parbox[t]{1.4cm}{Missing \\ Carbs} & 
  \parbox[t]{1.4cm}{Missing \\ Bolus}\\
  \hline
$D_e$1  & 42 & 7378  & 1 & 1 & 1 & 0 & 0 & 0 & Throwout & Impute Mean\\ \hline
$D_e$2  & 39 & 5978  & 1 & 1 & 1 & 0 & 0 & 0 & Throwout & Throwout\\ \hline
$D_e$3  & 47 & 16167 & 1 & 1 & 1 & 0 & 0 & 0 & Impute Mean & Impute 0\\ \hline
$D_e$4  & 47 & 16167 & 1 & 1 & 1 & 0 & 0 & 0 & Impute 0 & Impute Mean\\ \hline
$D_e$5  & 47 & 16167 & 1 & 7 & 1 & 1 & 0 & 0 & Impute Mean & Impute Mean\\ \hline 
$D_e$6  & 47 & 16167 & 1 & 1 & 1 & 0 & 0 & 0 & Impute Mean & Impute Mean\\ \hline
$D_e$7  & 47 & 16167 & 1 & 7 & 1 & 0 & 0 & 0 & Impute Mean & Impute Mean\\ \hline
$D_e$8  & 47 & 16167 & 1 & 0 & 0 & 0 & 0 & 0 & Impute Mean & Impute Mean\\ \hline
$D_e$9 & 37 & 7349  & 1 & 0 & 0 & 0 & 1 & 0 & N/A & N/A\\ \hline
$D_e$10  & 47 & 16167 & 1 & 1 & 1 & 0 & 0 & 0 & Impute 0 & Impute 0\\ \hline 
$D_e$11  & 45 & 10623 & 1 & 1 & 1 & 0 & 0 & 0 & Impute Mean & Throwout\\ \hline
$D_e$12 & 47 & 16167 & 1 & 0 & 0 & 0 & 0 & 1 & Impute Mean & Impute Mean\\ \hline
$D_e$13 & 37 & 7349  & 1 & 0 & 0 & 0 & 1 & 1 & N/A & N/A\\ \hline
\hline                             
$D_a$1 & 42 & 15961 & 0 & 1 & 1 & 0 & 0 & 0 & Throwout & Impute Mean\\ \hline
$D_a$2 & 39 & 13832 & 0 & 1 & 1 & 0 & 0 & 0 & Throwout & Throwout\\ \hline
$D_a$3 & 47 & 24896 & 0 & 1 & 1 & 0 & 0 & 0 & Impute Mean & Impute 0\\ \hline
$D_a$4 & 47 & 24896 & 0 & 1 & 1 & 0 & 0 & 0 & Impute 0 & Impute Mean\\ \hline
$D_a$5 & 47 & 24896 & 0 & 7 & 1 & 1 & 0 & 0 & Impute Mean & Impute Mean\\ \hline
$D_a$6 & 47 & 24896 & 0 & 1 & 1 & 0 & 0 & 0 & Impute Mean & Impute Mean\\ \hline  
$D_a$7 & 47 & 24896 & 0 & 7 & 1 & 0 & 0 & 0 & Impute Mean & Impute Mean\\ \hline
$D_a$8 & 47 & 24896 & 0 & 0 & 0 & 0 & 0 & 0 & Impute Mean & Impute Mean\\ \hline
$D_a$9 & 37 & 11888 & 0 & 0 & 0 & 0 & 1 & 0 & N/A & N/A\\ \hline
$D_a$10 & 47 & 24896 & 0 & 1 & 1 & 0 & 0 & 0 & Impute 0 & Impute 0\\ \hline 
$D_a$11 & 45 & 19122 & 0 & 1 & 1 & 0 & 0 & 0 & Impute Mean & Throwout\\ \hline
$D_a$12 & 47 & 24896 & 0 & 0 & 0 & 0 & 0 & 1 & Impute Mean & Impute Mean\\ \hline
$D_a$13 & 37 & 11888 & 0 & 0 & 0 & 0 & 1 & 1 & N/A & N/A\\ \hline
\end{tabular}}
\begin{flushleft} Here, ``Basal feature'' and ``Patient Specific Features'' are features that were included (1) or excluded (0) from datasets. ``DOW Features'' indicates if the day of the week was not included (0), included (1), or included as a one hot encoded feature (7). 
``PCA Transform'' indicates whether the data was reduced to 4 principle components. 
``Kok Features'' means that the data was preprocessed to replicate (as best as possible) the features used in Kok's MSc thesis~\cite{kok2004predicting}. 
In the final two columns, the value ``Throwout'' means that these records were removed from the dataset.
%\note[RG]{Is it Subjects or Histories?  Also is max still 51, or just 49?}
The ``\# of Subjects'' column shows that some datasets did not include all (47) patients. In these cases, patients were excluded because they had too few records (under 100) after the preprocessing steps were applied to their data. 
The table is partitioned so that datasets with the EP rules (\change[New]{D0 -- D12}{$D_e$1 -- $D_e$13}) precede their corresponding datasets without EP rules (\change[New]{D13 -- D25}{$D_a$1 -- $D_a$13}).
\end{flushleft}
\label{tab:datasets}
\end{adjustwidth}
\end{table}

\clearpage

\subsection{Model Variants}
\label{model_variants}

\begin{table}[!htbp] %[!ht]
\begin{adjustwidth}{-2.25in}{0in}
\centering
\caption{
{\bf Descriptions of the Different Learners Used.}}
\begin{tabular}{|l|l|c|c|c|}
\hline
\rowcolor[gray]{0.9}
name & Symbol & Algorithm & Confidence Weighting & Stacking \\ \hline
gpr\_be & $M^{w}_{gpr}$ & GPR & 1 & 0 \\ \hline
gpr\_be\_AllPat\_AllMeals & $M^{ws}_{gpr}$ & GPR & 1 & 1\\ \hline
gpr\_IndPat\_AllMeals & $M_{gpr}$ & GPR & 0 & 0 \\ \hline
gpr\_AllPat\_AllMeals & $M^{s}_{gpr}$ & GPR & 0 & 1 \\ \hline
svr1 & $M_{svr}$ & SVR (RBF Kernel) & 0 & 0 \\ \hline
svr1\_lin & $M^{lin}_{svr}$ & SVR (Linear Kernel) & 0 & 0 \\ \hline
svr1\_allpats & $M^{s}_{svr}$ & SVR (RBF Kernel) & 0 & 1 \\ \hline
rf4 & $M_{rf}$ & Random Forest & 0 & 0 \\ \hline
KNN10U & $M_{knn}$ & KNN & 0 & 0 \\ \hline
ridge & $M_{ridge}$ & Ridge Regression & 0 & 0 \\ \hline
wnn & $M_{wnn}$ & Wavelet Neural Network & 0 & 0 \\ \hline
NN & $M_{nn}$ & Feed-Forward NN & 0 & 0 \\ \hline
naive & $M_{avg}$ & BG History Average & 0 & 0 \\ \hline
\end{tabular}
\begin{flushleft} Confidence Weighting is explained in Section~\ref{weighted_gpr} and Stacking is explained in Appendix~\ref{stacking}.
\end{flushleft}
\label{tab:models}
\end{adjustwidth}
\end{table}

% \clearpage

\subsection{Inclusion of $\errrL$ in addition to $\errL$}
\label{sec:whyrl1}

\note[New]{Content moved here from the section on evaluating our models}

\begin{figure}[t] % [!h]
\centering
\ShowFig{
\includegraphics[width=\textwidth,height=2in]{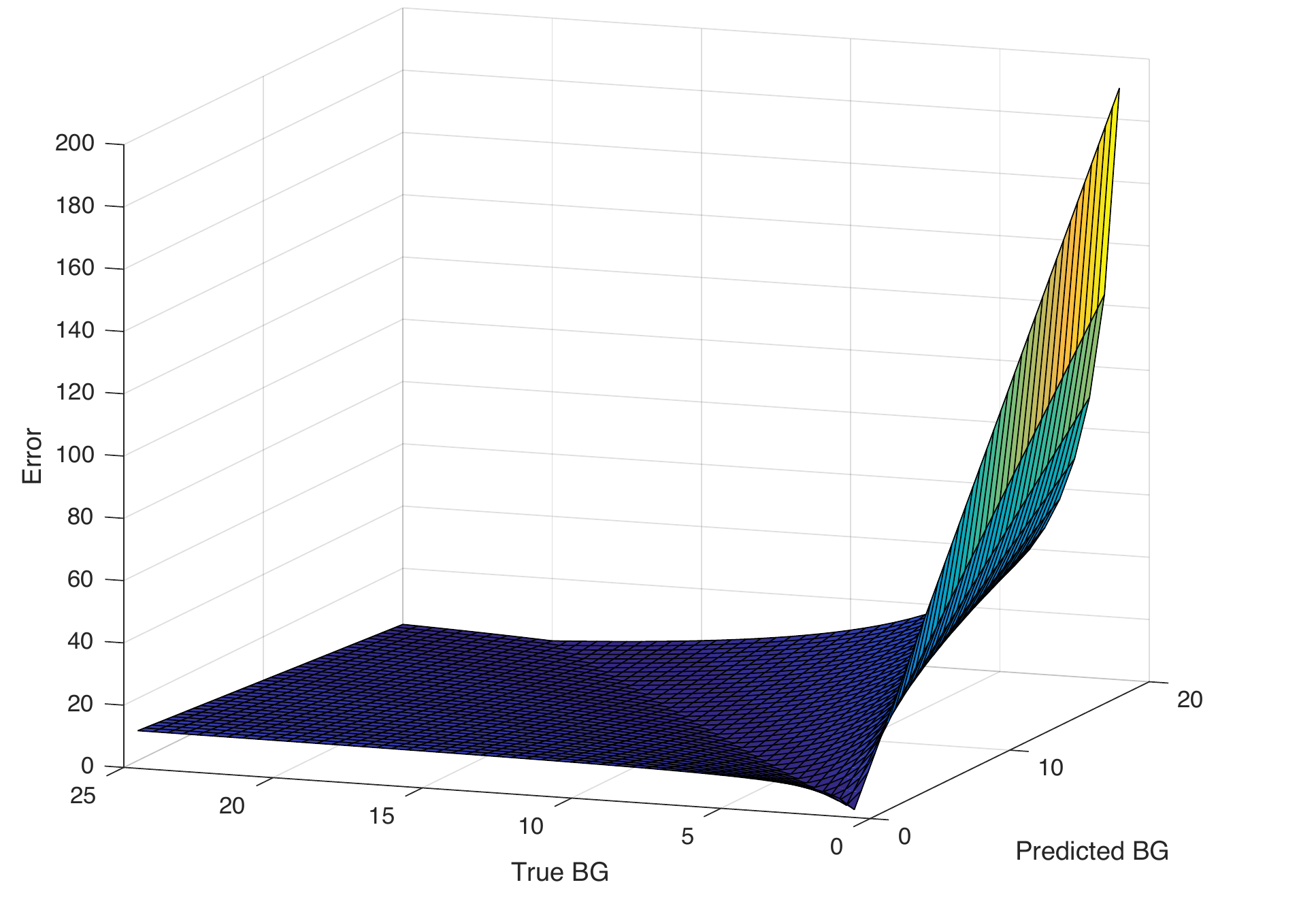}
}
\caption{{\bf Target $\errrL$ Loss Function.}
This function emphasizes loss when the model fails to predict hypoglycemic events. Here, ``True BG'' refers to 
\BG{i+1}\ and ``Predicted BG'' refers to 
$\widehat{\mBG{i+1}} %   \hbox{\BG{i+1}}
   \ =\ \Model{}(\, x_{i},\, \dts{i+1}\,)
   $. 
Note that the error is large when the true \BG{}\ is very low, but the predicted 
 $\widehat{\mBG{}} $  % \BG{}
    is relatively high.
  }
\label{fig:weightedL1_plot}
\end{figure}

While $\errL$ is the standard loss function,
% we show why 
it can be problematic: %  in the following example:
Note the [predicted, true] pair
$[ \Model{}(x_1,\, \dts{2}),\ \mBG{2}]\ =\ [5,\,3]$
and
$[ \Model{}(x_3, \,\dts{4}), \ \mBG{4}]\ =\ [10,\,12]$
each  have an $\errL$ of 2
-- \ie
$|\Model{}(x_1, \dts{2}) - \mBG{2}| =
2 = 
|\Model{}(x_3, \dts{4}) - \mBG{4}|$
--
but predicting 5 mmol/L instead of 3 mmol/L is potentially much more dangerous, in terms of patient health, than predicting 10 mmol/L instead of 12 mmol/L.
Here, the $\errrL$ function would correctly impose a larger penalty to the first
$\frac{|\Model{}(x_1, \dts{2}) - \mBG{2}|}{\mBG{2}} = \frac{2}{3}$
versus the second
$\frac{|\Model{}(x_3, \dts{4}) - \mBG{4}|}{ \mBG{4}} = \frac{2}{12}$.
See Fig~\ref{fig:weightedL1_plot} for a visualization of the $\errrL$ function,
showing that this error is especially large 
in the dangerous situation
when the true value \BG{}\ is small but the predicted value
$\Model{}(x, \dts{})$ is large.

%\add[New]{In addition to the above two metrics we also report the results in terms of the Root Mean Squared Error (RMSE). For each of these three metrics we also consider ``glucose-specific'' variants. These ``glucose-specific'' variants use a ``Clark Error Grid inspired penalty function'' to re-weight the relative costs of different mispredictions (see \hbox{Del Favero~\etal~\cite{del2012glucose}} for descriptions and definitions). Therefore in total, we consider the performance of our models across 6 different metrics. }

\subsection{Experimental Results: Heat-Maps}
\label{app:heatmaps}
\begin{figure}[!htbp]
\centering
\ShowFig{
\includegraphics[width=\textwidth]{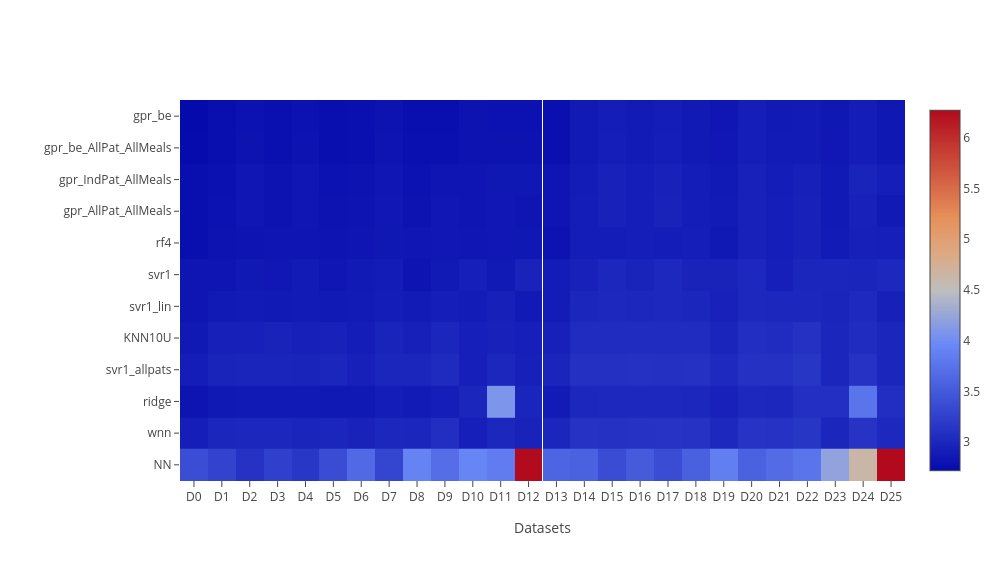}
}
\caption{{\bf Average L1 Loss: Datasets vs. Models.}
Datasets using the EP criteria are on the left half of the bisecting white line.
Each square represents the cross-validation L1 error, micro-averaged over patients.
\label{fig:AbsL1Te}
}
\end{figure}

\begin{figure}[!htbp]
\centering
\ShowFig{
\includegraphics[width=\textwidth]{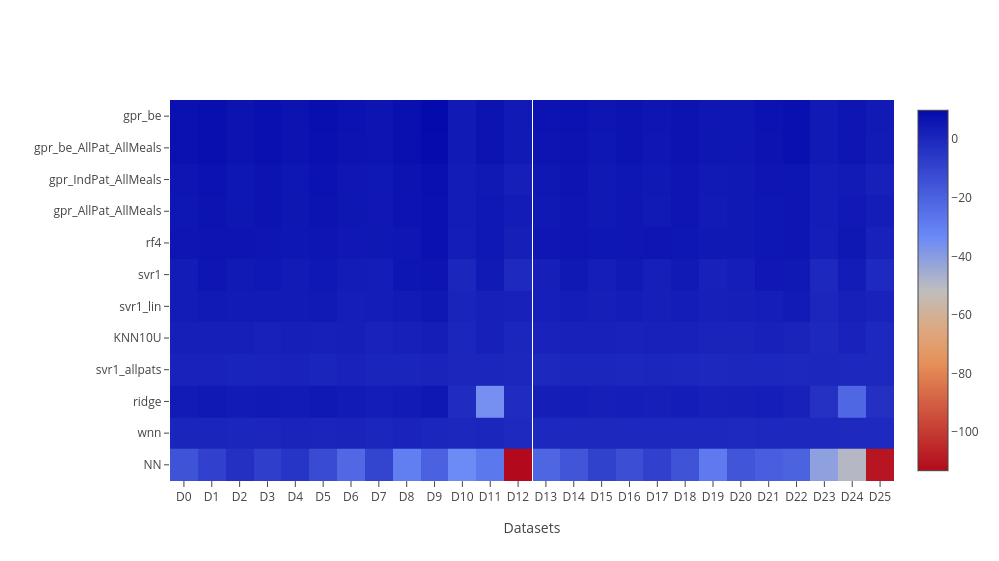}
}
\caption{{\bf Percent Improvement in Average L1 Loss for Models vs. Baseline.}
Datasets using the EP criteria are on the left half of the bisecting white line.
Each square corresponds with the percent change between the corresponding result in Figure~\ref{fig:AbsL1Te} and the performance of $M_{ave}$.}
\label{AbsL1Pi}
\end{figure}

\begin{figure}[!htbp]
\centering
\ShowFig{
\includegraphics[width=\textwidth]{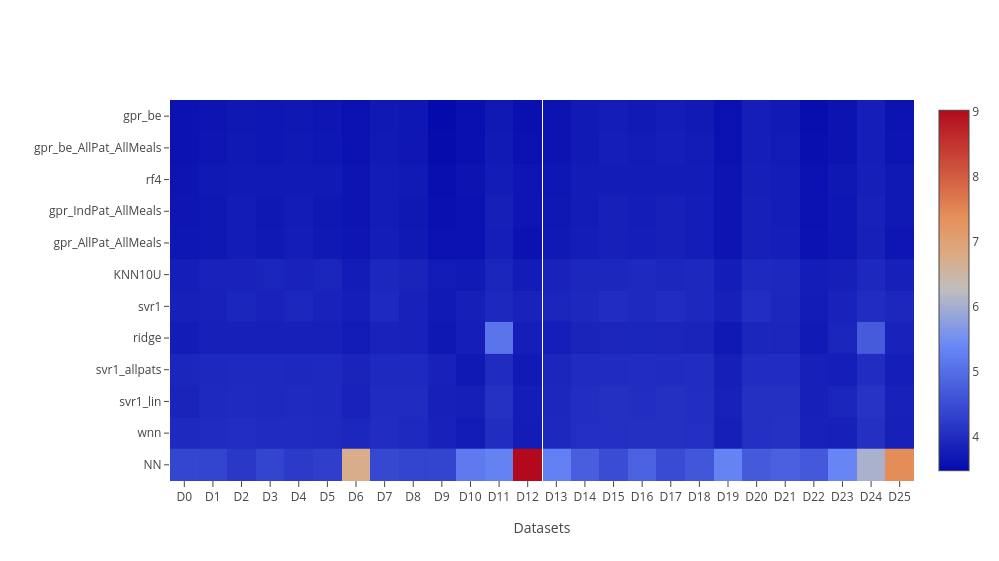}
}
\caption{{\bf Average Relative L1 Loss: Datasets vs. Models.}
Datasets using the EP criteria are on the left half of the bisecting white line. 
Each square represents the cross-validation relative L1 error, micro-averaged over patients.}
\label{AdjL1Te}
\end{figure}

\begin{figure}[!htbp]
\centering
\ShowFig{
\includegraphics[width=\textwidth]{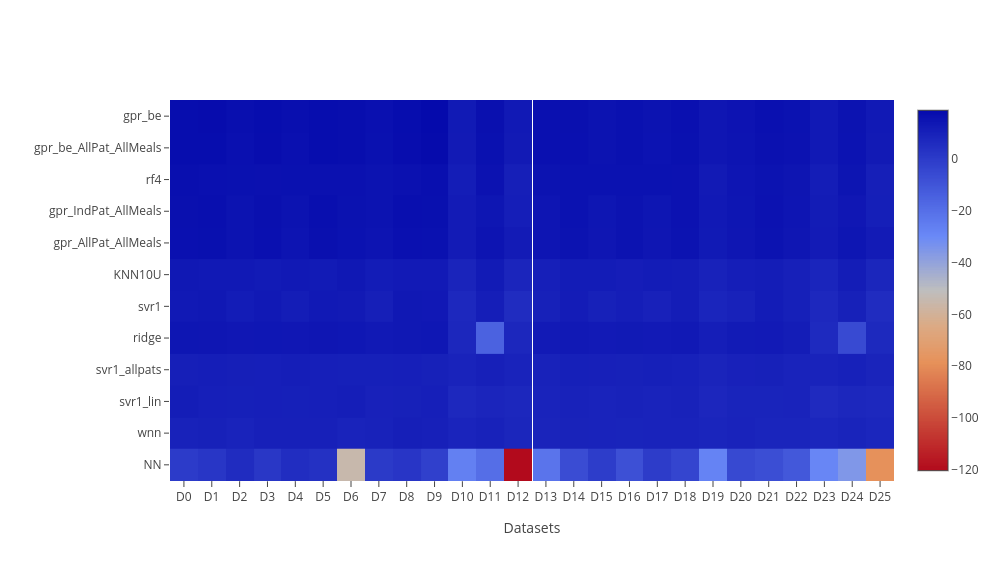}
}
\caption{{\bf Percent Improvement in Average Relative L1 Loss for Models vs. Baseline.}
Datasets using the EP criteria are on the left half of the bisecting white line. 
Each square corresponds with the percent change between the corresponding result in Figure~\ref{AdjL1Te} and the performance of $M_{ave}$.}
\label{AdjL1Pi}
\end{figure}

This sub-appendix provides a set of heat-maps (Figs~\ref{fig:AbsL1Te}, ~\ref{AbsL1Pi}, ~\ref{AdjL1Te} and ~\ref{AdjL1Pi}).
The left half of each heat-map contains the datasets that adhere to our EP rules, while the right half contains those datasets that do not.
Models (on the y axis) are  sorted in terms of their average $\errL$ error (over all 26 datasets),
so that the model with the best average $\errL$ error across all datasets appears at the top of these figures. 
Further, datasets \change[New]{D0 to D12}{$D_e$1 to $D_e$13} and \change[New]{D13 to D25}{$D_a$1 to $D_a$13} (in each half of these heat maps) are sorted horizontally,
in increasing order of average $\errL$ error across all models, with the left most dataset in each half having the smallest error. 
In Figure~\ref{fig:AbsL1Te}, 
the best model/dataset combination is found in the top left corner.
More detailed results for these heat-maps are seen in Borle~\cite{borle2017challenge}.

\pagebreak

\subsection{\add[New]{Experimental Results: Tables}}
\label{app:tables}

\begin{table}[!htbp] %[!ht]
\begin{adjustwidth}{-2.25in}{0in}
\centering
\caption{
{\bf Losses Corresponding to Fig~\ref{fig:AbsL1Te}.}}
\begin{tabular}{|>{\columncolor{LightCyan}}l||c|c|c|c|c|c|c|c|c|c|c|c|}
\hline
%Model & gpr_be_basic & gpr_be_AllPat_AllMeals_basic & gpr_IndPat_AllMeals_basic & gpr_AllPat_AllMeals_basic & rf4_basic & svr1_basic & svr1_lin_basic & bayes_basic & KNN10U_basic & svr1_allpats_basic & wnn_basic & NN_basic \\ \hline
\rowcolor[gray]{0.9}
& $M^{w}_{gpr}$ & $M^{ws}_{gpr}$ & $M_{gpr}$ & $M^{s}_{gpr}$ & $M_{rf}$ & $M_{svr}$ & $M^{lin}_{svr}$ & $M_{knn}$ & $M^{s}_{svr}$ & $M_{ridge}$ & $M_{wnn}$ & $M_{nn}$ \\ \hline
$D_e$1 & \bf{2.7} & 2.71 & 2.74 & 2.75 & 2.74 & 2.8 & 2.81 & 2.8 & 2.86 & 2.87 & 2.88 & 3.29 \\ \hline
$D_e$2 & 2.79 & 2.79 & 2.82 & 2.83 & 2.83 & 2.88 & 2.9 & 2.88 & 2.91 & 2.93 & 2.95 & 3.8 \\ \hline
$D_e$3 & 2.78 & 2.79 & 2.81 & 2.82 & 2.84 & 2.86 & 2.89 & 2.89 & 2.96 & 3.0 & 3.01 & 3.22 \\ \hline
$D_e$4 & 2.79 & 2.79 & 2.82 & 2.82 & 2.84 & 2.86 & 2.9 & 2.89 & 2.97 & 3.0 & 3.01 & 3.37 \\ \hline
$D_e$5 & 2.81 & 2.82 & 2.85 & 2.86 & 2.84 & 2.88 & 2.9 & 2.9 & 2.96 & 2.99 & 3.0 & 3.14 \\ \hline
$D_e$6 & 2.78 & 2.78 & 2.81 & 2.81 & 2.83 & 2.84 & 2.89 & 2.88 & 2.96 & 2.98 & 3.01 & 3.2 \\ \hline
$D_e$7 & 2.81 & 2.82 & 2.85 & 2.86 & 2.84 & 2.9 & 2.9 & 2.9 & 2.96 & 2.99 & 2.99 & 3.16 \\ \hline
$D_e$8 & 2.79 & 2.79 & 2.82 & 2.82 & 2.84 & 2.84 & 2.9 & 2.9 & 2.97 & 3.0 & 3.0 & 3.44 \\ \hline
$D_e$9 & 2.82 & 2.82 & 2.84 & 2.84 & 2.84 & 2.94 & 2.9 & 2.99 & 2.94 & 2.93 & 2.95 & 3.9 \\ \hline
$D_e$10 & 2.82 & 2.83 & 2.86 & 2.86 & 2.87 & 2.91 & 2.92 & 2.92 & 2.98 & 3.0 & 3.02 & 3.31 \\ \hline
$D_e$11 & 2.78 & 2.79 & 2.84 & 2.85 & 2.85 & 2.89 & 2.93 & 2.92 & 2.99 & 3.05 & 3.05 & 3.61 \\ \hline
$D_e$12 & 2.84 & 2.84 & 2.87 & 2.88 & 2.88 & 2.87 & 2.93 & 2.93 & 2.99 & 3.03 & 3.0 & 3.64 \\ \hline
$D_e$13 & 2.82 & 2.82 & 2.83 & 2.84 & 2.88 & 2.9 & 2.89 & 2.98 & 2.95 & 2.94 & 2.96 & 29.04 \\ \hline
\hline
$D_a$1 & 2.78 & 2.78 & 2.83 & 2.83 & 2.82 & 2.9 & 2.9 & 2.89 & 2.94 & 2.97 & 2.99 & 3.58 \\ \hline
$D_a$2 & 2.86 & 2.86 & 2.9 & 2.9 & 2.91 & 2.97 & 2.97 & 2.96 & 3.0 & 3.03 & 3.05 & 3.83 \\ \hline
$D_a$3 & 2.89 & 2.9 & 2.92 & 2.93 & 2.94 & 2.97 & 3.0 & 3.01 & 3.07 & 3.12 & 3.12 & 3.47 \\ \hline
$D_a$4 & 2.9 & 2.9 & 2.93 & 2.93 & 2.93 & 2.98 & 3.01 & 3.02 & 3.07 & 3.11 & 3.13 & 3.48 \\ \hline
$D_a$5 & 2.92 & 2.93 & 2.96 & 2.97 & 2.93 & 3.01 & 3.03 & 3.02 & 3.07 & 3.1 & 3.12 & 3.35 \\ \hline
$D_a$6 & 2.89 & 2.89 & 2.92 & 2.92 & 2.93 & 2.96 & 3.01 & 3.01 & 3.07 & 3.1 & 3.12 & 3.61 \\ \hline
$D_a$7 & 2.92 & 2.93 & 2.96 & 2.97 & 2.93 & 3.02 & 3.03 & 3.02 & 3.07 & 3.1 & 3.11 & 3.38 \\ \hline
$D_a$8 & 2.89 & 2.9 & 2.92 & 2.93 & 2.93 & 2.95 & 3.02 & 3.02 & 3.07 & 3.11 & 3.12 & 3.77 \\ \hline
$D_a$9 & 2.87 & 2.87 & 2.89 & 2.89 & 2.91 & 3.0 & 2.97 & 3.09 & 3.0 & 3.0 & 3.0 & 4.14 \\ \hline
$D_a$10 & 2.93 & 2.93 & 2.96 & 2.96 & 2.97 & 3.02 & 3.03 & 3.04 & 3.08 & 3.12 & 3.15 & 3.53 \\ \hline
$D_a$11 & 2.9 & 2.9 & 2.95 & 2.96 & 2.96 & 3.0 & 3.02 & 3.1 & 3.12 & 3.16 & 3.17 & 4.22 \\ \hline
$D_a$12 & 2.94 & 2.94 & 2.97 & 2.97 & 2.96 & 2.98 & 3.03 & 3.05 & 3.09 & 3.13 & 3.12 & 4.7 \\ \hline
$D_a$13 & 2.87 & 2.87 & 2.89 & 2.89 & 2.94 & 2.96 & 2.95 & 3.06 & 3.01 & 3.0 & 3.01 & 6.1 \\ \hline
\end{tabular}
\vspace{.1cm}
\begin{flushleft} See Table~\ref{tab:datasets} for descriptions of the different datasets and Table~\ref{tab:models} for descriptions of the models.
\end{flushleft}
\label{table:AbsL1Te}
\end{adjustwidth}
\end{table}

\begin{table}[!htbp] %[!ht]
\begin{adjustwidth}{-2.25in}{0in}
\centering
\caption{
{\bf Percentage Improvements Corresponding to Fig~\ref{AbsL1Pi}.}}
\begin{tabular}{|>{\columncolor{LightCyan}}l||c|c|c|c|c|c|c|c|c|c|c|c|}
\hline
\rowcolor[gray]{0.9}
& $M^{w}_{gpr}$ & $M^{ws}_{gpr}$ & $M_{gpr}$ & $M^{s}_{gpr}$ & $M_{rf}$ & $M_{svr}$ & $M^{lin}_{svr}$ & $M_{ridge}$ & $M_{knn}$ & $M^{s}_{svr}$ & $M_{wnn}$ & $M_{nn}$ \\ \hline
%Model & gpr_be & gpr_be_AllPat_AllMeals & gpr_IndPat_AllMeals & gpr_AllPat_AllMeals & rf4 & svr1 & svr1_lin & ridge & KNN10U & svr1_allpats & wnn & NN \\ \hline
$D_e$1 & 7.12 & 7.02 & 5.75 & 5.54 & 5.95 & 3.73 & 3.58 & 3.77 & 1.8 & 1.45 & 1.11 & -13.15 \\ \hline
$D_e$2 & 6.19 & 6.14 & 5.02 & 4.8 & 4.62 & 2.93 & 2.28 & 3.18 & 1.97 & 1.48 & 0.85 & -27.86 \\ \hline
$D_e$3 & 7.98 & 7.85 & 6.96 & 6.83 & 6.03 & 5.51 & 4.43 & 4.6 & 1.98 & 0.88 & 0.56 & -6.35 \\ \hline
$D_e$4 & 7.79 & 7.74 & 6.82 & 6.75 & 6.07 & 5.47 & 4.26 & 4.44 & 1.66 & 0.87 & 0.39 & -11.52 \\ \hline
$D_e$5 & 7.01 & 6.82 & 5.86 & 5.59 & 6.13 & 4.61 & 4.02 & 4.27 & 2.19 & 1.05 & 0.69 & -3.69 \\ \hline
$D_e$6 & 8.18 & 8.02 & 7.23 & 6.98 & 6.34 & 6.09 & 4.38 & 4.62 & 2.24 & 1.3 & 0.48 & -5.91 \\ \hline
$D_e$7 & 7.01 & 6.76 & 5.86 & 5.57 & 6.2 & 4.25 & 4.02 & 4.27 & 2.19 & 1.19 & 0.98 & -4.47 \\ \hline
$D_e$8 & 7.93 & 7.8 & 6.91 & 6.73 & 6.04 & 6.27 & 4.06 & 3.99 & 1.9 & 0.73 & 0.94 & -13.7 \\ \hline
$D_e$9 & 4.2 & 4.15 & 3.5 & 3.49 & 3.39 & 0.11 & 1.42 & -1.55 & 0.01 & 0.16 & -0.23 & -32.75 \\ \hline
$D_e$10 & 6.65 & 6.48 & 5.57 & 5.38 & 5.06 & 3.62 & 3.33 & 3.51 & 1.33 & 0.75 & 0.09 & -9.39 \\ \hline
$D_e$11 & \bf{9.94} & 9.72 & 8.04 & 7.61 & 7.61 & 6.47 & 5.15 & 5.33 & 3.11 & 1.09 & 1.23 & -16.83 \\ \hline
$D_e$12 & 6.27 & 6.03 & 5.0 & 4.71 & 4.96 & 4.99 & 3.17 & 2.85 & 1.29 & -0.13 & 0.72 & -20.35 \\ \hline
$D_e$13 & 4.1 & 3.99 & 3.7 & 3.53 & 2.15 & 1.19 & 1.8 & -1.24 & -0.4 & -0.18 & -0.75 & -887.75 \\ \hline
\hline
$D_a$1 & 6.14 & 6.08 & 4.63 & 4.49 & 4.79 & 2.14 & 2.13 & 2.37 & 0.9 & -0.18 & -0.73 & -20.68 \\ \hline
$D_a$2 & 4.97 & 4.9 & 3.69 & 3.53 & 3.31 & 1.27 & 1.31 & 1.59 & 0.5 & -0.53 & -1.49 & -27.2 \\ \hline
$D_a$3 & 7.02 & 6.97 & 6.1 & 6.0 & 5.62 & 4.57 & 3.54 & 3.27 & 1.42 & -0.06 & -0.26 & -11.37 \\ \hline
$D_a$4 & 6.9 & 6.86 & 5.91 & 5.83 & 5.78 & 4.42 & 3.34 & 2.97 & 1.35 & -0.01 & -0.52 & -11.91 \\ \hline
$D_a$5 & 6.07 & 5.93 & 4.84 & 4.61 & 6.04 & 3.33 & 2.83 & 2.84 & 1.49 & 0.28 & -0.28 & -7.63 \\ \hline
$D_a$6 & 7.23 & 7.07 & 6.34 & 6.1 & 5.96 & 4.96 & 3.43 & 3.27 & 1.51 & 0.38 & -0.12 & -15.97 \\ \hline
$D_a$7 & 6.07 & 5.9 & 4.84 & 4.59 & 6.05 & 2.89 & 2.83 & 2.84 & 1.49 & 0.35 & 0.06 & -8.41 \\ \hline
$D_a$8 & 7.13 & 6.93 & 6.22 & 5.97 & 5.84 & 5.4 & 3.13 & 2.99 & 1.33 & -0.01 & -0.03 & -20.94 \\ \hline
$D_a$9 & 4.21 & 4.2 & 3.29 & 3.25 & 2.8 & -0.19 & 0.56 & -3.19 & -0.22 & -0.15 & -0.16 & -38.48 \\ \hline
$D_a$10 & 5.95 & 5.85 & 4.92 & 4.79 & 4.73 & 3.05 & 2.72 & 2.46 & 0.93 & -0.17 & -1.19 & -13.24 \\ \hline
$D_a$11 & 7.91 & 7.81 & 6.26 & 5.99 & 6.17 & 4.73 & 4.22 & 1.68 & 1.1 & -0.39 & -0.74 & -34.05 \\ \hline
$D_a$12 & 5.64 & 5.53 & 4.66 & 4.49 & 4.8 & 4.47 & 2.55 & 2.15 & 0.75 & -0.59 & -0.27 & -51.06 \\ \hline
$D_a$13 & 4.11 & 4.0 & 3.51 & 3.36 & 1.62 & 1.1 & 1.47 & -2.33 & -0.69 & -0.46 & -0.61 & -103.85 \\ \hline
\end{tabular}
\vspace{.1cm}
\begin{flushleft} See Table~\ref{tab:datasets} for descriptions of the different datasets and Table~\ref{tab:models} for descriptions of the models.
\end{flushleft}
\label{table:AbsL1Pi}
\end{adjustwidth}
\end{table}

\begin{table}[!htbp] %[!ht]
\begin{adjustwidth}{-2.25in}{0in}
\centering
\caption{
{\bf Losses Corresponding to Fig~\ref{AdjL1Te}.}}
\begin{tabular}{|>{\columncolor{LightCyan}}l||c|c|c|c|c|c|c|c|c|c|c|c|}
\hline
\rowcolor[gray]{0.9}
& $M^{w}_{gpr}$ & $M^{ws}_{gpr}$ & $M_{gpr}$ & $M_{rf}$ & $M^{s}_{gpr}$ & $M_{ridge}$ & $M_{knn}$ & $M_{svr}$ & $M^{s}_{svr}$ & $M_{wnn}$ & $M^{lin}_{svr}$ & $M_{nn}$ \\ \hline
%Model & gpr_be & gpr_be_AllPat_AllMeals & gpr_IndPat_AllMeals & rf4 & gpr_AllPat_AllMeals & ridge & KNN10U & svr1 & svr1_allpats & wnn & svr1_lin & NN \\ \hline
$D_e$1 & 3.61 & 3.61 & 3.65 & 3.64 & 3.66 & 3.76 & 3.81 & 3.81 & 3.9 & 3.95 & 3.88 & 4.62 \\ \hline
$D_e$2 & 3.62 & 3.62 & 3.68 & 3.65 & 3.69 & 3.77 & 3.78 & 3.82 & 3.89 & 3.97 & 3.9 & 4.6 \\ \hline
$D_e$3 & 3.67 & 3.67 & 3.71 & 3.75 & 3.72 & 3.83 & 3.91 & 3.88 & 3.99 & 4.0 & 3.99 & 4.28 \\ \hline
$D_e$4 & 3.67 & 3.68 & 3.72 & 3.75 & 3.72 & 3.84 & 3.92 & 3.88 & 3.99 & 4.02 & 3.99 & 4.3 \\ \hline
$D_e$5 & 3.71 & 3.72 & 3.77 & 3.74 & 3.78 & 3.85 & 3.89 & 3.93 & 3.98 & 4.01 & 4.0 & 4.24 \\ \hline
$D_e$6 & 3.66 & 3.67 & 3.7 & 3.74 & 3.72 & 3.83 & 3.89 & 3.85 & 3.96 & 4.03 & 3.99 & 4.27 \\ \hline
$D_e$7 & 3.71 & 3.73 & 3.77 & 3.73 & 3.78 & 3.85 & 3.89 & 3.93 & 3.97 & 4.0 & 4.0 & 4.26 \\ \hline
$D_e$8 & 3.67 & 3.68 & 3.71 & 3.75 & 3.72 & 3.86 & 3.91 & 3.85 & 3.99 & 4.02 & 4.02 & 4.37 \\ \hline
$D_e$9 & 3.58 & 3.58 & 3.6 & 3.6 & 3.6 & 3.77 & 3.74 & 3.81 & 3.71 & 3.74 & 3.79 & 4.79 \\ \hline
$D_e$10 & 3.73 & 3.74 & 3.78 & 3.8 & 3.79 & 3.87 & 3.94 & 3.98 & 3.99 & 4.03 & 4.02 & 4.41 \\ \hline
$D_e$11 & \bf{3.48} & 3.5 & 3.57 & 3.57 & 3.59 & 3.69 & 3.74 & 3.72 & 3.86 & 3.85 & 3.84 & 4.68 \\ \hline
$D_e$12 & 3.75 & 3.76 & 3.8 & 3.8 & 3.81 & 3.89 & 3.95 & 3.93 & 4.03 & 4.0 & 4.05 & 4.73 \\ \hline
$D_e$13 & 3.58 & 3.58 & 3.59 & 3.64 & 3.6 & 3.77 & 3.77 & 3.77 & 3.73 & 3.76 & 3.77 & 6.02 \\ \hline
\hline
$D_a$1 & 3.64 & 3.64 & 3.71 & 3.68 & 3.72 & 3.81 & 3.89 & 3.9 & 3.91 & 3.93 & 3.95 & 4.73 \\ \hline
$D_a$2 & 3.62 & 3.62 & 3.67 & 3.66 & 3.67 & 3.75 & 3.8 & 3.86 & 3.81 & 3.85 & 3.88 & 4.73 \\ \hline
$D_a$3 & 3.75 & 3.75 & 3.78 & 3.79 & 3.79 & 3.9 & 3.98 & 3.97 & 4.03 & 4.06 & 4.05 & 4.66 \\ \hline
$D_a$4 & 3.75 & 3.75 & 3.79 & 3.79 & 3.79 & 3.91 & 3.98 & 3.98 & 4.03 & 4.08 & 4.06 & 4.82 \\ \hline
$D_a$5 & 3.79 & 3.79 & 3.84 & 3.79 & 3.85 & 3.92 & 3.96 & 4.02 & 4.02 & 4.07 & 4.08 & 4.45 \\ \hline
$D_a$6 & 3.74 & 3.74 & 3.77 & 3.78 & 3.78 & 3.9 & 3.96 & 3.95 & 4.01 & 4.06 & 4.05 & 4.63 \\ \hline
$D_a$7 & 3.79 & 3.8 & 3.84 & 3.78 & 3.85 & 3.92 & 3.96 & 4.03 & 4.02 & 4.04 & 4.08 & 4.48 \\ \hline
$D_a$8 & 3.75 & 3.75 & 3.78 & 3.8 & 3.79 & 3.93 & 3.98 & 3.94 & 4.03 & 4.09 & 4.09 & 4.78 \\ \hline
$D_a$9 & 3.63 & 3.63 & 3.66 & 3.69 & 3.66 & 3.88 & 3.81 & 3.87 & 3.78 & 3.84 & 3.88 & 5.35 \\ \hline
$D_a$10 & 3.8 & 3.8 & 3.84 & 3.82 & 3.85 & 3.93 & 4.0 & 4.05 & 4.04 & 4.05 & 4.09 & 4.68 \\ \hline
$D_a$11 & 3.55 & 3.56 & 3.61 & 3.6 & 3.62 & 3.73 & 3.78 & 3.78 & 3.83 & 3.88 & 3.84 & 4.73 \\ \hline
$D_a$12 & 3.82 & 3.82 & 3.86 & 3.85 & 3.87 & 3.97 & 4.02 & 4.01 & 4.06 & 4.07 & 4.13 & 11.06 \\ \hline
$D_a$13 & 3.63 & 3.64 & 3.65 & 3.72 & 3.66 & 3.86 & 3.84 & 3.82 & 3.8 & 3.85 & 3.85 & 7.52 \\ \hline
\end{tabular}
\vspace{.1cm}
\begin{flushleft} See Table~\ref{tab:datasets} for descriptions of the different datasets and Table~\ref{tab:models} for descriptions of the models. Here all values are multiplied by 10 so that they are the same order of magnitude as in Table~\ref{table:AbsL1Te}.
\end{flushleft}
\label{table:AdjL1Te}
\end{adjustwidth}
\end{table}

\begin{table}[!htbp] %[!ht]
\begin{adjustwidth}{-2.25in}{0in}
\centering
\caption{
{\bf Percentage Improvements Corresponding to Fig~\ref{AdjL1Pi}.}}
\begin{tabular}{|>{\columncolor{LightCyan}}l||c|c|c|c|c|c|c|c|c|c|c|c|}
\hline
\rowcolor[gray]{0.9}
& $M^{w}_{gpr}$ & $M^{ws}_{gpr}$ & $M_{gpr}$ & $M_{rf}$ & $M^{s}_{gpr}$ & $M_{ridge}$ & $M_{knn}$ & $M_{svr}$ & $M^{s}_{svr}$ & $M^{lin}_{svr}$ & $M_{wnn}$ & $M_{nn}$ \\ \hline
%Model & gpr_be & gpr_be_AllPat_AllMeals & gpr_IndPat_AllMeals & rf4 & gpr_AllPat_AllMeals & ridge & KNN10U & svr1 & svr1_allpats & svr1_lin & wnn & NN \\ \hline
$D_e$1 & 17.53 & 17.45 & 16.52 & 16.76 & 16.33 & 14.19 & 12.9 & 12.99 & 10.96 & 11.37 & 9.68 & -5.64 \\ \hline
$D_e$2 & 16.69 & 16.63 & 15.34 & 16.03 & 15.13 & 13.16 & 12.99 & 12.06 & 10.56 & 10.16 & 8.59 & -5.95 \\ \hline
$D_e$3 & 17.97 & 17.79 & 16.9 & 16.16 & 16.71 & 14.2 & 12.47 & 13.16 & 10.83 & 10.76 & 10.4 & 4.17 \\ \hline
$D_e$4 & 17.81 & 17.73 & 16.79 & 16.15 & 16.72 & 14.04 & 12.3 & 13.13 & 10.73 & 10.78 & 10.07 & 3.82 \\ \hline
$D_e$5 & 16.91 & 16.73 & 15.64 & 16.42 & 15.39 & 13.89 & 13.0 & 12.14 & 10.87 & 10.58 & 10.32 & 5.19 \\ \hline
$D_e$6 & 18.16 & 17.95 & 17.16 & 16.34 & 16.86 & 14.21 & 13.01 & 13.76 & 11.31 & 10.79 & 9.93 & 4.5 \\ \hline
$D_e$7 & 16.91 & 16.64 & 15.64 & 16.43 & 15.32 & 13.89 & 13.0 & 11.96 & 11.1 & 10.58 & 10.39 & 4.67 \\ \hline
$D_e$8 & 17.85 & 17.7 & 16.92 & 16.07 & 16.73 & 13.59 & 12.45 & 13.77 & 10.69 & 10.16 & 10.12 & 2.34 \\ \hline
$D_e$9 & 12.68 & 12.61 & 12.07 & 12.09 & 12.06 & 7.84 & 8.65 & 7.01 & 9.38 & 7.43 & 8.72 & -16.97 \\ \hline
$D_e$10 & 16.57 & 16.38 & 15.46 & 14.89 & 15.26 & 13.36 & 11.91 & 11.01 & 10.62 & 10.03 & 9.82 & 1.38 \\ \hline
$D_e$11 & \bf{18.97} & 18.7 & 16.87 & 16.89 & 16.42 & 14.1 & 12.95 & 13.47 & 10.26 & 10.72 & 10.5 & -8.95 \\ \hline
$D_e$12 & 16.13 & 15.86 & 15.06 & 14.9 & 14.73 & 12.82 & 11.7 & 12.11 & 9.8 & 9.42 & 10.51 & -5.72 \\ \hline
$D_e$13 & 12.65 & 12.49 & 12.27 & 11.01 & 12.06 & 7.96 & 7.93 & 7.99 & 9.01 & 7.83 & 8.31 & -46.92 \\ \hline
\hline
$D_a$1 & 15.85 & 15.83 & 14.3 & 14.9 & 14.17 & 12.11 & 10.15 & 9.93 & 9.79 & 8.86 & 9.22 & -9.33 \\ \hline
$D_a$2 & 13.7 & 13.63 & 12.54 & 12.67 & 12.41 & 10.56 & 9.29 & 7.96 & 9.05 & 7.39 & 8.14 & -12.83 \\ \hline
$D_a$3 & 16.61 & 16.54 & 15.77 & 15.59 & 15.66 & 13.19 & 11.36 & 11.66 & 10.18 & 9.73 & 9.56 & -3.7 \\ \hline
$D_a$4 & 16.55 & 16.5 & 15.63 & 15.71 & 15.56 & 13.02 & 11.39 & 11.41 & 10.22 & 9.71 & 9.15 & -7.34 \\ \hline
$D_a$5 & 15.68 & 15.54 & 14.52 & 15.67 & 14.31 & 12.76 & 11.85 & 10.45 & 10.48 & 9.26 & 9.36 & 0.96 \\ \hline
$D_a$6 & 16.84 & 16.65 & 16.03 & 15.85 & 15.76 & 13.17 & 11.84 & 12.02 & 10.71 & 9.75 & 9.58 & -3.01 \\ \hline
$D_a$7 & 15.68 & 15.51 & 14.52 & 15.91 & 14.28 & 12.76 & 11.85 & 10.18 & 10.62 & 9.26 & 10.0 & 0.23 \\ \hline
$D_a$8 & 16.64 & 16.43 & 15.83 & 15.36 & 15.59 & 12.6 & 11.44 & 12.2 & 10.29 & 9.07 & 9.0 & -6.37 \\ \hline
$D_a$9 & 12.88 & 12.89 & 12.11 & 11.57 & 12.1 & 6.83 & 8.53 & 7.19 & 9.21 & 6.86 & 7.8 & -28.32 \\ \hline
$D_a$10 & 15.47 & 15.35 & 14.51 & 14.86 & 14.38 & 12.45 & 11.04 & 9.85 & 10.0 & 9.02 & 9.74 & -4.16 \\ \hline
$D_a$11 & 15.88 & 15.74 & 14.61 & 14.71 & 14.33 & 11.55 & 10.49 & 10.47 & 9.25 & 9.15 & 8.15 & -11.98 \\ \hline
$D_a$12 & 15.02 & 14.9 & 14.15 & 14.34 & 13.95 & 11.59 & 10.63 & 10.81 & 9.73 & 7.97 & 9.49 & -146.23 \\ \hline
$D_a$13 & 12.89 & 12.76 & 12.47 & 10.71 & 12.29 & 7.49 & 7.98 & 8.28 & 8.85 & 7.64 & 7.68 & -80.39 \\ \hline
\end{tabular}
\vspace{.1cm}
\begin{flushleft} See Table~\ref{tab:datasets} for descriptions of the different datasets and Table~\ref{tab:models} for descriptions of the models.
\end{flushleft}
\label{table:AdjL1Pi}
\end{adjustwidth}
\end{table}

% gMAD gMARD SECTION

\begin{table}[!htbp] %[!ht]
\begin{adjustwidth}{-2.25in}{0in}
\centering
\caption{
{\bf Losses Using gMAD.}}
\begin{tabular}{|>{\columncolor{LightCyan}}l||c|c|c|c|c|c|c|c|c|c|c|c|}
\hline
%Model & gpr_be_basic & gpr_be_AllPat_AllMeals_basic & gpr_IndPat_AllMeals_basic & gpr_AllPat_AllMeals_basic & rf4_basic & svr1_basic & svr1_lin_basic & bayes_basic & KNN10U_basic & svr1_allpats_basic & wnn_basic & NN_basic \\ \hline
\rowcolor[gray]{0.9}
& $M_{svr}$ & $M^{w}_{gpr}$ & $M^{ws}_{gpr}$ & $M_{rf}$ & $M_{gpr}$ & $M^{s}_{gpr}$ & $M^{lin}_{svr}$ & $M_{ridge}$ & $M_{knn}$ & $M^{s}_{svr}$ & $M_{wnn}$ & $M_{nn}$ \\ \hline
%Model & svr1 & gpr_be & gpr_be_AllPat_AllMeals & rf4 & gpr_IndPat_AllMeals & gpr_AllPat_AllMeals & svr1_lin & ridge & KNN10U & svr1_allpats & wnn & NN \\ \hline
$D_e$1 & 5.03 & \bf{4.98} & 4.99 & 5.05 & 5.08 & 5.09 & 5.12 & 5.19 & 5.24 & 5.37 & 5.38 & 5.75 \\ \hline
$D_e$2 & 5.13 & 5.09 & 5.09 & 5.18 & 5.17 & 5.17 & 5.21 & 5.26 & 5.28 & 5.4 & 5.42 & 6.43 \\ \hline
$D_e$3 & 5.16 & 5.19 & 5.2 & 5.31 & 5.25 & 5.26 & 5.32 & 5.41 & 5.48 & 5.69 & 5.68 & 5.68 \\ \hline
$D_e$4 & 5.16 & 5.2 & 5.21 & 5.31 & 5.26 & 5.26 & 5.33 & 5.42 & 5.49 & 5.7 & 5.7 & 6.1 \\ \hline
$D_e$5 & 5.21 & 5.25 & 5.26 & 5.3 & 5.32 & 5.33 & 5.33 & 5.41 & 5.47 & 5.6 & 5.7 & 5.58 \\ \hline
$D_e$6 & 5.13 & 5.17 & 5.18 & 5.29 & 5.23 & 5.25 & 5.32 & 5.41 & 5.47 & 5.65 & 5.68 & 5.67 \\ \hline
$D_e$7 & 5.22 & 5.25 & 5.27 & 5.3 & 5.32 & 5.33 & 5.33 & 5.41 & 5.47 & 5.62 & 5.66 & 5.63 \\ \hline
$D_e$8 & 5.14 & 5.2 & 5.21 & 5.31 & 5.27 & 5.28 & 5.36 & 5.47 & 5.5 & 5.67 & 5.66 & 5.97 \\ \hline
$D_e$9 & 5.2 & 5.18 & 5.18 & 5.17 & 5.22 & 5.22 & 5.15 & 5.32 & 5.34 & 5.49 & 5.46 & 6.4 \\ \hline
$D_e$10 & 5.27 & 5.27 & 5.28 & 5.37 & 5.34 & 5.35 & 5.4 & 5.48 & 5.51 & 5.7 & 5.71 & 5.82 \\ \hline
$D_e$11 & 5.1 & 5.08 & 5.09 & 5.2 & 5.18 & 5.2 & 5.23 & 5.33 & 5.41 & 5.63 & 5.59 & 6.04 \\ \hline
$D_e$12 & 5.25 & 5.33 & 5.34 & 5.39 & 5.42 & 5.43 & 5.46 & 5.56 & 5.54 & 5.69 & 5.68 & 6.37 \\ \hline
$D_e$13 & 5.15 & 5.18 & 5.18 & 5.23 & 5.21 & 5.22 & 5.23 & 5.4 & 5.35 & 5.49 & 5.49 & 31.86 \\ \hline
\hline
$D_a$1 & 5.2 & 5.14 & 5.14 & 5.19 & 5.25 & 5.25 & 5.29 & 5.36 & 5.4 & 5.6 & 5.61 & 6.17 \\ \hline
$D_a$2 & 5.28 & 5.23 & 5.23 & 5.3 & 5.32 & 5.33 & 5.33 & 5.42 & 5.44 & 5.63 & 5.65 & 6.44 \\ \hline
$D_a$3 & 5.36 & 5.4 & 5.41 & 5.48 & 5.47 & 5.47 & 5.52 & 5.64 & 5.7 & 5.94 & 5.94 & 6.09 \\ \hline
$D_a$4 & 5.37 & 5.41 & 5.41 & 5.47 & 5.48 & 5.49 & 5.53 & 5.65 & 5.69 & 5.95 & 5.92 & 6.11 \\ \hline
$D_a$5 & 5.43 & 5.46 & 5.47 & 5.46 & 5.55 & 5.57 & 5.55 & 5.65 & 5.69 & 5.86 & 5.93 & 5.91 \\ \hline
$D_a$6 & 5.34 & 5.38 & 5.4 & 5.46 & 5.45 & 5.47 & 5.52 & 5.64 & 5.69 & 5.91 & 5.89 & 6.2 \\ \hline
$D_a$7 & 5.45 & 5.46 & 5.48 & 5.45 & 5.55 & 5.57 & 5.55 & 5.65 & 5.69 & 5.88 & 5.91 & 5.93 \\ \hline
$D_a$8 & 5.33 & 5.4 & 5.41 & 5.47 & 5.47 & 5.48 & 5.57 & 5.68 & 5.7 & 5.92 & 5.92 & 6.42 \\ \hline
$D_a$9 & 5.33 & 5.28 & 5.28 & 5.33 & 5.34 & 5.34 & 5.29 & 5.49 & 5.47 & 5.62 & 5.57 & 6.81 \\ \hline
$D_a$10 & 5.46 & 5.47 & 5.48 & 5.53 & 5.55 & 5.56 & 5.58 & 5.69 & 5.72 & 5.95 & 5.99 & 6.18 \\ \hline
$D_a$11 & 5.3 & 5.29 & 5.3 & 5.38 & 5.41 & 5.42 & 5.38 & 5.59 & 5.67 & 5.88 & 5.84 & 6.79 \\ \hline
$D_a$12 & 5.42 & 5.52 & 5.53 & 5.55 & 5.6 & 5.61 & 5.64 & 5.76 & 5.74 & 5.93 & 5.93 & 7.73 \\ \hline
$D_a$13 & 5.26 & 5.28 & 5.28 & 5.37 & 5.32 & 5.33 & 5.35 & 5.55 & 5.49 & 5.61 & 5.58 & 9.04 \\ \hline
\end{tabular}
\vspace{.1cm}
\begin{flushleft} See Table~\ref{tab:datasets} for descriptions of the different datasets and Table~\ref{tab:models} for descriptions of the models.
\end{flushleft}
\label{table:AbsL1Te}
\end{adjustwidth}
\end{table}

\begin{table}[!htbp] %[!ht]
\begin{adjustwidth}{-2.25in}{0in}
\centering
\caption{
{\bf Percentage Improvements for gMAD.}}
\begin{tabular}{|>{\columncolor{LightCyan}}l||c|c|c|c|c|c|c|c|c|c|c|c|}
\hline
\rowcolor[gray]{0.9}
& $M_{svr}$ & $M^{w}_{gpr}$ & $M^{ws}_{gpr}$ & $M_{rf}$ & $M_{gpr}$ & $M^{s}_{gpr}$ & $M^{lin}_{svr}$ & $M_{ridge}$ & $M_{knn}$ & $M^{s}_{svr}$ & $M_{wnn}$ & $M_{nn}$ \\ \hline
%Model & svr1 & gpr_be & gpr_be_AllPat_AllMeals & rf4 & gpr_IndPat_AllMeals & gpr_AllPat_AllMeals & svr1_lin & ridge & KNN10U & svr1_allpats & wnn & NN \\ \hline
$D_e$1 & 5.41 & 6.28 & 6.19 & 5.07 & 4.43 & 4.23 & 3.73 & 2.35 & 1.39 & -1.0 & -1.15 & -8.12 \\ \hline
$D_e$2 & 3.29 & 4.02 & 4.01 & 2.45 & 2.62 & 2.46 & 1.8 & 0.91 & 0.57 & -1.73 & -2.17 & -21.15 \\ \hline
$D_e$3 & 7.47 & 6.9 & 6.69 & 4.73 & 5.81 & 5.67 & 4.52 & 2.93 & 1.7 & -2.08 & -1.95 & -1.99 \\ \hline
$D_e$4 & 7.38 & 6.68 & 6.57 & 4.77 & 5.61 & 5.56 & 4.36 & 2.81 & 1.46 & -2.23 & -2.35 & -9.52 \\ \hline
$D_e$5 & 6.58 & 5.81 & 5.56 & 4.97 & 4.6 & 4.28 & 4.44 & 2.85 & 1.8 & -0.43 & -2.27 & -0.07 \\ \hline
$D_e$6 & \bf{8.03} & 7.2 & 6.97 & 5.07 & 6.17 & 5.89 & 4.6 & 3.02 & 1.83 & -1.37 & -2.0 & -1.82 \\ \hline
$D_e$7 & 6.41 & 5.81 & 5.5 & 4.94 & 4.6 & 4.29 & 4.44 & 2.85 & 1.8 & -0.76 & -1.49 & -1.1 \\ \hline
$D_e$8 & 7.78 & 6.72 & 6.59 & 4.72 & 5.5 & 5.35 & 3.79 & 1.92 & 1.37 & -1.62 & -1.41 & -7.01 \\ \hline
$D_e$9 & 1.36 & 1.79 & 1.72 & 1.85 & 1.0 & 0.98 & 2.23 & -0.93 & -1.26 & -4.19 & -3.6 & -21.35 \\ \hline
$D_e$10 & 5.43 & 5.4 & 5.19 & 3.63 & 4.19 & 3.97 & 3.16 & 1.7 & 1.06 & -2.32 & -2.54 & -4.43 \\ \hline
$D_e$11 & 6.54 & 6.94 & 6.73 & 4.74 & 5.03 & 4.65 & 4.09 & 2.35 & 0.87 & -3.16 & -2.35 & -10.64 \\ \hline
$D_e$12 & 5.9 & 4.49 & 4.31 & 3.27 & 2.85 & 2.61 & 2.07 & 0.07 & 0.59 & -2.08 & -1.9 & -14.29 \\ \hline
$D_e$13 & 2.37 & 1.79 & 1.68 & 0.82 & 1.28 & 1.07 & 0.85 & -2.49 & -1.47 & -4.03 & -4.12 & -504.3 \\ \hline
\hline
$D_a$1 & 4.59 & 5.7 & 5.67 & 4.83 & 3.76 & 3.63 & 2.95 & 1.58 & 0.85 & -2.83 & -2.89 & -13.21 \\ \hline
$D_a$2 & 2.45 & 3.29 & 3.27 & 1.98 & 1.65 & 1.52 & 1.47 & -0.14 & -0.62 & -4.07 & -4.34 & -19.05 \\ \hline
$D_a$3 & 6.94 & 6.24 & 6.14 & 4.87 & 5.08 & 4.98 & 4.09 & 2.07 & 1.11 & -3.21 & -3.06 & -5.73 \\ \hline
$D_a$4 & 6.72 & 6.07 & 6.0 & 5.07 & 4.83 & 4.76 & 3.92 & 1.89 & 1.25 & -3.23 & -2.81 & -6.07 \\ \hline
$D_a$5 & 5.64 & 5.15 & 4.96 & 5.24 & 3.63 & 3.35 & 3.71 & 1.89 & 1.24 & -1.71 & -3.0 & -2.58 \\ \hline
$D_a$6 & 7.35 & 6.51 & 6.29 & 5.18 & 5.41 & 5.11 & 4.08 & 2.12 & 1.24 & -2.56 & -2.34 & -7.68 \\ \hline
$D_a$7 & 5.37 & 5.15 & 4.9 & 5.31 & 3.63 & 3.32 & 3.71 & 1.89 & 1.24 & -2.0 & -2.67 & -2.94 \\ \hline
$D_a$8 & 7.5 & 6.22 & 6.05 & 5.11 & 5.07 & 4.84 & 3.38 & 1.36 & 0.99 & -2.66 & -2.81 & -11.38 \\ \hline
$D_a$9 & 1.14 & 2.03 & 2.02 & 1.13 & 0.88 & 0.85 & 1.73 & -1.85 & -1.55 & -4.29 & -3.46 & -26.43 \\ \hline
$D_a$10 & 5.21 & 5.02 & 4.88 & 3.9 & 3.69 & 3.53 & 3.09 & 1.23 & 0.75 & -3.36 & -4.06 & -7.38 \\ \hline
$D_a$11 & 5.62 & 5.66 & 5.6 & 4.15 & 3.63 & 3.38 & 4.11 & 0.35 & -1.02 & -4.77 & -4.08 & -21.08 \\ \hline
$D_a$12 & 5.98 & 4.13 & 4.07 & 3.76 & 2.86 & 2.7 & 2.16 & -0.08 & 0.3 & -2.92 & -2.91 & -34.21 \\ \hline
$D_a$13 & 2.3 & 2.05 & 1.96 & 0.35 & 1.25 & 1.06 & 0.72 & -3.01 & -1.83 & -4.22 & -3.55 & -67.75 \\ \hline
\end{tabular}
\vspace{.1cm}
\begin{flushleft} See Table~\ref{tab:datasets} for descriptions of the different datasets and Table~\ref{tab:models} for descriptions of the models.
\end{flushleft}
\label{table:AbsL1Pi}
\end{adjustwidth}
\end{table}

\begin{table}[!htbp] %[!ht]
\begin{adjustwidth}{-2.25in}{0in}
\centering
\caption{
{\bf Losses Using gMARD}}
\begin{tabular}{|>{\columncolor{LightCyan}}l||c|c|c|c|c|c|c|c|c|c|c|c|}
\hline
\rowcolor[gray]{0.9}
& $M^{w}_{gpr}$ & $M^{ws}_{gpr}$ & $M_{rf}$ & $M_{gpr}$ & $M^{s}_{gpr}$ & $M_{svr}$ & $M_{ridge}$ & $M_{knn}$ & $M^{lin}_{svr}$ & $M^{s}_{svr}$ & $M_{wnn}$ & $M_{nn}$ \\ \hline
%Model & gpr_be & gpr_be_AllPat_AllMeals & rf4 & gpr_IndPat_AllMeals & gpr_AllPat_AllMeals & svr1 & ridge & KNN10U & svr1_lin & svr1_allpats & wnn & NN \\ \hline
$D_e$1 & 6.88 & 6.89 & 6.95 & 7.01 & 7.02 & 7.1 & 7.22 & 7.23 & 7.38 & 7.58 & 7.68 & 8.63 \\ \hline
$D_e$2 & 6.74 & 6.74 & 6.78 & 6.88 & 6.89 & 6.97 & 7.06 & 6.96 & 7.21 & 7.33 & 7.44 & 7.99 \\ \hline
$D_e$3 & 7.1 & 7.12 & 7.29 & 7.21 & 7.23 & 7.32 & 7.49 & 7.51 & 7.7 & 7.88 & 7.9 & 7.94 \\ \hline
$D_e$4 & 7.12 & 7.13 & 7.27 & 7.22 & 7.23 & 7.33 & 7.5 & 7.52 & 7.7 & 7.9 & 7.92 & 7.92 \\ \hline
$D_e$5 & 7.2 & 7.23 & 7.25 & 7.33 & 7.35 & 7.41 & 7.5 & 7.47 & 7.69 & 7.75 & 7.94 & 7.89 \\ \hline
$D_e$6 & 7.08 & 7.1 & 7.28 & 7.18 & 7.21 & 7.27 & 7.49 & 7.47 & 7.69 & 7.81 & 7.94 & 7.89 \\ \hline
$D_e$7 & 7.2 & 7.23 & 7.25 & 7.33 & 7.36 & 7.41 & 7.5 & 7.47 & 7.69 & 7.76 & 7.91 & 7.94 \\ \hline
$D_e$8 & 7.13 & 7.14 & 7.3 & 7.23 & 7.24 & 7.3 & 7.58 & 7.54 & 7.79 & 7.84 & 7.93 & 8.11 \\ \hline
$D_e$9 & 6.65 & 6.66 & 6.64 & 6.71 & 6.71 & 6.85 & 6.83 & 6.9 & 6.87 & 7.04 & 7.04 & 8.11 \\ \hline
$D_e$10 & 7.24 & 7.26 & 7.4 & 7.35 & 7.37 & 7.53 & 7.57 & 7.55 & 7.78 & 7.91 & 7.96 & 8.22 \\ \hline
$D_e$11 & \bf{6.48} & 6.5 & 6.63 & 6.64 & 6.67 & 6.74 & 6.87 & 6.86 & 7.05 & 7.22 & 7.19 & 7.91 \\ \hline
$D_e$12 & 7.33 & 7.34 & 7.42 & 7.45 & 7.48 & 7.5 & 7.69 & 7.62 & 7.92 & 7.88 & 7.91 & 8.64 \\ \hline
$D_e$13 & 6.65 & 6.66 & 6.71 & 6.68 & 6.7 & 6.77 & 6.94 & 6.94 & 6.95 & 7.03 & 7.04 & 9.39 \\ \hline
\hline
$D_a$1 & 6.97 & 6.97 & 7.01 & 7.15 & 7.16 & 7.28 & 7.33 & 7.44 & 7.51 & 7.67 & 7.68 & 8.48 \\ \hline
$D_a$2 & 6.75 & 6.75 & 6.8 & 6.87 & 6.88 & 7.03 & 7.01 & 7.06 & 7.16 & 7.24 & 7.28 & 8.2 \\ \hline
$D_a$3 & 7.26 & 7.27 & 7.34 & 7.35 & 7.37 & 7.48 & 7.61 & 7.69 & 7.82 & 8.01 & 8.05 & 8.53 \\ \hline
$D_a$4 & 7.26 & 7.27 & 7.33 & 7.37 & 7.38 & 7.51 & 7.62 & 7.66 & 7.82 & 8.02 & 8.07 & 8.63 \\ \hline
$D_a$5 & 7.35 & 7.37 & 7.34 & 7.48 & 7.51 & 7.6 & 7.63 & 7.63 & 7.84 & 7.88 & 8.08 & 8.23 \\ \hline
$D_a$6 & 7.23 & 7.25 & 7.32 & 7.32 & 7.35 & 7.45 & 7.61 & 7.63 & 7.81 & 7.95 & 8.04 & 8.44 \\ \hline
$D_a$7 & 7.35 & 7.37 & 7.32 & 7.48 & 7.51 & 7.6 & 7.63 & 7.63 & 7.84 & 7.9 & 8.0 & 8.3 \\ \hline
$D_a$8 & 7.27 & 7.29 & 7.36 & 7.36 & 7.39 & 7.46 & 7.7 & 7.68 & 7.91 & 7.96 & 8.08 & 8.65 \\ \hline
$D_a$9 & 6.81 & 6.81 & 6.85 & 6.89 & 6.88 & 7.02 & 7.06 & 7.1 & 7.08 & 7.23 & 7.28 & 9.06 \\ \hline
$D_a$10 & 7.37 & 7.38 & 7.41 & 7.48 & 7.5 & 7.66 & 7.68 & 7.69 & 7.89 & 8.03 & 8.03 & 8.57 \\ \hline
$D_a$11 & 6.6 & 6.61 & 6.67 & 6.72 & 6.73 & 6.84 & 6.9 & 6.98 & 7.02 & 7.22 & 7.26 & 8.05 \\ \hline
$D_a$12 & 7.46 & 7.47 & 7.47 & 7.57 & 7.59 & 7.63 & 7.83 & 7.76 & 8.06 & 7.98 & 8.03 & 15.13 \\ \hline
$D_a$13 & 6.8 & 6.81 & 6.91 & 6.84 & 6.86 & 6.94 & 7.14 & 7.13 & 7.14 & 7.24 & 7.27 & 11.0 \\ \hline
\end{tabular}
\vspace{.1cm}
\begin{flushleft} See Table~\ref{tab:datasets} for descriptions of the different datasets and Table~\ref{tab:models} for descriptions of the models. Here all values are multiplied by 10 so that they are the same order of magnitude as in Table~\ref{table:AbsL1Te}.
\end{flushleft}
\label{table:AdjL1Te}
\end{adjustwidth}
\end{table}

\begin{table}[!htbp] %[!ht]
\begin{adjustwidth}{-2.25in}{0in}
\centering
\caption{
{\bf Percentage Improvements Corresponding to gMARD.}}
\begin{tabular}{|>{\columncolor{LightCyan}}l||c|c|c|c|c|c|c|c|c|c|c|c|}
\hline
\rowcolor[gray]{0.9}
& $M^{w}_{gpr}$ & $M^{ws}_{gpr}$ & $M_{rf}$ & $M_{gpr}$ & $M^{s}_{gpr}$ & $M_{svr}$ & $M_{ridge}$ & $M_{knn}$ & $M^{lin}_{svr}$ & $M^{s}_{svr}$ & $M_{wnn}$ & $M_{nn}$ \\ \hline
%Model & gpr_be & gpr_be_AllPat_AllMeals & rf4 & gpr_IndPat_AllMeals & gpr_AllPat_AllMeals & svr1 & ridge & KNN10U & svr1_lin & svr1_allpats & wnn & NN \\ \hline
$D_e$1 & \bf{18.38} & 18.28 & 17.53 & 16.81 & 16.63 & 15.73 & 14.29 & 14.18 & 12.44 & 10.05 & 8.87 & -2.41 \\ \hline
$D_e$2 & 16.37 & 16.35 & 15.83 & 14.59 & 14.45 & 13.49 & 12.39 & 13.57 & 10.54 & 9.04 & 7.65 & 0.76 \\ \hline
$D_e$3 & 18.29 & 18.02 & 16.06 & 17.05 & 16.85 & 15.71 & 13.76 & 13.52 & 11.37 & 9.33 & 9.13 & 8.61 \\ \hline
$D_e$4 & 18.08 & 17.93 & 16.27 & 16.88 & 16.81 & 15.58 & 13.64 & 13.5 & 11.41 & 9.05 & 8.86 & 8.86 \\ \hline
$D_e$5 & 17.09 & 16.84 & 16.54 & 15.64 & 15.36 & 14.67 & 13.67 & 14.06 & 11.49 & 10.84 & 8.65 & 9.22 \\ \hline
$D_e$6 & 18.56 & 18.27 & 16.22 & 17.41 & 17.04 & 16.32 & 13.84 & 14.03 & 11.52 & 10.12 & 8.66 & 9.18 \\ \hline
$D_e$7 & 17.09 & 16.76 & 16.52 & 15.64 & 15.29 & 14.69 & 13.67 & 14.06 & 11.49 & 10.64 & 9.0 & 8.61 \\ \hline
$D_e$8 & 18.03 & 17.84 & 16.03 & 16.84 & 16.66 & 15.98 & 12.77 & 13.29 & 10.41 & 9.84 & 8.82 & 6.75 \\ \hline
$D_e$9 & 11.56 & 11.47 & 11.72 & 10.82 & 10.81 & 8.9 & 9.18 & 8.32 & 8.65 & 6.42 & 6.37 & -7.75 \\ \hline
$D_e$10 & 16.68 & 16.43 & 14.87 & 15.38 & 15.14 & 13.35 & 12.84 & 13.07 & 10.46 & 8.98 & 8.36 & 5.35 \\ \hline
$D_e$11 & 17.28 & 17.02 & 15.34 & 15.17 & 14.77 & 13.99 & 12.29 & 12.4 & 10.01 & 7.78 & 8.15 & -1.04 \\ \hline
$D_e$12 & 15.72 & 15.51 & 14.61 & 14.26 & 13.98 & 13.72 & 11.35 & 12.34 & 8.88 & 9.35 & 9.02 & 0.62 \\ \hline
$D_e$13 & 11.65 & 11.47 & 10.81 & 11.15 & 10.89 & 10.0 & 7.79 & 7.81 & 7.65 & 6.49 & 6.36 & -24.83 \\ \hline
\hline
$D_a$1 & 16.96 & 16.95 & 16.46 & 14.85 & 14.73 & 13.24 & 12.71 & 11.35 & 10.48 & 8.62 & 8.47 & -1.04 \\ \hline
$D_a$2 & 13.43 & 13.42 & 12.78 & 11.9 & 11.81 & 9.83 & 10.13 & 9.48 & 8.2 & 7.1 & 6.7 & -5.09 \\ \hline
$D_a$3 & 17.1 & 16.98 & 16.09 & 15.98 & 15.85 & 14.53 & 13.05 & 12.19 & 10.66 & 8.47 & 8.07 & 2.52 \\ \hline
$D_a$4 & 17.01 & 16.91 & 16.24 & 15.78 & 15.7 & 14.16 & 12.98 & 12.52 & 10.69 & 8.41 & 7.86 & 1.39 \\ \hline
$D_a$5 & 16.03 & 15.84 & 16.12 & 14.5 & 14.25 & 13.2 & 12.82 & 12.88 & 10.45 & 9.93 & 7.74 & 5.95 \\ \hline
$D_a$6 & 17.39 & 17.13 & 16.38 & 16.33 & 15.98 & 14.92 & 13.08 & 12.85 & 10.77 & 9.23 & 8.2 & 3.61 \\ \hline
$D_a$7 & 16.03 & 15.8 & 16.41 & 14.5 & 14.21 & 13.12 & 12.82 & 12.88 & 10.46 & 9.72 & 8.63 & 5.17 \\ \hline
$D_a$8 & 16.99 & 16.8 & 15.89 & 15.89 & 15.65 & 14.8 & 12.08 & 12.32 & 9.68 & 9.12 & 7.73 & 1.19 \\ \hline
$D_a$9 & 11.96 & 11.98 & 11.45 & 10.97 & 10.98 & 9.21 & 8.7 & 8.14 & 8.49 & 6.45 & 5.81 & -17.13 \\ \hline
$D_a$10 & 15.81 & 15.63 & 15.32 & 14.5 & 14.34 & 12.46 & 12.3 & 12.11 & 9.85 & 8.22 & 8.2 & 2.13 \\ \hline
$D_a$11 & 14.5 & 14.4 & 13.59 & 13.04 & 12.79 & 11.45 & 10.59 & 9.64 & 9.07 & 6.54 & 6.03 & -4.28 \\ \hline
$D_a$12 & 14.74 & 14.69 & 14.68 & 13.53 & 13.36 & 12.81 & 10.49 & 11.35 & 7.94 & 8.88 & 8.34 & -72.84 \\ \hline
$D_a$13 & 12.11 & 11.99 & 10.67 & 11.53 & 11.32 & 10.31 & 7.69 & 7.8 & 7.65 & 6.43 & 6.01 & -42.24 \\ \hline
\end{tabular}
\vspace{.1cm}
\begin{flushleft} See Table~\ref{tab:datasets} for descriptions of the different datasets and Table~\ref{tab:models} for descriptions of the models.
\end{flushleft}
\label{table:AdjL1Pi}
\end{adjustwidth}
\end{table}

% RMSE SECTION

\begin{table}[!htbp] %[!ht]
\begin{adjustwidth}{-2.25in}{0in}
\centering
\caption{
{\bf Losses Using RMSE.}}
\begin{tabular}{|>{\columncolor{LightCyan}}l||c|c|c|c|c|c|c|c|c|c|c|c|}
\hline
%Model & gpr_be_basic & gpr_be_AllPat_AllMeals_basic & gpr_IndPat_AllMeals_basic & gpr_AllPat_AllMeals_basic & rf4_basic & svr1_basic & svr1_lin_basic & bayes_basic & KNN10U_basic & svr1_allpats_basic & wnn_basic & NN_basic \\ \hline
\rowcolor[gray]{0.9}
& $M^{w}_{gpr}$ & $M^{ws}_{gpr}$ & $M_{gpr}$ & $M^{s}_{gpr}$ & $M_{rf}$ & $M_{svr}$ & $M^{lin}_{svr}$ & $M_{knn}$ & $M_{wnn}$ & $M^{s}_{svr}$ & $M_{ridge}$ & $M_{nn}$ \\ \hline
%Model & gpr_be & gpr_be_AllPat_AllMeals & gpr_IndPat_AllMeals & gpr_AllPat_AllMeals & rf4 & svr1 & svr1_lin & KNN10U & wnn & svr1_allpats & ridge & NN \\ \hline
$D_e$1 & \bf{3.47} & 3.47 & 3.52 & 3.53 & 3.52 & 3.59 & 3.57 & 3.66 & 3.64 & 3.64 & 3.58 & 4.37 \\ \hline
$D_e$2 & 3.57 & 3.57 & 3.6 & 3.61 & 3.63 & 3.69 & 3.75 & 3.73 & 3.72 & 3.7 & 3.65 & 6.71 \\ \hline
$D_e$3 & 3.59 & 3.6 & 3.62 & 3.63 & 3.66 & 3.66 & 3.67 & 3.81 & 3.83 & 3.84 & 3.69 & 4.24 \\ \hline
$D_e$4 & 3.6 & 3.6 & 3.63 & 3.63 & 3.66 & 3.67 & 3.68 & 3.83 & 3.85 & 3.84 & 3.7 & 6.02 \\ \hline
$D_e$5 & 3.63 & 3.63 & 3.66 & 3.67 & 3.66 & 3.69 & 3.69 & 3.81 & 3.83 & 3.85 & 3.71 & 4.17 \\ \hline
$D_e$6 & 3.59 & 3.59 & 3.61 & 3.62 & 3.65 & 3.65 & 3.68 & 3.81 & 3.84 & 3.83 & 3.7 & 4.35 \\ \hline
$D_e$7 & 3.63 & 3.63 & 3.66 & 3.67 & 3.66 & 3.71 & 3.69 & 3.81 & 3.81 & 3.84 & 3.71 & 4.24 \\ \hline
$D_e$8 & 3.6 & 3.6 & 3.63 & 3.63 & 3.66 & 3.65 & 3.69 & 3.82 & 3.83 & 3.86 & 3.72 & 5.76 \\ \hline
$D_e$9 & 3.64 & 3.64 & 3.66 & 3.66 & 3.66 & 3.76 & 3.74 & 3.78 & 3.78 & 3.77 & 4.73 & 5.94 \\ \hline
$D_e$10 & 3.64 & 3.64 & 3.67 & 3.68 & 3.69 & 3.73 & 3.71 & 3.84 & 3.86 & 3.84 & 3.73 & 4.6 \\ \hline
$D_e$11 & 3.58 & 3.58 & 3.63 & 3.65 & 3.66 & 3.68 & 3.7 & 3.82 & 3.85 & 3.88 & 3.72 & 6.61 \\ \hline
$D_e$12 & 3.65 & 3.65 & 3.69 & 3.7 & 3.7 & 3.68 & 3.72 & 3.84 & 3.84 & 3.88 & 3.75 & 6.07 \\ \hline
$D_e$13 & 3.64 & 3.64 & 3.65 & 3.65 & 3.71 & 3.71 & 3.69 & 3.78 & 3.8 & 3.78 & 4.52 & 254.95 \\ \hline
\hline
$D_a$1 & 3.58 & 3.58 & 3.63 & 3.63 & 3.63 & 3.71 & 3.71 & 3.76 & 3.81 & 3.78 & 3.71 & 5.56 \\ \hline
$D_a$2 & 3.66 & 3.66 & 3.7 & 3.71 & 3.72 & 3.78 & 3.82 & 3.83 & 3.87 & 3.85 & 3.78 & 6.48 \\ \hline
$D_a$3 & 3.72 & 3.73 & 3.75 & 3.76 & 3.78 & 3.8 & 3.82 & 3.94 & 3.99 & 3.99 & 3.95 & 5.01 \\ \hline
$D_a$4 & 3.73 & 3.73 & 3.76 & 3.77 & 3.78 & 3.81 & 3.83 & 3.94 & 3.99 & 3.98 & 4.11 & 5.17 \\ \hline
$D_a$5 & 3.76 & 3.76 & 3.8 & 3.81 & 3.77 & 3.85 & 3.86 & 3.94 & 3.98 & 3.99 & 3.96 & 5.04 \\ \hline
$D_a$6 & 3.72 & 3.72 & 3.75 & 3.76 & 3.77 & 3.79 & 3.83 & 3.94 & 3.98 & 3.98 & 3.94 & 6.64 \\ \hline
$D_a$7 & 3.76 & 3.77 & 3.8 & 3.81 & 3.77 & 3.87 & 3.86 & 3.94 & 3.97 & 3.98 & 3.96 & 5.02 \\ \hline
$D_a$8 & 3.72 & 3.73 & 3.75 & 3.76 & 3.78 & 3.78 & 3.84 & 3.95 & 3.98 & 3.99 & 3.88 & 9.06 \\ \hline
$D_a$9 & 3.7 & 3.7 & 3.73 & 3.73 & 3.74 & 3.84 & 3.84 & 3.86 & 3.85 & 3.84 & 5.16 & 7.55 \\ \hline
$D_a$10 & 3.76 & 3.77 & 3.8 & 3.8 & 3.81 & 3.86 & 3.85 & 3.96 & 4.02 & 3.98 & 4.07 & 5.7 \\ \hline
$D_a$11 & 3.71 & 3.71 & 3.77 & 3.78 & 3.78 & 3.82 & 3.83 & 3.98 & 4.01 & 4.02 & 5.81 & 13.15 \\ \hline
$D_a$12 & 3.77 & 3.78 & 3.81 & 3.81 & 3.82 & 3.8 & 3.86 & 3.97 & 3.99 & 4.01 & 3.9 & 20.01 \\ \hline
$D_a$13 & 3.7 & 3.7 & 3.72 & 3.72 & 3.79 & 3.79 & 3.76 & 3.86 & 3.86 & 3.86 & 4.86 & 41.78 \\ \hline
\end{tabular}
\vspace{.1cm}
\begin{flushleft} See Table~\ref{tab:datasets} for descriptions of the different datasets and Table~\ref{tab:models} for descriptions of the models.
\end{flushleft}
\label{table:AbsL1Te}
\end{adjustwidth}
\end{table}

\begin{table}[!htbp] %[!ht]
\begin{adjustwidth}{-2.25in}{0in}
\centering
\caption{
{\bf Percentage Improvements for RMSE.}}
\begin{tabular}{|>{\columncolor{LightCyan}}l||c|c|c|c|c|c|c|c|c|c|c|c|}
\hline
\rowcolor[gray]{0.9}
& $M^{w}_{gpr}$ & $M^{ws}_{gpr}$ & $M_{gpr}$ & $M^{s}_{gpr}$ & $M_{rf}$ & $M_{svr}$ & $M^{lin}_{svr}$ & $M_{knn}$ & $M^{s}_{svr}$ & $M_{wnn}$ & $M_{ridge}$ & $M_{nn}$ \\ \hline
%Model & gpr_be & gpr_be_AllPat_AllMeals & gpr_IndPat_AllMeals & gpr_AllPat_AllMeals & rf4 & svr1 & svr1_lin & KNN10U & svr1_allpats & wnn & ridge & NN \\ \hline
$D_e$1 & 2.98 & 2.93 & 1.65 & 1.5 & 1.6 & -0.3 & 0.38 & -2.38 & -1.62 & -1.75 & 0.05 & -22.06 \\ \hline
$D_e$2 & 2.06 & 2.18 & 1.22 & 1.2 & 0.61 & -1.05 & -2.65 & -2.15 & -1.39 & -1.95 & -0.14 & -83.81 \\ \hline
$D_e$3 & 4.29 & 4.15 & 3.53 & 3.36 & 2.55 & 2.47 & 2.24 & -1.51 & -2.31 & -2.16 & 1.59 & -12.86 \\ \hline
$D_e$4 & 4.05 & 3.97 & 3.3 & 3.2 & 2.46 & 2.2 & 2.0 & -1.97 & -2.21 & -2.52 & 1.41 & -60.39 \\ \hline
$D_e$5 & 3.39 & 3.26 & 2.53 & 2.33 & 2.54 & 1.56 & 1.62 & -1.45 & -2.51 & -2.18 & 1.16 & -11.05 \\ \hline
$D_e$6 & 4.45 & 4.24 & 3.75 & 3.48 & 2.74 & 2.77 & 2.02 & -1.43 & -2.04 & -2.3 & 1.46 & -15.86 \\ \hline
$D_e$7 & 3.39 & 3.15 & 2.53 & 2.27 & 2.57 & 1.04 & 1.62 & -1.45 & -2.36 & -1.6 & 1.16 & -13.0 \\ \hline
$D_e$8 & 4.14 & 4.02 & 3.3 & 3.18 & 2.44 & 2.88 & 1.76 & -1.73 & -2.71 & -1.95 & 0.88 & -53.49 \\ \hline
$D_e$9 & 0.93 & 0.89 & 0.35 & 0.31 & 0.24 & -2.47 & -2.02 & -2.95 & -2.77 & -3.0 & -28.94 & -61.96 \\ \hline
$D_e$10 & 3.11 & 2.94 & 2.19 & 1.99 & 1.63 & 0.64 & 1.23 & -2.24 & -2.32 & -2.89 & 0.61 & -22.57 \\ \hline
$D_e$11 & \bf{6.25} & 6.05 & 4.81 & 4.39 & 4.17 & 3.43 & 2.88 & -0.26 & -1.63 & -1.07 & 2.54 & -73.25 \\ \hline
$D_e$12 & 2.84 & 2.67 & 1.7 & 1.47 & 1.39 & 1.96 & 1.01 & -2.26 & -3.31 & -2.24 & -0.1 & -61.66 \\ \hline
$D_e$13 & 0.8 & 0.71 & 0.61 & 0.48 & -0.98 & -1.2 & -0.6 & -3.09 & -3.03 & -3.54 & -23.2 & -6846.22 \\ \hline
\hline
$D_a$1 & 2.49 & 2.49 & 1.11 & 1.03 & 1.18 & -1.01 & -0.99 & -2.54 & -3.04 & -3.73 & -1.1 & -51.5 \\ \hline
$D_a$2 & 1.8 & 1.85 & 0.64 & 0.62 & 0.2 & -1.51 & -2.44 & -2.71 & -3.17 & -3.79 & -1.28 & -73.68 \\ \hline
$D_a$3 & 3.67 & 3.61 & 2.87 & 2.75 & 2.29 & 1.69 & 1.1 & -1.93 & -3.12 & -3.26 & -2.26 & -29.54 \\ \hline
$D_a$4 & 3.49 & 3.43 & 2.64 & 2.53 & 2.3 & 1.45 & 0.82 & -2.07 & -3.0 & -3.12 & -6.4 & -33.72 \\ \hline
$D_a$5 & 2.72 & 2.6 & 1.63 & 1.46 & 2.52 & 0.47 & 0.13 & -1.99 & -3.18 & -2.93 & -2.55 & -30.45 \\ \hline
$D_a$6 & 3.83 & 3.62 & 3.06 & 2.78 & 2.43 & 1.92 & 0.88 & -1.99 & -2.87 & -2.94 & -1.85 & -71.78 \\ \hline
$D_a$7 & 2.72 & 2.53 & 1.63 & 1.39 & 2.51 & -0.07 & 0.13 & -1.99 & -3.07 & -2.69 & -2.55 & -29.79 \\ \hline
$D_a$8 & 3.68 & 3.53 & 2.87 & 2.67 & 2.32 & 2.28 & 0.66 & -2.11 & -3.32 & -2.91 & -0.26 & -134.27 \\ \hline
$D_a$9 & 1.04 & 1.03 & 0.17 & 0.13 & -0.27 & -2.68 & -2.7 & -3.23 & -2.92 & -3.11 & -38.13 & -102.02 \\ \hline
$D_a$10 & 2.69 & 2.58 & 1.77 & 1.6 & 1.34 & 0.25 & 0.44 & -2.44 & -3.1 & -4.01 & -5.4 & -47.41 \\ \hline
$D_a$11 & 5.02 & 4.94 & 3.51 & 3.28 & 3.3 & 2.1 & 1.91 & -1.91 & -2.9 & -2.74 & -48.79 & -236.69 \\ \hline
$D_a$12 & 2.41 & 2.34 & 1.52 & 1.36 & 1.29 & 1.61 & 0.03 & -2.6 & -3.74 & -3.17 & -0.91 & -417.57 \\ \hline
$D_a$13 & 1.01 & 0.93 & 0.51 & 0.38 & -1.48 & -1.46 & -0.76 & -3.39 & -3.28 & -3.48 & -30.03 & -1018.57 \\ \hline
\end{tabular}
\vspace{.1cm}
\begin{flushleft} See Table~\ref{tab:datasets} for descriptions of the different datasets and Table~\ref{tab:models} for descriptions of the models.
\end{flushleft}
\label{table:AbsL1Pi}
\end{adjustwidth}
\end{table}

% gRMSE SECTION

\begin{table}[!htbp] %[!ht]
\begin{adjustwidth}{-2.25in}{0in}
\centering
\caption{
{\bf Losses Using gRMSE.}}
\begin{tabular}{|>{\columncolor{LightCyan}}l||c|c|c|c|c|c|c|c|c|c|c|c|}
\hline
%Model & gpr_be_basic & gpr_be_AllPat_AllMeals_basic & gpr_IndPat_AllMeals_basic & gpr_AllPat_AllMeals_basic & rf4_basic & svr1_basic & svr1_lin_basic & bayes_basic & KNN10U_basic & svr1_allpats_basic & wnn_basic & NN_basic \\ \hline
\rowcolor[gray]{0.9}
& $M^{w}_{gpr}$ & $M^{ws}_{gpr}$ & $M_{svr}$ & $M_{gpr}$ & $M^{s}_{gpr}$ & $M_{rf}$ & $M^{lin}_{svr}$ & $M_{knn}$ & $M_{wnn}$ & $M^{s}_{svr}$ & $M_{ridge}$ & $M_{nn}$ \\ \hline
%Model & gpr_be & gpr_be_AllPat_AllMeals & svr1 & gpr_IndPat_AllMeals & gpr_AllPat_AllMeals & rf4 & svr1_lin & KNN10U & wnn & svr1_allpats & ridge & NN \\ \hline
$D_e$1 & \bf{4.86} & 4.87 & 4.96 & 4.94 & 4.95 & 4.93 & 4.97 & 5.12 & 5.14 & 5.13 & 5.02 & 5.88 \\ \hline
$D_e$2 & 5.0 & 4.99 & 5.09 & 5.05 & 5.04 & 5.07 & 5.15 & 5.18 & 5.22 & 5.2 & 5.1 & 8.62 \\ \hline
$D_e$3 & 5.06 & 5.07 & 5.07 & 5.1 & 5.11 & 5.16 & 5.14 & 5.35 & 5.44 & 5.45 & 5.21 & 5.72 \\ \hline
$D_e$4 & 5.07 & 5.08 & 5.09 & 5.11 & 5.12 & 5.16 & 5.15 & 5.37 & 5.46 & 5.45 & 5.22 & 8.68 \\ \hline
$D_e$5 & 5.11 & 5.12 & 5.12 & 5.16 & 5.17 & 5.15 & 5.16 & 5.34 & 5.45 & 5.43 & 5.23 & 5.62 \\ \hline
$D_e$6 & 5.05 & 5.06 & 5.06 & 5.09 & 5.1 & 5.15 & 5.15 & 5.34 & 5.44 & 5.43 & 5.22 & 5.92 \\ \hline
$D_e$7 & 5.11 & 5.12 & 5.14 & 5.16 & 5.17 & 5.16 & 5.16 & 5.34 & 5.41 & 5.43 & 5.23 & 5.78 \\ \hline
$D_e$8 & 5.07 & 5.08 & 5.07 & 5.12 & 5.13 & 5.16 & 5.18 & 5.36 & 5.42 & 5.46 & 5.26 & 7.28 \\ \hline
$D_e$9 & 5.1 & 5.1 & 5.17 & 5.13 & 5.13 & 5.11 & 5.14 & 5.27 & 5.33 & 5.34 & 6.12 & 7.35 \\ \hline
$D_e$10 & 5.13 & 5.13 & 5.17 & 5.18 & 5.19 & 5.21 & 5.2 & 5.38 & 5.47 & 5.45 & 5.27 & 6.1 \\ \hline
$D_e$11 & 4.99 & 5.0 & 5.05 & 5.07 & 5.09 & 5.09 & 5.11 & 5.31 & 5.38 & 5.43 & 5.18 & 8.06 \\ \hline
$D_e$12 & 5.15 & 5.16 & 5.13 & 5.22 & 5.23 & 5.23 & 5.24 & 5.4 & 5.45 & 5.48 & 5.32 & 7.63 \\ \hline
$D_e$13 & 5.1 & 5.11 & 5.11 & 5.12 & 5.12 & 5.17 & 5.14 & 5.27 & 5.36 & 5.34 & 5.97 & 256.36 \\ \hline
\hline
$D_a$1 & 5.02 & 5.02 & 5.12 & 5.1 & 5.1 & 5.07 & 5.15 & 5.26 & 5.38 & 5.35 & 5.2 & 7.08 \\ \hline
$D_a$2 & 5.12 & 5.12 & 5.21 & 5.19 & 5.19 & 5.19 & 5.26 & 5.33 & 5.44 & 5.42 & 5.27 & 7.99 \\ \hline
$D_a$3 & 5.24 & 5.25 & 5.26 & 5.29 & 5.3 & 5.32 & 5.35 & 5.54 & 5.67 & 5.67 & 5.51 & 6.55 \\ \hline
$D_a$4 & 5.25 & 5.26 & 5.28 & 5.31 & 5.31 & 5.32 & 5.36 & 5.54 & 5.65 & 5.66 & 5.65 & 6.73 \\ \hline
$D_a$5 & 5.3 & 5.31 & 5.33 & 5.36 & 5.37 & 5.31 & 5.39 & 5.53 & 5.65 & 5.64 & 5.52 & 6.57 \\ \hline
$D_a$6 & 5.23 & 5.25 & 5.25 & 5.28 & 5.3 & 5.31 & 5.36 & 5.53 & 5.64 & 5.65 & 5.49 & 8.14 \\ \hline
$D_a$7 & 5.3 & 5.31 & 5.35 & 5.36 & 5.38 & 5.31 & 5.39 & 5.53 & 5.64 & 5.65 & 5.52 & 6.44 \\ \hline
$D_a$8 & 5.25 & 5.25 & 5.24 & 5.3 & 5.31 & 5.32 & 5.38 & 5.54 & 5.65 & 5.67 & 5.47 & 10.56 \\ \hline
$D_a$9 & 5.19 & 5.19 & 5.28 & 5.24 & 5.24 & 5.24 & 5.26 & 5.39 & 5.43 & 5.44 & 6.57 & 9.26 \\ \hline
$D_a$10 & 5.3 & 5.31 & 5.35 & 5.36 & 5.37 & 5.37 & 5.39 & 5.56 & 5.71 & 5.67 & 5.63 & 7.19 \\ \hline
$D_a$11 & 5.17 & 5.18 & 5.24 & 5.26 & 5.27 & 5.26 & 5.28 & 5.54 & 5.62 & 5.65 & 7.22 & 14.56 \\ \hline
$D_a$12 & 5.33 & 5.33 & 5.3 & 5.39 & 5.39 & 5.38 & 5.43 & 5.57 & 5.66 & 5.68 & 5.51 & 21.96 \\ \hline
$D_a$13 & 5.18 & 5.19 & 5.22 & 5.22 & 5.22 & 5.29 & 5.24 & 5.39 & 5.44 & 5.45 & 6.31 & 43.34 \\ \hline
\end{tabular}
\vspace{.1cm}
\begin{flushleft} See Table~\ref{tab:datasets} for descriptions of the different datasets and Table~\ref{tab:models} for descriptions of the models.
\end{flushleft}
\label{table:AbsL1Te}
\end{adjustwidth}
\end{table}

\begin{table}[!htbp] %[!ht]
\begin{adjustwidth}{-2.25in}{0in}
\centering
\caption{
{\bf Percentage Improvements for gRMSE.}}
\begin{tabular}{|>{\columncolor{LightCyan}}l||c|c|c|c|c|c|c|c|c|c|c|c|}
\hline
\rowcolor[gray]{0.9}
& $M^{w}_{gpr}$ & $M^{ws}_{gpr}$ & $M_{svr}$ & $M_{gpr}$ & $M^{s}_{gpr}$ & $M_{rf}$ & $M^{lin}_{svr}$ & $M_{knn}$ & $M_{wnn}$ & $M^{s}_{svr}$ & $M_{ridge}$ & $M_{nn}$ \\ \hline
%Model & gpr_be & gpr_be_AllPat_AllMeals & svr1 & gpr_IndPat_AllMeals & gpr_AllPat_AllMeals & rf4 & svr1_lin & KNN10U & wnn & svr1_allpats & ridge & NN \\ \hline
$D_e$1 & 2.95 & 2.88 & 1.01 & 1.45 & 1.32 & 1.58 & 0.9 & -2.07 & -2.49 & -2.38 & -0.13 & -17.3 \\ \hline
$D_e$2 & 1.22 & 1.37 & -0.51 & 0.29 & 0.31 & -0.19 & -1.75 & -2.4 & -3.19 & -2.67 & -0.86 & -70.36 \\ \hline
$D_e$3 & 4.33 & 4.15 & 4.06 & 3.54 & 3.36 & 2.42 & 2.75 & -1.11 & -2.82 & -3.12 & 1.36 & -8.25 \\ \hline
$D_e$4 & 4.05 & 3.93 & 3.7 & 3.26 & 3.16 & 2.33 & 2.52 & -1.51 & -3.26 & -3.1 & 1.2 & -64.21 \\ \hline
$D_e$5 & 3.37 & 3.2 & 3.22 & 2.49 & 2.26 & 2.51 & 2.36 & -1.06 & -3.05 & -2.63 & 1.1 & -6.34 \\ \hline
$D_e$6 & 4.52 & 4.29 & 4.35 & 3.8 & 3.51 & 2.63 & 2.62 & -1.06 & -2.92 & -2.71 & 1.32 & -11.89 \\ \hline
$D_e$7 & 3.37 & 3.1 & 2.83 & 2.49 & 2.24 & 2.49 & 2.36 & -1.06 & -2.25 & -2.69 & 1.1 & -9.25 \\ \hline
$D_e$8 & 4.1 & 3.98 & 4.19 & 3.17 & 3.08 & 2.35 & 2.08 & -1.43 & -2.56 & -3.21 & 0.49 & -37.66 \\ \hline
$D_e$9 & 0.23 & 0.18 & -1.16 & -0.39 & -0.43 & 0.03 & -0.5 & -3.07 & -4.24 & -4.44 & -19.8 & -43.78 \\ \hline
$D_e$10 & 3.05 & 2.87 & 2.13 & 2.08 & 1.87 & 1.51 & 1.6 & -1.83 & -3.56 & -3.19 & 0.3 & -15.42 \\ \hline
$D_e$11 & \bf{4.86} & 4.68 & 3.67 & 3.46 & 3.08 & 2.91 & 2.53 & -1.12 & -2.53 & -3.44 & 1.35 & -53.6 \\ \hline
$D_e$12 & 2.54 & 2.4 & 2.95 & 1.24 & 1.03 & 1.08 & 0.89 & -2.05 & -2.98 & -3.61 & -0.87 & -44.25 \\ \hline
$D_e$13 & 0.16 & 0.06 & 0.02 & -0.09 & -0.24 & -1.12 & -0.53 & -3.12 & -4.81 & -4.44 & -16.7 & -4914.91 \\ \hline
\hline
$D_a$1 & 2.7 & 2.71 & 0.66 & 1.15 & 1.08 & 1.62 & 0.01 & -2.1 & -4.29 & -3.75 & -0.88 & -37.37 \\ \hline
$D_a$2 & 1.27 & 1.35 & -0.52 & -0.03 & -0.03 & -0.12 & -1.35 & -2.85 & -4.83 & -4.47 & -1.55 & -54.15 \\ \hline
$D_a$3 & 3.84 & 3.76 & 3.5 & 2.95 & 2.82 & 2.42 & 1.89 & -1.53 & -4.02 & -3.98 & -1.04 & -20.19 \\ \hline
$D_a$4 & 3.64 & 3.56 & 3.17 & 2.69 & 2.56 & 2.48 & 1.65 & -1.56 & -3.62 & -3.91 & -3.61 & -23.41 \\ \hline
$D_a$5 & 2.83 & 2.69 & 2.26 & 1.61 & 1.42 & 2.68 & 1.22 & -1.51 & -3.63 & -3.51 & -1.26 & -20.54 \\ \hline
$D_a$6 & 4.03 & 3.8 & 3.72 & 3.18 & 2.86 & 2.59 & 1.76 & -1.51 & -3.42 & -3.63 & -0.79 & -49.27 \\ \hline
$D_a$7 & 2.83 & 2.61 & 1.84 & 1.61 & 1.34 & 2.69 & 1.22 & -1.51 & -3.41 & -3.56 & -1.26 & -18.21 \\ \hline
$D_a$8 & 3.79 & 3.65 & 3.88 & 2.87 & 2.68 & 2.49 & 1.29 & -1.68 & -3.66 & -3.93 & -0.33 & -93.6 \\ \hline
$D_a$9 & 0.47 & 0.45 & -1.35 & -0.52 & -0.55 & -0.56 & -0.91 & -3.38 & -4.25 & -4.46 & -26.04 & -77.7 \\ \hline
$D_a$10 & 2.79 & 2.65 & 1.92 & 1.75 & 1.57 & 1.45 & 1.15 & -1.95 & -4.8 & -3.98 & -3.21 & -31.92 \\ \hline
$D_a$11 & 4.12 & 4.06 & 2.88 & 2.48 & 2.26 & 2.57 & 2.26 & -2.6 & -4.08 & -4.66 & -33.8 & -169.81 \\ \hline
$D_a$12 & 2.25 & 2.2 & 2.89 & 1.23 & 1.08 & 1.34 & 0.5 & -2.23 & -3.81 & -4.19 & -1.23 & -302.7 \\ \hline
$D_a$13 & 0.5 & 0.43 & -0.25 & -0.09 & -0.24 & -1.6 & -0.58 & -3.5 & -4.47 & -4.61 & -21.15 & -731.8 \\ \hline
\end{tabular}
\vspace{.1cm}
\begin{flushleft} See Table~\ref{tab:datasets} for descriptions of the different datasets and Table~\ref{tab:models} for descriptions of the models.
\end{flushleft}
\label{table:AbsL1Pi}
\end{adjustwidth}
\end{table}

\clearpage

\subsection{Feature Stacking with models trained on all patients}
\label{stacking}

\begin{figure}[t] % [!ht]
\centering
\ShowFig{
\includegraphics[width=\textwidth]{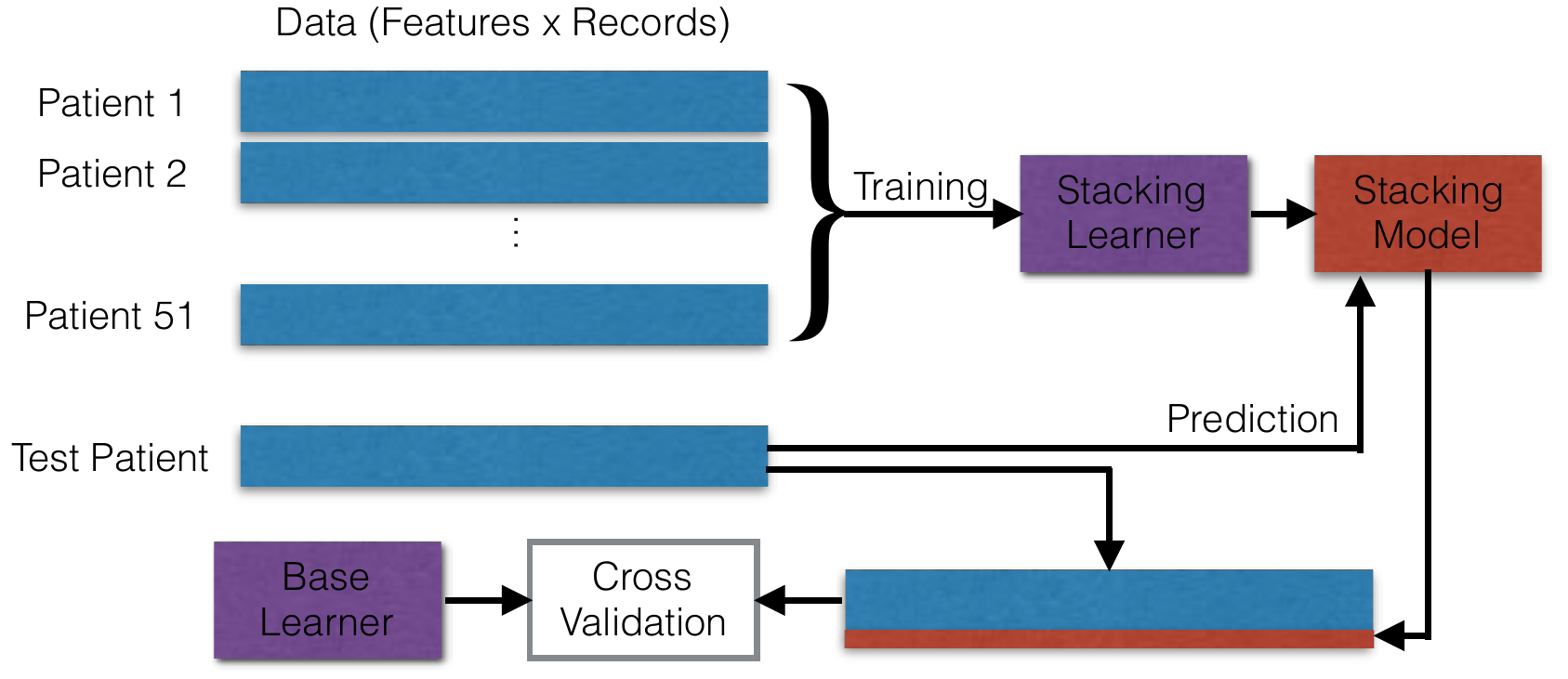}
}
\caption{{\bf How we Implement Stacking.}}
\label{fig:stacking}
\end{figure}

Up to this point, all the previous learning algorithms described have been trained specifically on a single patient and then evaluated on new data from that same patient. 
To incorporate the available data from the other patients in the dataset, 
we use stacking (originally introduced as ``stacked generalization''%
% by David Wolpert
~\cite{wolpert1992stacked})%
, which
involves training higher levels of learners on a combination of meta-data/data with the predictions of other basic learners~\cite{marton2013ensembles}. 
In our case, this involves training a stacking learner on all the other patients to produce a stacking model,
which is then used to make BG predictions for each record associated with the current patient. % under test.
(See Fig~\ref{fig:stacking} for a description of the stacking process that we use).
These BG predictions then become a new feature for the patient under test.%, 
%\annote[RG]{and CV with}{?? and we run CV over these features ??, using ?? to evaluate ??} 
%this patient proceeds as described before. 
Specifically, we use SVR as our stacking learner and 
use GPR (resp., confidence weighted GPR and SVR) as our base learner,
to produce the $M^{s}_{gpr}$ model (resp., $M^{ws}_{gpr}$, $M^{s}_{svr}$). Here the superscript s indicates that the model uses stacking, whereas the superscript w indicates confidence weighting.
%\note[RG]{What is the notation here?}
% These produce the models $M^{s}_{gpr}$, $M^{ws}_{gpr}$, and $M^{s}_{svr}$ respectively.

% The rationale behind
This stacking approach is motivated by the 
realization % fact
that, as our data is self-reported, some patients (such as patient 16) have a large number of records, while other patients have relatively few records.
Consider the datasets (features and outcome) for two patients,
$D_1 = \{(\mathbf{x}^1_i, y^1_i)| i = 1..n\}$ and 
$D_2 = \{(\mathbf{x}^2_j, y^2_j)| j = 1..m\}$ where $m \ll n$. 
Let
$f_1: \mathbf{X} \rightarrow Y$ and $f_2: \mathbf{X} \rightarrow Y$
be the true underlying glucose functions for these patients. Both functions are sufficiently complex that a learning algorithm cannot learn a good approximation $\hat{f}_2$ when only learning with $D_2$,
but a learning algorithm can learn a good approximation $\hat{f}_1$ when only learning with $D_1$. 
If $f_1$ and $f_2$ are closely related in some sense
-- \eg affinely related  $f_2(\mathbf{x}) =  a\,f_1(\mathbf{x}) + b$ --
then $\hat{f}_2$ can be learned using $D_1$ and $D_2$
by first learning $\hat{f}_1$. 
As another perspective,
we could view 
this stacking approach as mimicking health care professionals,
who apply the knowledge that they have acquired from previous patients, to a new patient. 
% From this perspective the stacking model represents knowledge that has been acquired from previous patients, and that is applied to a new patient when learning about that new patient.
\comment{
Let's say that we have the datasets (features and outcome) for two patients, $D_1 = \{(\mathbf{x}_i, y_i)| i = 0..n\}$ and 
$D_2 = \{(\mathbf{x}_j, y_j)| j = 0..m\}$ where $m \ll n$. 
\note[RG]{Reworded}
Additionally, 
we'll denote 
$f_1: \mathbf{x}_i \rightarrow y_i$ and $f_2: \mathbf{x}_j \rightarrow y_j$ as the true underlying glucose functions for these patient where both functions are sufficiently complex that a learning algorithm cannot learn a good approximation $\hat{f}_2$ when only learning with $D_2$ but that a learning algorithm can learn a good approximation $\hat{f}_1$ when only learning with $D_1$. If it is the case that $f_1$ and $f_2$ are closely related in some sense
, e.g. affinely related so that $f_2 =  af_1 + b$,
then $\hat{f}_1$ can be learned using $D_1$ and $D_2$ can be used to learn the relationship between the two functions.
}

% Include only the SI item label in the paragraph heading. Use the \nameref{label} command to cite SI items in the text.

% \paragraph*{S1 Appendix.}
% \label{S1_Appendix}
% {\bf Lorem Ipsum.} Maecenas convallis mauris sit amet sem ultrices gravida. Etiam eget sapien nibh. Sed ac ipsum eget enim egestas ullamcorper nec euismod ligula. Curabitur fringilla pulvinar lectus consectetur pellentesque.

% \paragraph*{S1 Table.}
% \label{S1_Table}

\clearpage

\newpage
\clearpage

\section*{Competing interests}
  The authors declare that they have no competing interests.

\section*{Author's contributions}
\begin{tabular}{l l}
{\bf Conceptualization:} & NCB EAR RG\\
{\bf Data curation:} & NCB EAR RG\\
{\bf Formal analysis:} & NCB\\
{\bf Funding acquisition:} & RG\\
{\bf Investigation:} & NCB\\
{\bf Methodology:} & NCB EAR RG\\
{\bf Project administration:} & EAR RG\\
{\bf Resources:} & EAR RG\\
{\bf Software:} & NCB\\
{\bf Supervision:} & EAR RG\\
{\bf Validation:} & NCB EAR\\
{\bf Visualization:} & NCB RG\\
{\bf Writing - original draft:} & NCB EAR RG\\
{\bf Writing - review and editing:} & NCB EAR RG\\
\end{tabular}

\section*{Acknowledgements}
This study was funded in part by a pilot project grant from ADI.
RG gratefully acknowledges funding from AMII, and NSERC;
ER from ADI;
NB from NSERC.
The authors gratefully acknowledge help from Haipeng (Paul) Li,
as well as our visiting summer interns from the Indian Institute of Technology, Kharagpur.
% and interns:
% Abhinav Agrawalla,
% Prachi Agrawal and
% Pranjal Daga.

%\nolinenumbers

\bibliographystyle{plos2015}

\end{document}